\documentclass{article}

\PassOptionsToPackage{numbers}{natbib}

\usepackage[preprint]{neurips_2026}

\usepackage[utf8]{inputenc} 
\usepackage[T1]{fontenc}    
\usepackage{hyperref}       
\usepackage{url}            
\usepackage{booktabs}       
\usepackage{amsfonts}       
\usepackage{nicefrac}       
\usepackage{microtype}      
\usepackage{xcolor}         

\usepackage{graphicx}
\usepackage{subcaption}
\usepackage{multirow}
\usepackage{xurl}
\usepackage{siunitx}
\usepackage{tabularx}
\usepackage{wrapfig}

\usepackage{amsmath}
\usepackage{amssymb}
\usepackage{mathtools}
\usepackage{amsthm}
\usepackage{thmtools}

\usepackage[capitalize,noabbrev]{cleveref}

\theoremstyle{plain}
\newtheorem{theorem}{Theorem}[section]
\newtheorem{proposition}[theorem]{Proposition}

\theoremstyle{definition}

\theoremstyle{remark}
\newtheorem{remark}[theorem]{Remark}

\usepackage{amsmath,amsfonts,bm}

\def\eqref#1{equation~\ref{#1}}

\def\1{\bm{1}}

\def\vzero{{\bm{0}}}

\def\vmu{{\bm{\mu}}}

\def\vb{{\bm{b}}}
\def\vc{{\bm{c}}}

\def\ve{{\bm{e}}}

\def\vm{{\bm{m}}}

\def\vu{{\bm{u}}}
\def\vv{{\bm{v}}}
\def\vw{{\bm{w}}}
\def\vx{{\bm{x}}}
\def\vy{{\bm{y}}}
\def\vz{{\bm{z}}}

\def\mA{{\bm{A}}}

\def\mC{{\bm{C}}}

\def\mI{{\bm{I}}}

\def\mM{{\bm{M}}}

\def\mLambda{{\bm{\Lambda}}}
\def\mSigma{{\bm{\Sigma}}}

\DeclareMathAlphabet{\mathsfit}{\encodingdefault}{\sfdefault}{m}{sl}
\SetMathAlphabet{\mathsfit}{bold}{\encodingdefault}{\sfdefault}{bx}{n}

\def\gA{{\mathcal{A}}}

\def\gC{{\mathcal{C}}}
\def\gD{{\mathcal{D}}}

\def\gF{{\mathcal{F}}}
\def\gG{{\mathcal{G}}}

\def\gN{{\mathcal{N}}}

\def\gP{{\mathcal{P}}}
\def\gQ{{\mathcal{Q}}}

\def\gU{{\mathcal{U}}}

\def\sN{{\mathbb{N}}}

\def\sP{{\mathbb{P}}}

\def\sR{{\mathbb{R}}}

\newcommand{\E}{\mathbb{E}}

\newcommand{\R}{\mathbb{R}}

\newcommand{\normltwo}{L^2}

\newcommand{\dTV}{d_{\mathrm{TV}}}
\newcommand{\tinygray}[1]{{\scriptsize\textcolor{gray}{#1}}}

\usepackage[textsize=tiny]{todonotes}
\usepackage[ruled,vlined]{algorithm2e}

\title{Conformal Prediction for Time-Dependent PDEs}

\author{%
  Joshua Stiller\thanks{Corresponding author (\texttt{joshua.stiller@lmu.de})} \\
  LMU Munich, \\MCML\\
  \And
  Annika Schneider \\
  TU Munich, Helmholtz AI, \\relAI, MCML\\
  \And
  Eyke Hüllermeier \\
  LMU Munich, \\MCML, DFKI \\
}

\begin{document}

\maketitle
\setcounter{footnote}{0}

\begin{abstract}
  Uncertainty quantification is crucial in scientific machine learning, where models inform safety-critical tasks such as flood forecasting, financial risk management, and thermal control in machines. 
  Conformal prediction provides distribution-free coverage guarantees, but in time-dependent settings common to physics and engineering, these guarantees can break down, leading to systematic undercoverage.
  We study this problem in the context of surrogate models for time-dependent physical systems described by partial differential equations.
  We prove that in a function space setting, distributions at arbitrarily close times can be mutually singular, making exact coverage guarantees impossible. We further show that in discretized settings, the total variation distance between distributions often saturates exponentially fast with the grid resolution, mandating restrictions on the spatial discretization to ensure coverage. Finally, for certain settings, we show how to use principled weighted conformal prediction to obtain finite-sample coverage guarantees over growing time horizons.
\end{abstract}

\section{Introduction}\label{sec:intro}

Many problems in physics and engineering, such as weather prediction, aerodynamics, and financial modeling, are governed by partial differential equations (PDEs). Classical numerical solvers are accurate but computationally expensive, often scaling poorly with the dimensionality of the data. AI-based surrogate models have emerged as a promising alternative, providing fast approximations of PDE solutions. Prominent examples include physics-informed neural networks (PINNs)~\citep{raissi_physics-informed_2019}, DeepONets~\citep{lu_learning_2021}, and Neural Operators~\citep{li_neural_2020,li_fourier_2021}.

Despite these advances, surrogate models still lack principled mechanisms for uncertainty quantification, which are vital, as model predictions often inform safety-critical decisions in science and engineering. Conformal prediction (CP) \cite{vovk_algorithmic_2022} provides a principled framework, producing distribution-free uncertainty sets with guaranteed marginal coverage. These guarantees, however, rely on exchangeability between calibration and test samples, a condition that is frequently violated in time-dependent PDEs.

\begin{figure*}[ht]
    \centering
    \includegraphics[width=\linewidth]{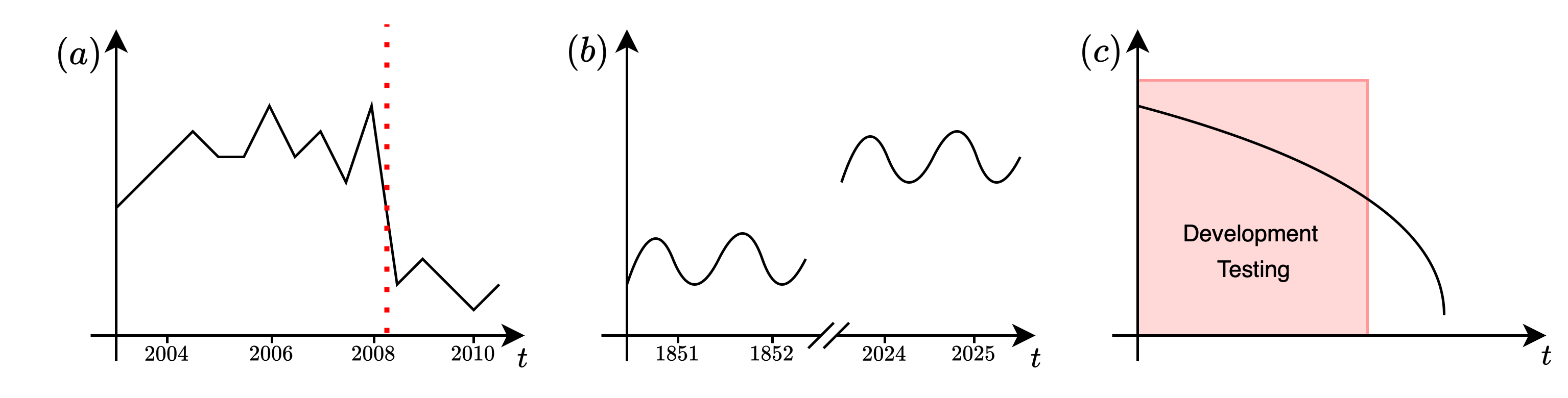}
    \caption{Examples of different types of non-stationarity that can arise in time-dependent systems. 
    \emph{(a)} Sudden distribution shift due to external shocks, illustrated by a stock market crash. 
    \emph{(b)} Long-term structural changes, as in climate time series. 
    \emph{(c)} Limited observation window during system development, where calibration and testing occur on different parts of the trajectory.}
    \label{fig:nonstationarity}
\end{figure*}

\paragraph{Non-stationarity in time-dependent PDEs}
Let $u_t$ denote the solution of a time-dependent PDE at time $t$. In practice, we are interested in predicting $u_{t+\delta}$ for several time steps $\delta > 0$, beyond the available training and calibration data. Unless $u_t$ is a stationary process, test samples follow a different distribution than observed calibration samples, breaking the exchangeability assumption required by conformal prediction (see \cref{fig:cp_timestep}).
This type of non-stationarity is ubiquitous: sudden shocks (e.g., stock market crashes), long-term structural changes (e.g., climate trends), and limited development windows (e.g., laboratory testing) all produce systematic shifts in the data distribution (see \cref{fig:nonstationarity}).

\paragraph{Why existing methods fall short}
Existing CP approaches for time-dependent settings either rely on assumptions like mixing, which generally fail for PDE dynamics, or relax to asymptotic guarantees that average coverage over time and cannot guarantee per-step coverage \citep{gibbs_adaptive_2021, gibbs_conformal_2024}.
In theory, weighted conformal prediction (WCP)~\cite{tibshirani_conformal_2019, barber_conformal_2023} provides finite-sample coverage under distribution shift, but it requires likelihood ratios $p_{\text{test}}/p_{\text{cal}}$ that are generally unavailable for PDE surrogates. As a result, no current approach provides finite-sample coverage at every future time step. This raises a structural question: when can CP guarantees hold for time-dependent PDEs, and how should CP be applied to obtain them? We answer with theoretical limits and practical guidelines.

\paragraph{Our contributions}
\begin{enumerate}
\item We show that in function-space, already for the heat equation with Gaussian initial conditions, the total variation (TV) distance between solutions at any two distinct times is maximal. This rules out non-exchangeable CP guarantees on the function-space view common in the neural operator literature.
\item We show that for discretized PDEs the TV distance often approaches $1$ exponentially fast with the grid granularity. This gives a practical guideline: rather than calibrating on the full prediction grid, we should restrict to a coarser grid or the spatial subregion of interest, as we do in our real-world experiments.
\item We identify assumptions under which closed-form likelihood ratios and finite-sample WCP guarantees at every future time step exist in the discretized setting and further lift this discrete guarantee to the true continuous solution.
\end{enumerate}
We empirically validate these results on synthetic PDEs and three real-world datasets.

\section{Related work}\label{sec:related}
CP for PDEs is closely related to conformal prediction for time series.
Existing approaches can be roughly categorized as follows:

\paragraph{Access to exchangeable data}
Some work treats entire solution trajectories as exchangeable units (e.g., \citet{ma_calibrated_2024} for Neural Operators,
\citet{moya_conformalized-deeponet_2025} for DeepONets, or
\citet{gray_guaranteed_2025} and \citet{gopakumar_uncertainty_2026, gopakumar_calibrated_2025} for general surrogate models). These approaches fail when extrapolated to future time steps beyond the observed trajectory, as the distribution of the solution changes with time.

\paragraph{Relaxing to asymptotic guarantees}
Instead of requiring marginal coverage at every time step, asymptotic coverage provides guarantees on average over time. The seminal work on adaptive conformal prediction \citep{gibbs_adaptive_2021} tightens or loosens conformal bands based on past miscoverage. Several refinements have been proposed \citep[e.g.,][]{angelopoulos_conformal_2023, xu_sequential_2023, gibbs_conformal_2024}.
For non-critical tasks and slowly drifting time series, this can be a viable approach. In many PDE problems, however, undercoverage at a single time step can be fatal and it is a priori unknown how fast changes appear.

\paragraph{Exploiting additional properties}
To extend CP beyond exchangeability, some work assumes additional properties on the data. For example, stationarity, (strong) mixing of the process or residuals, or requirements on the accuracy of the predictive model
\citep[e.g.,][]{oliveira_split_2024, xu_sequential_2023, auer_conformal_2023, xu_conformal_2024, xu_conformal_2021}.
Depending on the assumptions, both asymptotic and per-time-step guarantees can be obtained.

\paragraph{Distributional assumptions}
Generally, CP is distribution-free, but when certain properties of the distribution are known, they can be exploited to obtain coverage guarantees without exchangeability. 
In full generality, \citet{barber_conformal_2023} show that coverage can be lower bounded in terms of the total variation (TV) distance between calibration and test distributions. \citet{farinhas_non-exchangeable_2024} extend this to conformal risk control.
Under covariate or label shift, more specialized approaches achieve marginal coverage when the distribution shift can be quantified \citep{tibshirani_conformal_2019, podkopaev_distribution-free_2021}.
In practice, these methods are rarely used in a principled way, as the required density ratios are typically unknown.
A related approach comes from \citet{harris_locally_2025}, which replaces global exchangeability with a local assumption across nearby time points.

\section{Background}\label{sec:background}

\subsection{Conformal prediction}\label{sec:cp_background}

Conformal prediction (CP) is a framework for constructing prediction sets with marginal finite-sample coverage guarantees~\cite{vovk_algorithmic_2022}. In the standard split setting, a model is trained on a dataset $\gD_{\mathrm{train}}$ and calibrated on a dataset $\gD_{\mathrm{cal}}$, yielding a set-valued predictor $\mathcal{C}$ such that, for a test sample $(x_{\text{test}}, y_{\text{test}})$,
\[
  \sP\big(y_{\text{test}} \in \mathcal{C}(x_{\text{test}})\big) \;\geq\; 1-\alpha,
\]
where $\alpha$ denotes the miscoverage level. This guarantee relies on \textbf{exchangeability} of calibration and test samples, meaning that their joint distribution is invariant under permutations. When exchangeability is violated, coverage may fail.

\paragraph{Conformal prediction beyond exchangeability}
When calibration and test distributions differ but are related through a likelihood ratio, weighted conformal prediction (WCP) provides a natural extension of standard CP~\cite{tibshirani_conformal_2019}. 
A canonical example is \emph{covariate shift}, where calibration samples $(x_i,y_i)$ for $i\in\{1, \dots, n\}$ are drawn from $\gP_X \times \gP_{Y|X}$, while test samples $(x_{\text{test}},y_{\text{test}})$ follow $\gQ_X \times \gP_{Y|X}$, i.e., the conditional distribution of $Y$ given $X$ remains unchanged. 
In this setting, valid coverage guarantees can be recovered by reweighting the calibration quantiles according to the density ratio
\begin{equation}\label{eq:weighted-cp}
       w_i \;\propto\; \frac{p_{\mathrm{test}}(x_i)}{p_{\mathrm{cal}}(x_i)}, 
       \quad \forall x_i \in \gD_{\mathrm{cal}} \cup \{x_{\text{test}}\}, 
       \quad \sum_i w_i = 1,
\end{equation}
where $p_{\mathrm{cal}}$ and $p_{\mathrm{test}}$ denote the densities of $\gP_X$ and $\gQ_X$ (see Appendix~\ref{app:wcp_formula} for more details).

In the most general case, \citet{barber_conformal_2023} and, similarly, \citet{farinhas_non-exchangeable_2024} provide corrections for the conformal guarantees based on the TV distance between calibration and test distribution:\footnote{The TV distance is originally defined on probability measures, and whenever we write $\dTV(\vx,\vy)$ for random vectors $\vx$ and $\vy$, or $\dTV(\gP,\gQ)$ for probability distributions $\gP$ and $\gQ$, we refer to the TV distance between their corresponding probability measures.}
\begin{equation}\label{eq:tv-distance}
 \sP\big(y_{\mathrm{test}} \in \mathcal{C}(x_{\mathrm{test}})\big)  
 \;\geq\; 1 - \alpha - \sum_{i=1}^n \tilde w_i \, \dTV(\vz, \vz^i),
\end{equation}
where $\tilde w_i=w_i/(w_1+\cdots+w_n + 1)$, with $w_i\in[0,1]$, $\vz = ((x_1, y_1), \dots, (x_{n}, y_{n}), (x_\mathrm{test}, y_{\mathrm{test}}))$ and $\vz^i$ arises from permuting the test pair in $\vz$ with the $i^\mathrm{th}$ calibration pair.
These approaches are rarely used in a principled way because they require knowing the corresponding distributions. We show when and how they can be used on PDE problems to regain coverage over time.

\subsection{PDEs as operator mappings}\label{sec:pde_background}

Many dynamical systems in physics and engineering can be described by evolution equations of the form  
\[
    \frac{\partial u}{\partial t}(\vx,t) = \mathcal{L}_\vx u(\vx,t),
\]  
where $u : \Omega \times [0,\infty) \to \mathbb{R}$ is the state variable, $\vx$ denotes spatial coordinates on a bounded domain $\Omega \subset \mathbb{R}^d$, $t \geq 0$ is the time, and $\mathcal{L}_\vx$ is a (possibly nonlinear) differential operator acting on the spatial variable $\vx$. In this paper, we are interested in the Cauchy-type problem, where we consider boundary conditions on $\bar{\Omega}$ and initial conditions $u(\vx, 0)$ from some Banach space of functions $(\gA, \|\cdot\|_{\gA})$ and are interested in a solution $u(\vx, t)$ in some Banach space of functions $(\gU_t, \|\cdot\|_{\gU_t})$.
We will further only consider well-posed problems with a time-invariant differential operator, where we can define solution operators $\gG_t: \gA \rightarrow \gU_t,\, \gG_t(a) = u(\cdot,t)$ that uniquely map an initial condition to a solution function $u(\vx, t)$ and the map $t \mapsto \gG_t$ is continuous in $t$.
In the rest of the paper, we will assume that all functions come from the same space, so $\gA = \gU_t$ for all $t\geq0$, to simplify the notation, but the results apply more generally.

\section{WCP for time-dependent PDEs}\label{sec:cp_pdes}
\begin{figure*}[t]
    \centering
    \includegraphics[width=\linewidth]{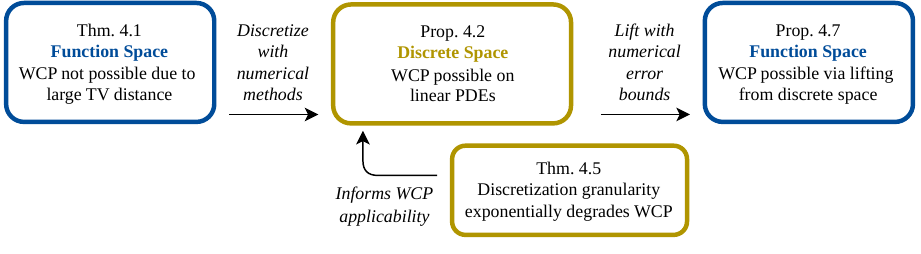}
    \caption{Overview of the theoretical results. 
    The colors differentiate between function space (blue) and discretized space (yellow).
    }
    \label{fig:theorem_overview}
\end{figure*}
We begin by specifying the conformal setup, followed by the main theoretical contributions which we summarize in \cref{fig:theorem_overview}.

\subsection{Problem setting}\label{sec:problem_setting}

\paragraph{From initial conditions to solutions}  
To apply CP in the PDE setting, we first specify the underlying distributions. Let $u(\vx,t)$ be the solution of a PDE and $S_{\delta} : \gU \to \gU$ be the solution operator mapping $u(\vx,t)$ to $u(\vx,t+\delta)$ for any $t\geq0$. We may assume the initial condition (or in practice: our initial measurement) is distributed according to $u(\vx, 0) \sim \gP_0$. Then, $S_{\delta}$ induces a pushforward measure\footnote{We slightly abuse notation here by writing the pushforward in terms of the distribution instead of the measure corresponding to the distribution.} $\gP_{\delta} := (S_{\delta})_\# \gP_0 \,$.
If we now consider consecutive time points, our distribution changes over time:
\[
  \gP_0 \;\xrightarrow{S_{\delta}}\; \gP_{\delta} \;\xrightarrow{S_{\delta}}\; \gP_{2\delta} \;\xrightarrow{S_{\delta}}\; \cdots
\]  
Thus, we obtain a sequence of probability distributions $\{\gP_{k\delta}\}_{k\in\sN}$ on the same function space, evolving under the PDE dynamics. Calibration and test data drawn from different $\gP_t$ are therefore \textbf{not exchangeable}: although they live in the same function space, their distributions shift with time.

\paragraph{Datasets}
Datasets of size $n$ typically consist of pairs $(u_i(\vx,t), u_i(\vx,t+\delta))$, for $i\in\{1,\dots,n\}$, generated from initial conditions $u_i(\vx,0)\sim\gP_0$. In practice, the $u_i$ can either arise from measurements or numerical simulations. In both simulations and real-world settings, the states $u_i(\vx,t)$ are discretized on a grid with $p$ points, yielding vectors in $\sR^p$. In this case, we omit the spatial dependence and denote the discretized solution at time $t$ by $\vu_i^t:=(u_i(x_0, t), \dots, u_i(x_{p-1}, t))^T\in\sR^p$ (bold font). Nevertheless, since neural operators can generalize across grids, the function-space viewpoint remains relevant and is commonly adopted in the literature \citep{harris_locally_2025, gray_guaranteed_2025, mollaali_conformalized_2025}. Accordingly, we consider both the functional representation (\cref{sec:function_space_shifts}) and the discretized setting (\cref{sec:bounds}). Throughout the paper we assume that our calibration dataset consists of pairs $(u_{\mathrm{cal},i}(\vx,t_{\mathrm{cal}}), u_{\mathrm{cal},i}(\vx,t_{\mathrm{cal}}+\delta))$ from some time $t_{\mathrm{cal}}$ and our test dataset consists of pairs $(u_{\mathrm{test}, i}(\vx,t), u_{\mathrm{test},i}(\vx,t+\delta))$ for some $t \geq t_{\mathrm{cal}}$. We omit the subscripts \emph{cal} and \emph{test} for notational simplicity.

\subsection{Distribution shifts in function spaces}\label{sec:function_space_shifts}

We will now consider CP on PDEs, starting with the function-space perspective, where the data points $u_i(\vx,t)$ are elements of a suitable function space. As the data is not exchangeable, we must resort to a more general CP method. WCP is not suitable because the function-space densities\footnote{See Appendix~\ref{app:function_space_dist} for a concrete example of such a function-space distribution.} cannot be evaluated, but if the TV distance between the distributions at times $t$ and $t+\delta$ were moderate, CP coverage guarantees for the prediction at $t+\delta$ could still be recovered using the approach from \cref{eq:tv-distance}.
The following theorem shows that even for simple problems, like the heat equation with Gaussian initial distribution, the TV distance between the solution distributions $\gP_t, \gP_{t+\delta}$ at different time points $t, t+\delta$ is always maximal $\dTV(\gP_t, \gP_{t+\delta}) = 1, \text{for all } t \geq 0, \delta > 0 $.
This is representative of a broader phenomenon that ``[...] measures in infinite-dimensional spaces have a strong tendency of being mutually singular.''~\cite{hairer_introduction_2023}. 
As a direct consequence, regular CP and any method requiring distributional similarity between calibration and test fails. In particular, exact WCP via \cref{eq:weighted-cp} is inapplicable.

\begin{restatable}{theorem}{tvtheorem}\label{thm:heat_eq_no_tv}
Consider the one-dimensional heat equation $\partial_t u = \partial_{xx} u$ on the domain $\Omega=(0,1)$ with Dirichlet boundary conditions, $t \geq 0$.
Suppose the initial condition $u_0(x)$ is sampled from a Gaussian random field $\gP_0 = \gN(\boldsymbol{0}, (\mI - \mLambda)^{-1})$, where $\mLambda$ is the Laplace operator on $\Omega$.
Then, the TV distance between the solution distributions at any two time points is maximal:
\[\dTV(\gP_t, \gP_{t+\delta}) = 1, \quad \forall\, t \geq 0,\ \delta > 0.\]
\end{restatable}
\begin{proof}[Proof sketch:] The proof uses the Feldman--Hájek theorem \citep{da_prato_stochastic_1992}, which shows that two Gaussian measures on a Hilbert space are either equivalent or mutually singular. We then calculate the Cameron--Martin spaces of both measures and show they differ, which tells us that the measures cannot be equivalent. The full proof is provided in Appendix~\ref{app:proof_heat_eq_no_tv}.
\end{proof}

We will now discuss how, despite the issue above, coverage guarantees can be achieved in practice.

\subsection{Coverage guarantees in practice}\label{sec:bounds}

\cref{thm:heat_eq_no_tv} shows that a pure function-space perspective renders TV-based corrections vacuous, even for arbitrarily close time points.
This is not a contradiction to practical CP on PDE surrogate models: in practice, we always observe or simulate on a finite number of degrees of freedom.
Therefore, we shift the perspective to the \emph{discretized state} $ \vu_i^t \in \mathbb{R}^p$. As we still consider the non-exchangeable setting, we either need additional inductive biases or rely on TV-based corrections. We will consider both approaches, starting with the former. 

While CP is inherently distribution-free, related methods such as WCP \citep{tibshirani_conformal_2019, farinhas_non-exchangeable_2024} give up this property in exchange for relaxing the exchangeability assumption. We will now identify a setting relevant to time-dependent PDEs where closed-form likelihood ratios are available so that WCP can be applied.

First, note that in physics, Gaussianity is often assumed for the initial distribution, because the aggregate effect of many small independent perturbations forming the initial condition is approximately Gaussian by the central limit theorem. This is well supported by the literature~\citep{li_fourier_2021,santos_physics-informed_2023,gopakumar_uncertainty_2026,zhou_predicting_2025,gopakumar_calibrated_2025}.

Second, observe that when considering \emph{linear} PDEs solved with standard numerical schemes, the time evolution is affine 
$\vu_i^{t+\delta} = \mA_\delta \vu_i^t + \vb_\delta$, 
with 
$\mA_\delta \in \mathbb{R}^{p\times p}$ 
and 
$\vb_\delta \in \mathbb{R}^p$.
If the initial condition is Gaussian, $\vu_i^0 \sim \mathcal{N}(\vmu_0,\mSigma_0)$, then $\vu_i^t$ remains Gaussian for all $t \ge 0$, with mean $\vmu_t$ and covariance $\mSigma_t$ evolving explicitly. Consequently, for linear PDEs with Gaussian initial distribution the calibration and test distributions admit closed-form likelihood ratios, enabling WCP.

\begin{restatable}{proposition}{marginalCoverageProp}
\label{prop:wcp_linear_pde}
Consider a linear evolution equation on a bounded domain with linear boundary conditions and a deterministic source term. If the calibration dataset at time $t_{\mathrm{cal}}$ and a test sample $u_{\mathrm{test}}^t$ at time $t \geq t_{\mathrm{cal}}$ are generated from the same affine numerical scheme on independent initial conditions sampled from a Gaussian distribution, then we can construct a likelihood-weighted conformal predictor such that
\[
\sP\big( \vu_{\mathrm{test}}^{t+\delta} \in \mathcal{C}^{\mathrm{WCP}}_{t+\delta}( \vu_{\mathrm{test}}^{t}) \big) \ge 1 - \alpha .
\]
\end{restatable}
\begin{proof}[Proof sketch:]
As the initial distribution is Gaussian and the PDE linear, we can show that the solution at each time step is Gaussian with known mean and covariance. With this, we calculate the exact weights for WCP. The full proof is provided in Appendix~\ref{app:proof_wcp_linear_pde}.
\end{proof}

\begin{remark}[Other initial distributions]
    This result extends to any other initial distribution that is closed under affine transformations (e.g., Cauchy, Laplace, and Logistic). We include additional experiments sampling from different initial distributions, including heavy-tailed Cauchy initial conditions, in Appendix~\ref{app:additional_results}.
\end{remark}

We only know the input-time distribution, not the prediction-time one (Gaussian with unknown moments), so \textbf{the distribution itself cannot be used to infer uncertainty directly}.

Even though WCP is theoretically attainable in the setting we just identified, the resulting conformal bands can become too large in practice to provide useful information.
In the next section, we present a way to alleviate this: We show that the \emph{grid granularity} is a powerful lever for controlling the TV distance and conformal band size, and with this, the applicability of WCP.
\subsection{Effect of grid granularity}

\begin{figure}[t]
    \centering
    \begin{subfigure}[t]{0.49\linewidth}
        \centering
        \includegraphics[width=\linewidth]{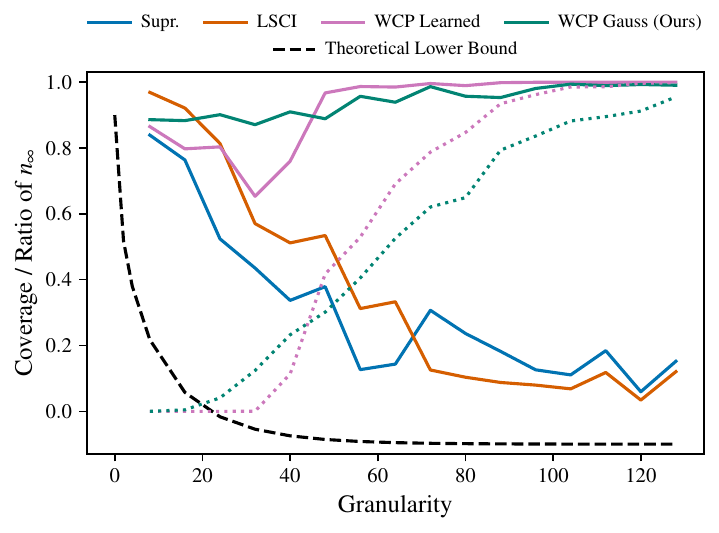}
        \caption{Drift PDE, \cref{thm:tv_gauss_same_cov}}
        \label{fig:granularity}
    \end{subfigure}\hfill
    \begin{subfigure}[t]{0.49\linewidth}
        \centering
        \includegraphics[width=\linewidth]{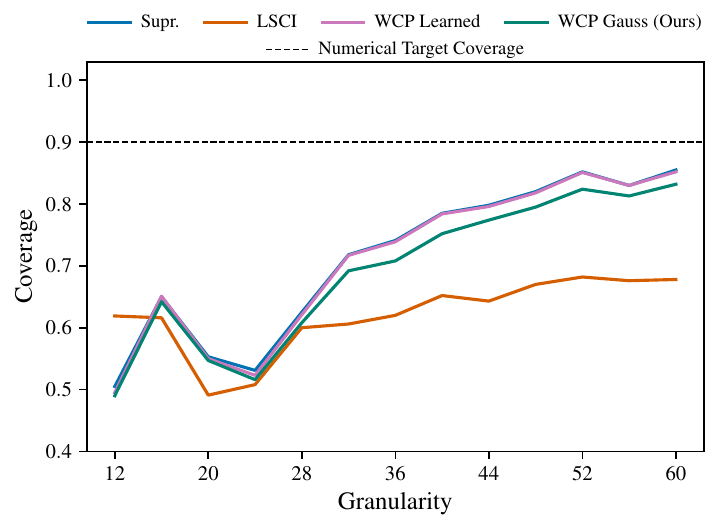}
        \caption{Backward heat equation, \cref{prop:cp_lifting}}
        \label{fig:analytic_heat_lifting}
    \end{subfigure}
    \caption{Visualization for \cref{thm:tv_gauss_same_cov} and \cref{prop:cp_lifting}. \emph{(a)} Coverage decays exponentially with grid granularity for the baselines, while WCP grows infinite bands (dotted lines). The theoretical lower bound follows \citet{barber_conformal_2023}. \emph{(b)} Coverage of the analytical solution of the backward heat equation with numerical conformal bands. With finer granularity, coverage over the analytic solution converges to the target as the reconstruction error vanishes. Also here, the coverage decline effect of (a) would start dominating at higher granularity. More in Appendix~\ref{app:add_exp_expl}.}
    \label{fig:analytic_heat}
\end{figure}

So far we worked with $\vu_i^t$ at arbitrary granularity. Refining the grid recovers the function-space limit, where \cref{thm:heat_eq_no_tv} gives $\dTV=1$, raising the question of how granularity affects CP. We analyze this question for the \emph{drift PDE} which represents slight non-stationary behavior. The following proposition shows that for this PDE, the TV distance approaches $1$ exponentially fast in the grid size, so standard CP approaches become inapplicable even under slight drifts.

\begin{restatable}{proposition}{blowupTheorem}\label{thm:tv_gauss_same_cov}
Consider the drift PDE $\partial_t u(x,t)=c,$ with $x\in(0,1), t\ge 0, c\in\sR$, and compatible boundary conditions. Define $\vz$ and $\vz^i$ as in \cref{eq:tv-distance} with $n$ i.i.d.\ calibration pairs at times $(t_1, t_1+\delta)$ and a test pair at time $(t_2, t_2+\delta)$ from an affine scheme on a grid with $p$ points and Gaussian initial conditions. Then, the total variation distance between the laws of $\vz$ and $\vz^i$ is
\[
  \dTV\!\left(\vz,\vz^i\right)
  \;=\;
  2\Phi\!\Big(\frac{\sqrt{2pc^2}\,|t_1 - t_2|}{2}\Big)-1,
\]
where $\Phi$ is the cumulative distribution function of the standard normal distribution.
Hence, $\dTV(\vz,\vz^i)\to 1$ exponentially fast in the grid size $p$ and with it, the coverage guarantee from \citet{barber_conformal_2023} vanishes exponentially.

Moreover, for likelihood-ratio weighting as in \cref{eq:weights}
evaluated at the distribution centers $\vx_{\mathrm{test}}:=\E[\vu_{n+1}^{t_2}]$ and $\vx_{\mathrm{cal}}:=\E[\vu_i^{t_1}]$, the test weight satisfies
\[
w_{\mathrm{test}}=1-O\!\big(n e^{-pc^2(t_1 - t_2)^2}\big), \quad   
w_{\mathrm{cal}}=O\!\big(e^{-pc^2(t_1-t_2)^2}\big). 
\]
Hence, the weighted conformal bands become infinite at an exponential rate.
\end{restatable}
\begin{proof}[Proof sketch:]
The TV distance follows from \cref{thm:tv_general_drift} with $\vw=\mathbf{1}$ and $\mSigma_0=\mI_p$, and for the weights, we calculate the explicit density ratio. The full proof is provided in Appendix~\ref{app:proof_tv_gauss_same_cov}.
\end{proof}
We visualize \cref{thm:tv_gauss_same_cov} in \cref{fig:granularity}. The drift example provides a good heuristic that a slight non-stationarity can lead to exponential decay of the coverage guarantees for general (non)linear PDEs. We extend this result to general linear PDEs in \cref{app:prop:general_tv_distance}.

The exponential growth in $p$ relies on white-noise initial conditions, where every grid point carries independent information about the drift. In real data, neighboring grid points are usually correlated, which requires a more general statement. The following theorem replaces the grid size $p$ by an \emph{effective dimension} $p_{\mathrm{eff}}$, which is small when the drift points along directions of high initial variance. Restricting CP to a subregion of the domain reduces the effective dimension and thereby improves the CP guarantees, though other approaches can be more suitable depending on the covariance structure of the data.

\begin{restatable}{theorem}{generalDriftTheorem}\label{thm:tv_general_drift}
Consider the drift PDE on a grid with $p$ points, $\partial_t \vu^t=c\,\vw$ with $t\ge 0$, drift strength $c\in\sR$, and drift profile $\vw\in\sR^p$ (e.g., $\vw=\mathbf{1}$ for a constant drift), and let the initial conditions be non-degenerate Gaussian, $\vu^0\sim\gN(\vmu_0,\mSigma_0)$. For $\vz$ and $\vz^i$ as in \cref{thm:tv_gauss_same_cov},
\[
  \dTV\!\left(\vz,\vz^i\right)
  \;=\;
  2\Phi\!\Big(\frac{\sqrt{2\,p_{\mathrm{eff}}\,c^2}\,|t_1 - t_2|}{2}\Big)-1, \quad \text{with } p_{\mathrm{eff}}:= \vw^\top\mSigma_0^{-1}\vw.
\]
\end{restatable}
\begin{proof}[Proof sketch:]
Whitening the covariance reduces the statement to the equal-covariance identity used for \cref{thm:tv_gauss_same_cov}. The full proof is provided in Appendix~\ref{app:proof_tv_general_drift}.
\end{proof}
For $\vw=\mathbf{1}$ and $\mSigma_0=\mI_p$, we have $p_{\mathrm{eff}}=p$, which recovers \cref{thm:tv_gauss_same_cov}. Thus, the TV distance is dependent on the smoothness of the initial state and how the PDE drifts relative to it, even approaching $1$ exponentially fast in the grid size. 

Knowing that many PDEs of interest are non-stationary, and that fine grids are often required for accurate predictions, it may seem futile to apply non-exchangeable CP on PDEs at all. However, while the full grid is usually necessary for prediction, we often do not need uncertainty quantification over the whole space. Referencing our real-world examples in \cref{sec:experiments}, we may find that the temperature of Lake Maggiore is most relevant at depths where certain fish are living, or that bleaching of the FRAP target is most interesting in the middle of the plate where the laser is focused. In such cases, \cref{thm:tv_general_drift} tells us to restrict CP to coherent regions of interest, such as the center of the bleach spot or a few neighboring depths in the lake, and leverage the more favorable properties of the marginal distribution. We demonstrate this in our experiments.

\subsection{Applications and extensions of the framework}
The previous theorem showed that non-exchangeable CP should only be applied to a subset of the grid to avoid exponential decay of the coverage guarantees and uninformative conformal bands. While WCP methods ``only'' suffer from vacuous prediction intervals instead of undercoverage, they face additional practical challenges: For numerical data, we can evolve the exact mean and covariance from the initial distribution. For real-world measurements, however, the true mean and covariance must be \emph{estimated} from data, introducing additional uncertainty. While it is common in CP literature that some assumptions cannot be checked on real-world data (e.g., \citep{tibshirani_conformal_2019,barber_conformal_2023}), we nonetheless want to give a \emph{theoretical outlook} on how to account for this estimation uncertainty and recover coverage guarantees when additional assumptions can be made.

\begin{restatable}[Robust WCP under Gaussian distributions]{proposition}{robustCoverage}
\label{thm:robust_coverage}
Under the assumptions of \cref{prop:wcp_linear_pde}, suppose we additionally observe independent estimation samples
$\{\vu^{t_{\mathrm{cal}}}_j\}_{j=1}^{m_0}$ and
$\{\vu^t_j\}_{j=1}^{m_1}$ with $m_k>p$ for $k\in\{0,1\}$, and fix an arbitrary estimation uncertainty $\eta >0$.
Then, we can construct a conformal prediction set $C^{\mathrm{rob}}_{t+\delta}$ with coverage guarantee
\[
\mathbb P\!\left(\vu_{\mathrm{test}}^{t+\delta}\in C^{\mathrm{rob}}_{t+\delta}(\vu_{\mathrm{test}}^t)\right)\ \ge\ 1-\eta-\alpha.
\]
Smaller $\eta$ will usually correspond to wider bands.
\end{restatable}
\begin{proof}[Proof sketch:]
    We use the sample distributions of the estimated means and covariances to construct a $1-\eta$ confidence region over the true means and covariances. Then, we take the worst case quantiles over these regions. The full proof can be found in Appendix~\ref{app:proof:robust_coverage}.
\end{proof}
Without additional structure on the covariance, this procedure is intractable in practice, and only of theoretical value. Exploring tractable problem classes (e.g., with covariances diagonal in Fourier space) is left for future work. Empirically, such corrections are often unnecessary, as mean and covariance estimates converge rapidly with the number of samples. In particular, we find that constructing prediction bands from estimated mean and covariance yields consistent coverage, whereas estimating the full distribution shift, as the original WCP does, often leads to undercoverage.

\subsection{Recovering guarantees for the true solution}

While we saw in \cref{sec:function_space_shifts} that function-space coverage is generally intractable, numerical schemes typically come with deterministic error bounds of order $O(h^k)$ in the grid size $h$. The following proposition lifts a discrete CP set $C_h$ to a conformal set on the true solution by inflating $C_h$ with this error: the convergence \emph{rate} carries over (even though unknown constants typically prevent computing $C$ explicitly), and depends only on the prediction granularity rather than that used for $C_h$. See \cref{fig:analytic_heat_lifting} for an empirical demonstration.
\begin{restatable}{proposition}{certifiedLiftingTheorem}\label{prop:cp_lifting}
Let  $u(\cdot,t)$ be a PDE solution in some function space $(\gU,\|\cdot\|_{\gU})$ and $\vu^t$ be its approximation via some numerical scheme on grid size $h$. Let $C_h$ be a conformal prediction set for the discretized state with coverage $1-\alpha$ and assume that an error bound $\big\|u(\cdot,t) - \mI_h \vu^t\big\|_{\gU} \;\le\; \varepsilon_{h,t}(u_0)$ is available via some reconstruction operator $\mI_h:\R^p\to\gU$ (e.g., piecewise interpolation).
Then we can define a conformal set for the function-space setting, $C\subset\gU$, with $\sP\big(u(\cdot,t)\in C\big)\;\ge\; 1-\alpha$ by
\begin{equation}\label{eq:lifted_set}
  C := \big\{ v\in\gU : \exists\ \vz\in C_h \text{ with } \|v-\mI_h \vz\|_{\gU}\le \varepsilon_{h,t}(u_0)\big\}.
\end{equation}
\end{restatable}

\begin{proof}[Proof sketch:]
We use the error bound to extend the conformal set for the numerical solution using the triangle inequality. The full proof is given in Appendix~\ref{app:proof_cp_lifting} and a visualization in \cref{fig:analytic_heat_lifting}.
\end{proof}

\section{Experiments}\label{sec:experiments}
We demonstrate our theoretical results on both synthetic and real-world data, in particular the coverage of WCP with Gaussian likelihood ratio.
\begin{figure*}[t]
    \centering
    \includegraphics[width=\linewidth]{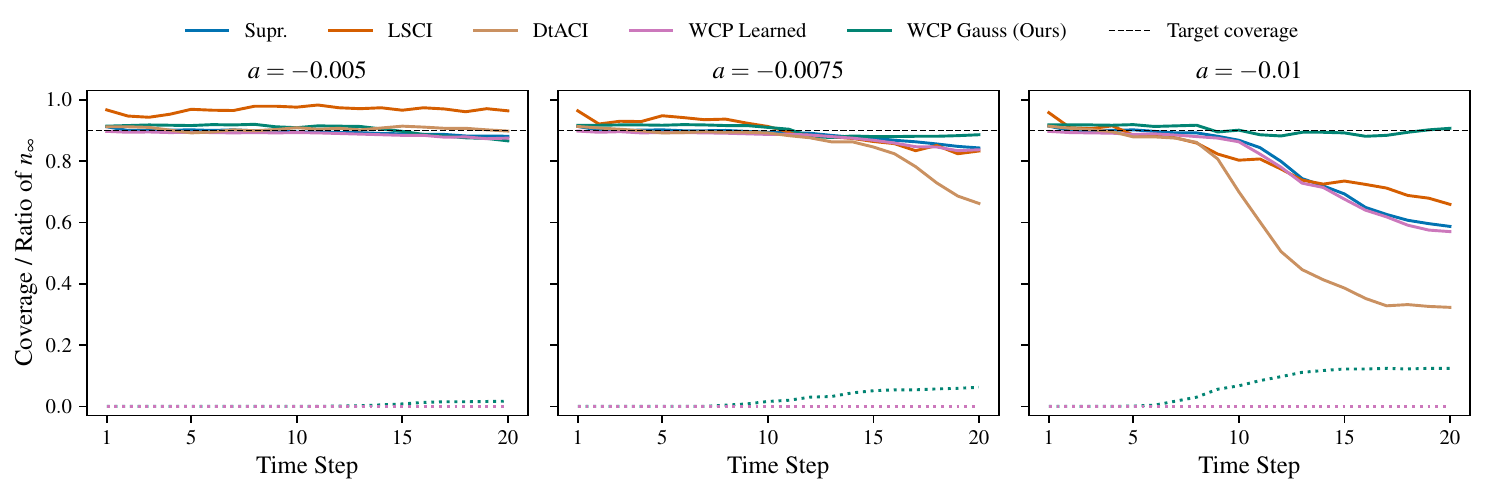}
    \caption{Coverages for varying $a$ and fixed $b=0, c=0.5$ on the recalibration task. The 90\% target is marked by a black dashed line. For WCP Gauss and WCP Learned, the fraction of samples with infinite bandwidth $n_\infty$ is indicated by green and pink dotted lines, respectively. Coverage holds for some parameters, but small variations induce undercoverage at later time steps. Only WCP Gauss meets the target throughout.}
    \label{fig:synthetic_one_row_beta0}
\end{figure*}

\paragraph{Experimental setup}
For the \textbf{synthetic case}, we choose the general second-order PDE framework, which covers many well-known PDEs
$u_t = au_{xx} +bu_{x} + cu, x \in (0,2\pi),\ t \geq 0$
where $a, b, c\in\R$ are tunable parameters and $u_0$ is sampled from a multivariate Gaussian.
We target setups where the PDE becomes more unstable over time ($a<0$). Otherwise, the PDE smooths out over time, leading to trivial coverage at future time steps.
Hence, we consider $a =\{ -0.005, -0.0075, -0.01\}$ and for the remaining parameters we choose $b,c\in\{-0.5,0,0.5\}$. See \cref{fig:conf_bands} for exemplary data.
For \textbf{real-world scenarios}, we use pulsed-thermography measurements from \citet{wei_pulsed_2023}, diffusion dynamics from FRAP-microscopy (Fluorescence recovery after photobleaching) \cite{wahlstrand_skarstrom_deepfrap_2021} and heat-diffusion from sensors in Lake Maggiore under seasonal drift \cite{scheu_observation_2018} (details in Appendices~\ref{app:heat_pulse},~\ref{app:lake_maggiore}, and~\ref{app:frap}). 
We consider the following tasks, targeting 90\% coverage:\footnote{
    A sample is considered covered only if all spatial points are covered. The score is defined as the maximum absolute deviation over the spatial dimension.
}
\textbf{(1) Recalibration case:} Starting from $t_0 = 0$, we generate states at time steps $t_i = t_0 + i\delta$. We use pairs $(t_i, t_{i+1})$ for calibration and test on $(t_{i+1}, t_{i+2})$ for $i\in\{0, \dots, 18\}$ (we also include the calibration time coverage as the first step). The inverse scenario is obtained by reversing time.
\textbf{(2) General case:} As above, but no recalibration is allowed after $t_0$, corresponding to deployment at an unknown future time. 

\paragraph{Baselines}
We define four baselines for our experiments. The first is a na\"ive implementation (\textbf{Supr.}) which depends on exchangeability \cite{diquigiovanni_conformal_2022}.
Secondly, we use WCP with learned weights, as described in \citet{tibshirani_conformal_2019} (\textbf{WCP Learned}), which serves as our main baseline.
Thirdly, we use the \textbf{LSCI} method \citep{harris_locally_2025}, a novel approach for CP on PDEs (note Appendix~\ref{app:lsci_assumption}).
Finally, we use Dynamically-Tuned Adaptive Conformal Inference (\textbf{DtACI}) by \citet{gibbs_conformal_2024}, an updated version of the original Adaptive CP \citep{gibbs_adaptive_2021} with automatic parameter selection. However, as adaptive CP requires an evaluation of the error at every step, this method is only applicable to the recalibration task. We compare these to our WCP method (\textbf{WCP Gauss}), which only estimates mean and covariance for each model input to construct the weights. In cases where the distributional dissimilarity of $u_t$ and $u_{t+\delta}$ is too large, the WCP methods predict infinite bands. We report the fraction of these samples as $n_\infty$. A band is wider the larger the likelihood ratio of its test sample, and infinite once the normalized test weight exceeds $\alpha$ (Appendix~\ref{app:widths}). Note that in safety-critical applications, reporting trivial bands is often a more valuable result than delivering bands with undercoverage.

\paragraph{Additional experiments}
To demonstrate the effect of our derived results, \cref{fig:analytic_heat} provides visualizations of \cref{thm:tv_gauss_same_cov} and \cref{prop:cp_lifting}. More details in Appendix~\ref{app:add_exp_expl}. We additionally include an ablation study in Appendix~\ref{app:ablation_study} that stress-tests WCP Gauss under reduced sample sizes and increased dimensionality. As expected, when the number of samples is insufficient relative to the problem dimensionality, instability in covariance estimation can lead to undercoverage. Coverage is stable across random calibration/test splits and, for the pulsed-thermography data, across plate assignments with retrained surrogates (Appendix~\ref{app:multi_seed}).

\paragraph{Results}
Exemplary figures for synthetic and real-world coverage results can be seen in \cref{fig:synthetic_one_row_beta0} and \cref{fig:heat_coverage} respectively. Results for all remaining parameter combinations (incl. tables) are provided in Appendix~\ref{app:additional_results}.\footnote{
    Bandwidth results are reported in Appendix~\ref{app:additional_results}. Our WCP method does not optimize for bandwidth, but could certainly be extended with more advanced score functions or CP algorithms.
}  
Across both synthetic and real-world tasks, \textbf{WCP Gauss} is the only method that consistently attains the target coverage. In general, WCP Gauss can estimate the weights more easily than WCP Learned, and DtACI would require longer time windows to adapt (which may require costly measurements). This does not mean that coverage with other methods is impossible. While other methods can achieve coverage in specific settings or with careful tuning or more expressive models (e.g., for \emph{WCP Learned}), this behavior does not generalize reliably. In practice, small changes such as grid resolution, PDE parameters, or evaluation time steps can already lead to undercoverage. This issue is often hidden in smoothing forward problems, where residuals shrink over time, but becomes apparent in inverse problems, unstable dynamics, or blow-up regimes. Overall, \textbf{WCP is the only method providing formal guarantees}, and we can see empirically that this is a clear advantage as soon as our system exhibits significant dynamics.

\section{Limitations}
The convergence results for increasing grid granularity (\cref{thm:tv_gauss_same_cov}) should be read as a heuristic for (non)linear PDEs: with increasing granularity, even small drifts can cause the TV distance to deteriorate rapidly. Of course, one can construct PDEs with no drift at all, which may exhibit more favorable TV behavior. Similar considerations apply to the vacuous TV bound in function spaces. Identifying the conditions under which these effects occur more generally is left for future work.
Finally, while the lifting proposition (\cref{prop:cp_lifting}) is independent of the choice of PDE, the proposed WCP method (\cref{prop:wcp_linear_pde} together with \cref{thm:robust_coverage}) relies strictly on linearity and loses its coverage guarantees for nonlinear PDEs, where coverage depends on the PDE parameters (Appendix~\ref{app:nonlinear}). 

\section{Discussion}\label{sec:discussion}
Conformal prediction for time-dependent physical phenomena is often constrained by non-exchangeable data. In this work, we investigated whether coverage guarantees can be maintained beyond the exchangeability assumption and show that this depends strongly on the problem representation. On function spaces, measures are typically mutually singular, making coverage guarantees unattainable. On discretized domains, however, we derived how weighted CP can be applied to linear PDEs to obtain coverage guarantees. We further show that these guarantees can be preserved for real-world measurements, and that CP should be restricted to a coarse, low-dimensional grid to provide meaningful guarantees. A natural next step is to extend these results to specific classes of (non)linear PDEs.

\begin{ack}
The authors gratefully acknowledge the computational and data resources provided by the Leibniz Supercomputing Centre (LRZ). JS acknowledges funding from the Munich Center for Machine Learning (MCML), as well as the support he received from the METEOR Research Training Group, which is sponsored by the German Research Foundation.
AS is supported by the DAAD programme Konrad Zuse Schools of Excellence in Artificial Intelligence, sponsored by the Federal Ministry of Research, Technology and Space.
The authors declare no competing interests.
\end{ack}

{\small
\bibliography{Conformal-Operators}
}

\newpage
\appendix
\crefalias{section}{appendix}
\crefalias{subsection}{appendix}

\numberwithin{figure}{section}
\numberwithin{table}{section}

\numberwithin{definition}{section}
\numberwithin{theorem}{section}
\numberwithin{proposition}{section}
\numberwithin{lemma}{section}
\numberwithin{corollary}{section}
\numberwithin{remark}{section}
\numberwithin{assumption}{section}
\section{LLM disclaimer}
Large language models (LLMs) were used to support the development and execution of this project. They contributed to improving the clarity and readability of the manuscript, suggesting alternative phrasings, helping with the code, offering feedback on mathematical arguments, and providing ideas throughout the research and coding process. All research contributions, results, and final formulations were manually verified.

\section{Licenses of external assets}\label{app:licenses}
All external assets used in this work are publicly available. The pulsed-thermography dataset of \citet{wei_pulsed_2023} is released under CC-BY~4.0 and hosted on Kaggle.\footnote{\url{https://www.kaggle.com/datasets/ziangwei/irtpvc}} The Lake Maggiore thermistor data of \citet{scheu_observation_2018} is archived on Zenodo under CC-BY~4.0.\footnote{\url{https://doi.org/10.5281/zenodo.1206188}} The FRAP dataset and reference code of \citet{wahlstrand_skarstrom_deepfrap_2021} are archived on Zenodo under CC-BY~4.0.\footnote{\url{https://doi.org/10.5281/zenodo.3874218}}

\section{Example: a Gaussian random field on a function space}\label{app:function_space_dist}

A probability distribution on a function space is very similar to a distribution on $\sR^p$. Where a draw from a multivariate Gaussian $\gN(\vmu,\mSigma)$ on $\sR^p$ produces a vector of $p$ numbers, a draw from a distribution on $\normltwo(\Omega)$ produces an entire function $u:\Omega\to\sR$. Each sample is a function over $\Omega$, and different draws give different functions.

The simplest and most common such distributions are \emph{Gaussian random fields} (GRFs), the infinite-dimensional analogue of multivariate Gaussians. Like a Gaussian on $\sR^p$, a GRF is determined by two ingredients: a mean function $\vm\in\normltwo(\Omega)$ playing the role of $\vmu$, and a covariance operator $\mC$ playing the role of $\mSigma$. The covariance operator $\mC:\normltwo(\Omega)\to\normltwo(\Omega)$ must be positive, symmetric, and trace class~\cite{hairer_introduction_2023}. The trace-class condition is what makes the infinite sum of variances finite, ensuring samples actually lie in $\normltwo(\Omega)$. We write $u\sim\gN(\vm,\mC)$.

The most useful way to picture a sample is via its \emph{Karhunen--Lo\`eve expansion}. If $(\mu_n,e_n)_{n\geq 1}$ are the eigenvalues and eigenfunctions of $\mC$ (so $\mC e_n = \mu_n e_n$), then a draw $u\sim\gN(\vzero,\mC)$ can be written as
\begin{equation}\label{eq:karhunen_loeve}
u(x) \;=\; \sum_{n=1}^\infty \sqrt{\mu_n}\,\xi_n\,e_n(x),
\qquad \xi_n \stackrel{\mathrm{iid}}{\sim} \gN(0,1),
\end{equation}
i.e., a random superposition of the basis functions $e_n$ with independent Gaussian coefficients whose variances $\mu_n$ are set by $\mC$. The decay rate of $\mu_n$ controls how rough or smooth the samples look: fast decay suppresses high-frequency $e_n$ and gives smooth samples, slow decay gives rough ones.

For \cref{thm:heat_eq_no_tv}, we take $\Omega=(0,1)$, $\vm = \vzero$, and $\mC = (\mI-\mLambda)^{-1}$ with $\mLambda$ the Laplace operator under homogeneous Dirichlet boundary conditions. The eigenfunctions $e_n(x)=\sqrt{2}\sin(n\pi x)$ are sine modes of increasing frequency, and the eigenvalues $\mu_n = 1/(1+(n\pi)^2)$ decay like $1/n^2$. The trace condition $\sum_n \mu_n < \infty$ holds, so $\gN(\vzero,\mC)$ is a well-defined GRF on $\normltwo(\Omega)$. Each sample is a continuous function on $(0,1)$ that vanishes at the endpoints (because all $e_n$ do), with most of its energy in the low-frequency modes.

\section{Validating local exchangeability}\label{app:lsci_assumption}

To have formal guarantees with the LSCI approach from~\cite{harris_locally_2025}, it is required that the model's \textit{residuals} $r_t$ are locally exchangeable. Considering this for the simplest setup of two time points $t$ and $t+\delta$ for some $\delta>0$, this means that it must hold that
\[\dTV \big(\begin{pmatrix} r_t \\ r_{t+\delta} \end{pmatrix},\begin{pmatrix} r_{t+\delta} \\ r_{t} \end{pmatrix}\big) \leq d(t,t+\delta)\]
for a pre-metric $d$ on the time domain.
Note, however, that since we do not have access to the laws of the random vectors above, we cannot reason about their TV distance.
Even though in \cref{prop:wcp_linear_pde} we derive the laws of the solutions $u_t$, it is not clear how to reason about the above left-hand side without further assumptions (like independence of the residuals over time, which is not plausible as the solution has a clear time-dependence in all non-stationary PDEs).

\section{Additional figures}

\begin{figure}[h]
    \centering
    \includegraphics[width=\columnwidth]{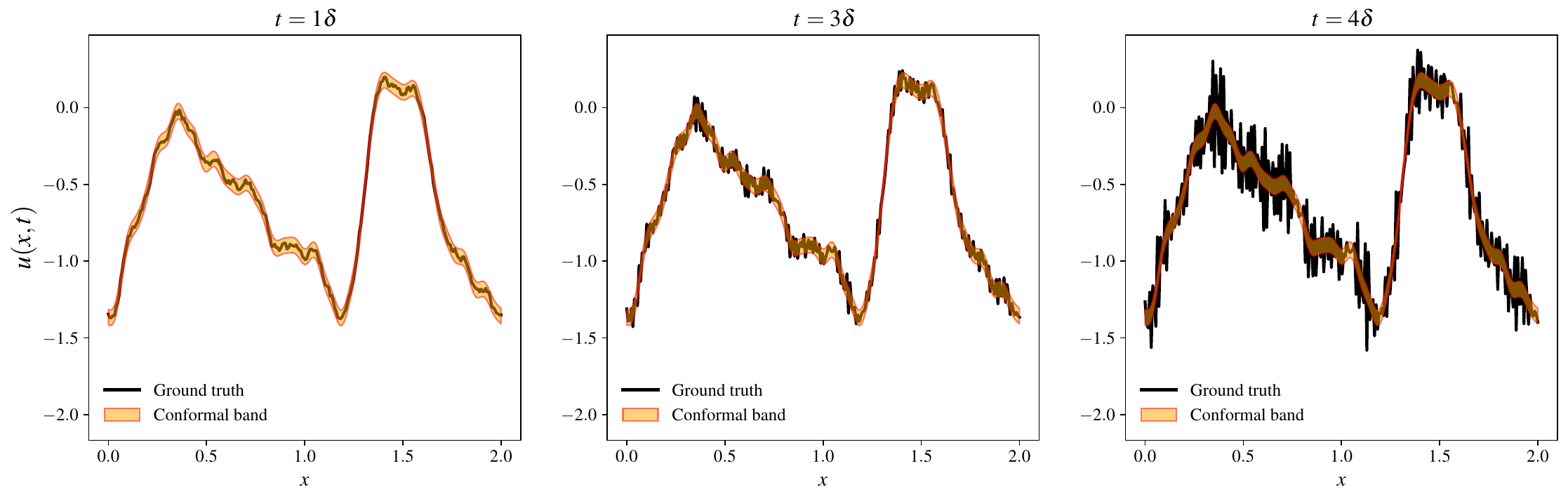}
    \caption{CP bands of the backward heat equation calibrated at $t=0$. Each panel shows one time point of the solution $u(x,t)$ and CP bands over the spatial domain $x$. While at time $t=\delta$, the solution is properly covered, coverage is lost at later time steps.}
    \label{fig:cp_timestep}
\end{figure}

\section{Weighted conformal prediction formula}\label{app:wcp_formula}

For completeness, we recall the explicit form of the weighted split conformal predictor of \citet{tibshirani_conformal_2019} that underlies our \emph{WCP Gauss} method (\cref{prop:wcp_linear_pde}).

Given a calibration set $\gD_{\mathrm{cal}}=\{(\vx_i,y_i)\}_{i=1}^n$, a fixed score function $S$, and unnormalized weights $w_i$ on $\gD_{\mathrm{cal}}\cup\{\vx_{\mathrm{test}}\}$ (e.g., \cref{eq:weighted-cp} or, in our case, \cref{eq:weights}), define the normalized weights
\begin{equation}\label{eq:wcp_normalized_weights}
\tilde w_i := \frac{w_i}{\sum_{j=1}^n w_j + w_{\mathrm{test}}},\ i=1,\dots,n,
\qquad
\tilde w_{n+1} := \frac{w_{\mathrm{test}}}{\sum_{j=1}^n w_j + w_{\mathrm{test}}},
\end{equation}
where $w_{\mathrm{test}}$ is the weight evaluated at $\vx_{\mathrm{test}}$. Writing $S_i:=S(\vx_i,y_i)$ for the calibration scores, the weighted split conformal prediction set at level $1-\alpha$ is
\begin{equation}\label{eq:wcp_set}
\gC^{\mathrm{WCP}}(\vx_{\mathrm{test}}) := \left\{ y\in\sR\ :\ S(\vx_{\mathrm{test}}, y) \le Q_{1-\alpha}\!\Big(\sum_{i=1}^n \tilde w_i\,\delta_{S_i} + \tilde w_{n+1}\,\delta_{+\infty}\Big) \right\},
\end{equation}
where $\delta_a$ denotes a point mass at $a$ and $Q_{1-\alpha}(\mu)$ the $(1-\alpha)$-quantile of the discrete measure $\mu$. The atom at $+\infty$ forces $\gC^{\mathrm{WCP}}=\sR$ whenever $\tilde w_{n+1} > \alpha$, i.e., whenever the test point is so unlike the calibration set that no finite quantile carries enough mass. This corresponds to the infinite bands reported as $n_\infty$ in our experiments. With weights set to the true density ratio between test and calibration distributions, the predictor satisfies $\sP\big(y_{\mathrm{test}}\in\gC^{\mathrm{WCP}}(\vx_{\mathrm{test}})\big)\ge 1-\alpha$~\citep{tibshirani_conformal_2019}. Standard split CP (\cref{sec:cp_background}) is recovered by setting $w_i\equiv 1$, which collapses all normalized weights to $1/(n+1)$.

\paragraph{Gaussian WCP algorithm}
Algorithm~\ref{alg:wcp_pde} below instantiates the WCP set in \cref{eq:wcp_set} using Gaussian density estimates at calibration and prediction times, with covariance estimated by the Ledoit--Wolf shrinkage estimator \citep{ledoit_well-conditioned_2004} for stability in the high-dimensional regime. We list the algorithm with different sets for calibration and test. In practice, it should be common to use the same trajectories at different time points for calibration and testing. We show both in our real-world tasks, where the pulsed-thermography and FRAP datasets use the same trajectories, but Lake Maggiore splits the data (at least it splits them into early-spring and late-spring).

\begin{algorithm}[H]
\DontPrintSemicolon
\textbf{Input:} calibration pairs $\{(\vu_i^{t_{\mathrm{cal}}},\vu_i^{t_{\mathrm{cal}}+\delta})\}_{i=1}^n$, test-time inputs $\{\tilde\vu_j^t\}_{j=1}^{m+1}$ at time $t\geq t_{\mathrm{cal}}$ (with $\tilde\vu_{m+1}^t$ the query), nonconformity score $S$, miscoverage level $\alpha$.\\
\textbf{Output:} prediction set $\gC^{\mathrm{WCP}}_{t+\delta}(\tilde\vu_{m+1}^t)$.\\[0.3em]
$\gD_{t_{\mathrm{cal}}} \gets \{\vu_i^{t_{\mathrm{cal}}}\}_{i=1}^n$,\ \  $\gD_t \gets \{\tilde\vu_j^t\}_{j=1}^{m}$\;
\For{$\tau\in\{t_{\mathrm{cal}},\,t\}$}{
  $\hat\vmu_\tau \gets \tfrac{1}{|\gD_\tau|}\sum_{\vu\in\gD_\tau}\vu$\tcp*{sample mean}
  $\hat\mSigma_\tau \gets \mathrm{LedoitWolf}(\gD_\tau)$\tcp*{Ledoit--Wolf shrinkage}
}
\For{$i=1,\dots,n$}{
  $w_i \gets \varphi(\vu_i^{t_{\mathrm{cal}}};\hat\vmu_t,\hat\mSigma_t)/\varphi(\vu_i^{t_{\mathrm{cal}}};\hat\vmu_{t_{\mathrm{cal}}},\hat\mSigma_{t_{\mathrm{cal}}})$\tcp*{importance weight}
  $S_i \gets S(\vu_i^{t_{\mathrm{cal}}},\vu_i^{t_{\mathrm{cal}}+\delta})$\tcp*{calibration score}
}
$w_{m+1} \gets \varphi(\tilde\vu_{m+1}^t;\hat\vmu_t,\hat\mSigma_t)/\varphi(\tilde\vu_{m+1}^t;\hat\vmu_{t_{\mathrm{cal}}},\hat\mSigma_{t_{\mathrm{cal}}})$\tcp*{test weight}
$\tilde w_i \gets w_i/(\sum_j w_j + w_{m+1})$, \ $\tilde w_{n+1} \gets w_{m+1}/(\sum_j w_j + w_{m+1})$\tcp*{normalize, cf.\ \cref{eq:wcp_normalized_weights}}
$\hat q_{1-\alpha} \gets Q_{1-\alpha}\!\big(\sum_{i=1}^n \tilde w_i\,\delta_{S_i} + \tilde w_{n+1}\,\delta_{+\infty}\big)$\;
\Return $\gC^{\mathrm{WCP}}_{t+\delta}(\tilde\vu_{m+1}^t) = \{\,\vy : S(\tilde\vu_{m+1}^t,\vy)\leq \hat q_{1-\alpha}\,\}$\;
\caption{WCP for time-dependent PDEs.}
\label{alg:wcp_pde}
\end{algorithm}

\section{Proofs}\label{app:proofs}
\subsection{Proof of \texorpdfstring{\cref{thm:heat_eq_no_tv}}{Theorem 4.1}}\label{app:proof_heat_eq_no_tv}
\tvtheorem*

\begin{proof}
  Notice that our whole setup is on a Hilbert space. We begin by showing that $(\mI - \mLambda)^{-1}$ is a well-defined covariance operator on $\normltwo(\Omega)$. For this, according to 
  \cite[Proposition 4.17]{hairer_introduction_2023}, the operator must be positive, symmetric and trace class. As both $\mI$ and $-\mLambda$ are positive and symmetric, so is their sum. As $-\mLambda$ is positive, it follows that $(\mI-\mLambda)$ is invertible, due to strictly positive eigenvalues. Further, the eigenvalues of $-\mLambda$ with Dirichlet boundary conditions are given by $\lambda_n = (n\pi)^2$, with corresponding eigenfunctions $e_n(x) = \sqrt{2}\sin(n\pi x)$, which form an orthonormal basis of $\normltwo(\Omega)$. Therefore, the eigenvalues of $(\mI - \mLambda)^{-1}$ are given by $\mu_n = 1/(1+(n\pi)^2)$ and as
  \[
  \sum_{n=1}^{\infty} \mu_n < \infty,
  \]
  we conclude that $(\mI - \mLambda)^{-1}$ is a trace class operator and thus defines a Gaussian measure on $\normltwo(\Omega)$.\\

  Now our proof will be based on the Feldman--Hájek theorem \cite[Theorem 2.23]{da_prato_stochastic_1992}, which gives a characterization of when two Gaussian measures on a Hilbert space are either equivalent or mutually singular. We will briefly state the whole chain of reasoning, and then provide the necessary details.

  We will show that our measure at all times is Gaussian. By the Feldman--Hájek theorem, two Gaussian measures $\gN(\vm_1, \mC_1)$ and $\gN(\vm_2, \mC_2)$ on a Hilbert space are either equivalent or mutually singular. A necessary condition for equivalence is that the Cameron--Martin spaces, as given by $\mC^{1/2}$, of the two measures are equal as sets \cite[Theorem 2.23]{da_prato_stochastic_1992}. Thus, if the ranges of the covariance operators $\mC_1^{1/2}$ and $\mC_2^{1/2}$ are not equal, then the measures are mutually singular and their TV distance is $1$.

  \paragraph{Calculating the covariance operators} Starting with a measure $\mu_0$ of the initial distribution, the heat equation induces a semigroup $S(t) = \exp(t\mLambda)$, which maps the initial condition $u_0$ to the solution at time $t$, i.e., $u(\cdot, t) = S(t)u_0$. Therefore, the measure $\mu_t$ of $u(\cdot, t)$ is induced by the pushforward of $\mu_0$ under $S(t)$, i.e., $\mu_t = S(t)_{\#}\mu_0$. As $S(t)$ is linear, $\mu_t$ is also a Gaussian measure with mean $\boldsymbol{0}$ and covariance operator
  \[\mC_t = S(t)(\mI - \mLambda)^{-1}S(t)^* = \exp(t\mLambda)(\mI - \mLambda)^{-1}\exp(t\mLambda),\]
  where $S(t)^*$ denotes the adjoint of $S(t)$ \citep[][Chap. 4.3]{hairer_introduction_2023}.
  As we have seen above, the eigenvalues of $(\mI - \mLambda)^{-1}$ are given by $\mu_n = 1/(1+(n\pi)^2)$, with corresponding eigenfunctions $e_n(x)$. Further, the eigenfunctions of $\mLambda$ are also given by $e_n(x)$, with corresponding eigenvalues $\lambda_n = -(n\pi)^2$. Lastly, by functional calculus, the eigenfunctions of $\exp(t\mLambda)$ are also given by $e_n(x)$ with corresponding eigenvalues $\nu_n = \exp(-t(n\pi)^2)$. With this, we can compute 
  \begin{align*}
  \mC_t \ve_n &= \exp(t\mLambda)(\mI - \mLambda)^{-1}\exp(t\mLambda) \ve_n\\
  &= (\exp(-t(n\pi)^2))\left(\frac{1}{1+(n\pi)^2}\right)(\exp(-t(n\pi)^2)) \ve_n = \frac{\exp(-2t(n\pi)^2)}{1+(n\pi)^2} \ve_n.
  \end{align*}
  Thus, the eigenvalues of $\mC_t$ are given by $\lambda_n(t) = \nu_n^2 \mu_n = \exp(-2t(n\pi)^2)/(1+(n\pi)^2)$, with corresponding eigenfunctions $e_n(x)$.

  \paragraph{Calculating the Cameron--Martin spaces}
  The functions $e_n(x)$ form an orthonormal basis of $\normltwo(\Omega)$, so we can express every element $f \in \normltwo(\Omega)$ as
  \[f = \sum_{n=1}^{\infty} c_n e_n, \quad \sum_{n=1}^{\infty} c_n^2 < \infty.\]
  The Cameron--Martin space $H_t$ of $\gP_t$ is given by the range of $\mC_t^{1/2}$, which is given by
  \[\textrm{Ran}(\mC_t^{1/2}) = \left\{ \mC_t^{1/2} f \mid f \in \normltwo(\Omega) \right\} = \left\{ \sum_{n=1}^{\infty} \sqrt{\lambda_n(t)} c_n e_n \mid f \in \normltwo(\Omega) \right\}.\]
  Therefore, $g \in H_t$ if and only if $g$ can be expressed as
  \[g = \sum_{n=1}^{\infty} d_n e_n, \quad \sum_{n=1}^{\infty} \frac{d_n^2}{\lambda_n(t)} < \infty.\]
  Inserting the expression for $\lambda_n(t)$, we see that $g \in H_t$ if and only if
  \[\sum_{n=1}^{\infty} d_n^2 \frac{1+(n\pi)^2}{\exp(-2t(n\pi)^2)} < \infty.\]

  \paragraph{Showing that the Cameron--Martin spaces are not equal}
  Now it is easy to see that for any $t \geq 0$, $\delta > 0$, the Cameron--Martin spaces $H_t$ and $H_{t+\delta}$ are not equal. For example, the function
  \[h(x) = \sum_{n=1}^{\infty} \exp(-(t+\delta)(n\pi)^2) e_n(x)\]
  is an element of $H_t$, as
  \[\sum_{n=1}^{\infty} (\exp(-(t+\delta)(n\pi)^2))^2 \frac{1+(n\pi)^2}{\exp(-2t(n\pi)^2)} = \sum_{n=1}^{\infty} (1+(n\pi)^2) \exp(-2\delta(n\pi)^2) < \infty,\]
  but it is not an element of $H_{t+\delta}$, as
  \[\sum_{n=1}^{\infty} (\exp(-(t+\delta)(n\pi)^2))^2 \frac{1+(n\pi)^2}{\exp(-2(t+\delta)(n\pi)^2)} = \sum_{n=1}^{\infty} (1+(n\pi)^2) = \infty.\]
  
  Therefore, by the Feldman--Hájek theorem, the measures $\gP_t$ and $\gP_{t+\delta}$ are mutually singular, and their TV distance is $1$.
\end{proof}

\subsection{Proof of \texorpdfstring{\cref{prop:wcp_linear_pde}}{Proposition 4.2}}\label{app:proof_wcp_linear_pde}
\marginalCoverageProp*
\begin{proof}
By assumption, the numerical time--$\delta$ map is affine: there exist a matrix $\mA\in\R^{p\times p}$ and vector $\vb\in\R^p$ such that
\[
  \vu^{t+\delta}=\mA\vu^t+\vb .
\]
Iterating yields for $k\in\sN$,
\[
  \vu^{t_{\mathrm{cal}}+k\delta}
  =\mA^k \vu^{t_{\mathrm{cal}}} + \sum_{j=0}^{k-1}\mA^j\vb ,
\]
and in particular each $\vu^t$ is an affine function of $\vu^0$.

Since $\vu^0\sim\mathcal{N}(\vmu_0,\mSigma_0)$ and the multivariate normal family is closed under affine transformations, we have for every $t\ge 0$
\[
  \vu^t \sim \mathcal{N}(\vmu_t,\mSigma_t)
\]
for some mean $\vmu_t$ and covariance $\mSigma_t$ (e.g., $\vmu_t=\mM_t\vmu_0+\vc_t$ and $\mSigma_t=\mM_t\mSigma_0\mM_t^\top$ for the corresponding affine representation $\vu^t=\mM_t\vu^0+\vc_t$).
Hence the calibration and test laws admit Lebesgue densities $p_{t_{\mathrm{cal}}}$ and $p_t$ on $\R^p$ and the likelihood ratio
\begin{equation}\label{eq:weights}
  r(\vx)\;:=\;\frac{p_t(\vx)}{p_{t_{\mathrm{cal}}}(\vx)}
\end{equation}
is available in closed form. In particular, weighting by \cref{eq:weights} corresponds to importance weights proportional to $p_t/p_{t_{\mathrm{cal}}}$.

Now apply likelihood-weighted split conformal prediction (weighted CP) with these weights to any fixed score function.
The standard validity theorem for weighted split CP (under covariate shift with known likelihood ratio) implies that the resulting predictor $\mathcal{C}^{\mathrm{WCP}}_{t+\delta}$ satisfies
\[
  \sP\big(\vu^{t+\delta}\in \mathcal{C}^{\mathrm{WCP}}_{t+\delta}\big)\;\ge\;1-\alpha
\]
for each target time index $t\ge t_{\mathrm{cal}}$. This proves the claim.
\end{proof}

\subsection{Proof of \texorpdfstring{\cref{thm:robust_coverage}}{Proposition 4.6}}\label{app:proof:robust_coverage}
\robustCoverage*
\begin{proof}
As this proof is purely statistical, we will use random variables $X_i$ instead of PDE-solutions $u_i^t$ to ease the notation with more familiar symbols. In particular, we will use $Z_i^{(0)}$ and $Z_i^{(1)}$ for the estimation samples of the calibration and test distribution, respectively.\\
The proof extends WCP \citep{tibshirani_conformal_2019} to use worst case weights over a confidence region. Specifically, we know that we can construct a valid conformal set with the true parameters $\theta^\star=(\vmu_0,\mSigma_0,\vmu_1,\mSigma_1)$ using WCP. Now we construct a $(1-\eta)$ confidence region $\Theta$ for the unknown true parameters $\theta^\star$ using the covariate-only samples. The robust construction
then takes a worst-case weighted conformal threshold (meaning the threshold that yields the largest conformal set over $\Theta$) computed using the
WCP weights
\[
p_{i, \theta}(x)=\frac{w_\theta(X_i)}{\sum_{j=1}^n w_\theta(X_j)+w_\theta(x)},\ i=1,\dots,n,
\qquad
p_{n+1, \theta}(x)=\frac{w_\theta(x)}{\sum_{j=1}^n w_\theta(X_j)+w_\theta(x)},
\]
where $w_\theta=dP_1^\theta/dP_0^\theta$ is the covariate likelihood ratio of normal distributions with parameters $\theta$, $X_i$ are calibration samples and $x$ is the test sample. On the event $\{\theta^\star\in\Theta\}$,
the robust set contains the oracle WCP set. The stated coverage then follows by a union bound.

Throughout, $\varphi(\cdot;\vmu,\mSigma)$ denotes the $p$-variate Gaussian density with mean $\vmu$ and covariance
$\mSigma\succ0$.

\paragraph{Getting the \texorpdfstring{$\boldsymbol{(1-\eta)}$}{1-eta} confidence region for \texorpdfstring{$\boldsymbol{\theta^\star}$}{theta-star}}
From the covariate-only samples $\{Z^{(k)}_j\}_{j=1}^{m_k}$ for $k\in\{0,1\}$, define the sample mean and sample
covariance
\[
\widehat\vmu_k := \frac{1}{m_k}\sum_{j=1}^{m_k} Z^{(k)}_j,
\qquad
\widehat\mSigma_k := \frac{1}{m_k-1}\sum_{j=1}^{m_k}
(Z^{(k)}_j-\widehat\vmu_k)(Z^{(k)}_j-\widehat\vmu_k)^\top .
\]
Since $m_k>p$ and the data are Gaussian, $\widehat\mSigma_k$ is positive definite almost surely \citep[Thm. 3.1.4]{muirhead_aspects_1982}.

Fix numbers $\eta_1,\eta_2,\eta_3,\eta_4>0$ with
$\eta_1+\eta_2+\eta_3+\eta_4\le\eta$.

\emph{Mean confidence sets.}
By \citet[Cor. 5.3]{hardle_applied_2019}, for each $k\in\{0,1\}$, the Hotelling $T^2$ statistic
\[
T_k^2 := m_k(\widehat\vmu_k-\vmu_k)^\top \widehat\mSigma_k^{-1}(\widehat\vmu_k-\vmu_k)
\]
satisfies
\[
\frac{m_k-p}{p(m_k-1)}T_k^2 \sim F_{p,m_k-p}.
\]
Let $c_k^{(\vmu)}$ be the $(1-\eta_{3+k})$ quantile of
$\frac{p(m_k-1)}{m_k-p}F_{p,m_k-p}$ and define
\[
\mathcal M_k :=
\left\{
\vmu\in\mathbb R^p :
m_k(\widehat\vmu_k-\vmu)^\top \widehat\mSigma_k^{-1}(\widehat\vmu_k-\vmu)
\le c_k^{(\vmu)}
\right\}.
\]
Then $\mathcal M_k$ is compact and
$\mathbb P(\vmu_k\in\mathcal M_k)\ge 1-\eta_{3+k}$. Notice that  $\mathcal M_k$ is an ellipsoid around the sample mean.

\emph{Covariance confidence sets.}
For the sample covariance matrix, we know $W_k=(m_k-1)\widehat\mSigma_k\sim\mathrm{Wishart}_p(\mSigma_k,m_k-1)$ \citep[Thm. 5.7 (c)]{hardle_applied_2019}.
The eigenvalues of
$\mSigma_k^{-1/2}\widehat\mSigma_k\mSigma_k^{-1/2}$
have a distribution independent of $\mSigma_k$ \citep[Thm. 5.5]{hardle_applied_2019}.
Choose constants $0<a_k<b_k<\infty$ such that
\[
\mathbb P\!\left(
a_k I \preceq
\mSigma_k^{-1/2}\widehat\mSigma_k\mSigma_k^{-1/2}
\preceq b_k I
\right)
\ge 1-\eta_{k+1}. 
\]
This is well-defined, as the eigenvalues of $\mSigma_k^{-1/2}\widehat\mSigma_k\mSigma_k^{-1/2}$ are almost surely strictly positive and finite. Hence $L_k:=\lambda_{\text{min}}(\mSigma_k^{-1/2}\widehat\mSigma_k\mSigma_k^{-1/2})$ and $U_k := \lambda_{\text{max}}(\mSigma_k^{-1/2}\widehat\mSigma_k\mSigma_k^{-1/2})$ are well-defined random variables with support on a subset of $(0,\infty)$.\\
Define
\[
\mathcal S_k :=
\left\{
\mSigma\in\mathbb S_{++}^p :
\frac{1}{b_k}\widehat\mSigma_k
\preceq \mSigma
\preceq
\frac{1}{a_k}\widehat\mSigma_k
\right\}.
\]
Each set $\mathcal S_k$ is compact. 
Indeed, $\mathcal S_k$ is bounded since, for any $\mSigma\in\mathcal S_k$, its spectral norm satisfies
\[
\|\mSigma\|_2 = \lambda_{\max}(\mSigma),
\]
which is bounded above by the largest eigenvalue allowed by the defining inequality of $\mathcal S_k$.

Closedness follows as well. Let $(\mSigma_n)_{n\in\sN}\subset\mathcal S_k$ be a convergent sequence with limit $\mSigma_\infty$. 
By definition of $\mathcal S_k$, for every $n$ and every vector $z\in\mathbb R^p$,
\[
z^\top\!\left(\frac{1}{a_k}\widehat\mSigma_k - \mSigma_n\right)z \ge 0.
\]
By continuity of the quadratic form, passing to the limit yields
\[
z^\top\!\left(\frac{1}{a_k}\widehat\mSigma_k - \mSigma_\infty\right)z \ge 0
\quad\text{for all } z\in\mathbb R^p.
\]
An analogous argument applies to the lower bound in the definition of $\mathcal S_k$. Hence $\mSigma_\infty\in\mathcal S_k$, and $\mathcal S_k$ is closed.

Since $\mathcal S_k$ is both closed and bounded in the space of symmetric matrices, it is compact. 
Moreover, under the stated assumptions, the matrices in $\mathcal S_k$ are almost surely bounded away from singularity.

Define the joint confidence region
\[
\Theta := \mathcal M_0\times\mathcal S_0\times\mathcal M_1\times\mathcal S_1 .
\]
By a union bound,
\begin{equation}
\label{eq:theta_cover}
\mathbb P(\theta^\star\in\Theta)\ge 1-\eta .
\end{equation}
Let $E:=\{\theta^\star\in\Theta\}$.

\paragraph{Defining the weights and quantiles}
For $\theta=(\vmu_0,\mSigma_0,\vmu_1,\mSigma_1)\in\Theta$, define the Gaussian likelihood ratio
\[
w_\theta(x):=\frac{\varphi(x;\vmu_1,\mSigma_1)}{\varphi(x;\vmu_0,\mSigma_0)}.
\]
For a test covariate value $x\in\mathbb R^p$, define the associated WCP weights
\[
p_{i,\theta}(x)
:=
\frac{w_\theta(X_i)}{\sum_{j=1}^n w_\theta(X_j)+w_\theta(x)},
\quad i=1,\dots,n,
\qquad
p_{n+1,\theta}(x)
:=
\frac{w_\theta(x)}{\sum_{j=1}^n w_\theta(X_j)+w_\theta(x)}.
\]
Then $p_{i,\theta}(x)\ge 0$ and $\sum_{i=1}^{n+1}p_{i,\theta}(x)=1$.

Let $\theta^\star$ denote the true parameters and $w^\star:=w_{\theta^\star}$.
The \emph{oracle} WCP set is the WCP set obtained by using $w^\star$ in the weights above, and we denote it by
$\widehat C_{\theta^\star}$ below.

Fix $x\in\mathbb R^p$ and a candidate label $y\in\mathbb R$. Let
$S_i$, $i=1,\dots,n$, be the nonconformity scores for the sample $(X_i,Y_i)$.

For each $\theta\in\Theta$ and test point $(x,y)$ with score $S_{n+1}$, define the discrete probability measure on $[0,\infty]$,
\begin{equation}
\label{eq:nu_theta}
\widehat \nu_{\theta}^{(x,y)}
:=
\sum_{i=1}^n p_{i,\theta}(x)\,\delta_{S_i} + p_{n+1,\theta}(x)\,\delta_{\infty},
\end{equation}
and define the corresponding CDF on $\mathbb R$ by
\[
F_\theta(t;x,y)
:=
\widehat \nu_{\theta}^{(x,y)}\bigl((-\infty,t]\bigr)
=
\sum_{i=1}^n p_{i,\theta}(x)\,\mathbf 1\{S_i\le t\},
\qquad t\in\mathbb R,
\]
since the atom at $\infty$ contributes no mass to $(-\infty,t]$ for finite $t$.

Define the $(1-\alpha)$ weighted quantile as
\begin{equation}
\label{eq:q_theta_def}
q_\theta(x,y)
:=
\inf\Bigl\{t\in\mathbb R:\ F_\theta(t;x,y)\ge 1-\alpha\Bigr\}.
\end{equation}
Equivalently, $q_\theta(x,y)$ is the quantile of level $1-\alpha$ under the discrete law
$\widehat \nu_{\theta}^{(x,y)}$ in \cref{eq:nu_theta}.

\paragraph{Robust threshold and robust prediction set}
Define the robust threshold as the worst-case quantile over $\Theta$:
\begin{equation}
\label{eq:q_rob}
q_{\mathrm{rob}}(x,y):=\sup_{\theta\in\Theta} q_\theta(x,y).
\end{equation}
Define the robust WCP prediction set by
\begin{equation}
\label{eq:C_rob_def}
\widehat C_{\mathrm{rob}}(x)
:=
\Bigl\{y\in\mathbb R:\ S_{n+1}\le q_{\mathrm{rob}}(x,y)\Bigr\}.
\end{equation}

Also, define for any fixed $\theta\in\Theta$, the WCP set
\[
\widehat C_{\theta}(x)
:=
\Bigl\{y\in\mathbb R:\ S_{n+1}\le q_{\theta}(x,y)\Bigr\}.
\]
In particular, $\widehat C_{\theta^\star}$ is the oracle WCP set.

On the event $E=\{\theta^\star\in\Theta\}$, we have for all $(x,y)$,
\[
q_{\mathrm{rob}}(x,y)=\sup_{\theta\in\Theta} q_\theta(x,y)\ \ge\ q_{\theta^\star}(x,y),
\]
and therefore, for all $x$,
\begin{equation}
\label{eq:set_incl}
\widehat C_{\mathrm{rob}}(x)\supseteq \widehat C_{\theta^\star}(x)
\qquad\text{on }E.
\end{equation}

\paragraph{Robust coverage}
Let $F:=\{Y_{n+1}\notin \widehat C_{\mathrm{rob}}(X_{n+1})\}$. By \cref{eq:set_incl},
\[
F\cap E
\subseteq
\{Y_{n+1}\notin \widehat C_{\theta^\star}(X_{n+1})\}\cap E.
\]
Hence,
\[
\mathbb P(F)
\le
\mathbb P\!\left(Y_{n+1}\notin \widehat C_{\theta^\star}(X_{n+1})\right)
+ \mathbb P(E^c).
\]
By the oracle WCP guarantee (i.e., the covariate-shift WCP coverage result applied with the true likelihood
ratio $w^\star=w_{\theta^\star}$),
\[
\mathbb P\!\left(Y_{n+1}\in \widehat C_{\theta^\star}(X_{n+1})\right)\ge 1-\alpha,
\]
so $\mathbb P(Y_{n+1}\notin \widehat C_{\theta^\star}(X_{n+1}))\le\alpha$.
Also, by \cref{eq:theta_cover}, $\mathbb P(E^c)\le\eta$.
Therefore,
\[
\mathbb P\!\left(Y_{n+1}\in \widehat C_{\mathrm{rob}}(X_{n+1})\right)
\ge 1-\eta-\alpha.
\]
\end{proof}

\subsection{Proof of \texorpdfstring{\cref{prop:cp_lifting}}{Proposition 4.7}}\label{app:proof_cp_lifting}
\certifiedLiftingTheorem*
\begin{proof}
Let $(\Omega,\gF,\gP)$ be an underlying probability space supporting the random initial condition $u_0:\Omega\to\gU$ and hence the (random) PDE solution at time $t$,
\[
  u(\cdot,t):\Omega\to\gU,
\]
as well as the random numerical state
\[
  \vu^t:\Omega\to\R^p.
\]
Assume $\mI_h:\R^p\to\gU$ is measurable and that the deterministic error bound assumed in \cref{prop:cp_lifting} holds for all $\omega\in\Omega$,
\begin{equation}\label{eq:pathwise_err}
  \big\|u(\cdot, t)(\omega)-\mI_h \vu^t(\omega)\big\|_{\gU}\le \varepsilon_{h,t}(u_0(\omega)).
\end{equation}
Let $C_h\subset\R^p$ be a (measurable) prediction set such that
\[
  \sP\big(\vu^t\in C_h\big)\ge 1-\alpha.
\]
Define the lifted set $C\subset\gU$ by \cref{eq:lifted_set}, i.e.
\[
  C := \Big\{ u(\cdot,t)\in\gU:\ \exists\,\vz\in C_h\ \text{s.t.}\ \|u(\cdot,t)-\mI_h\vz\|_{\gU}\le \varepsilon_{h,t}(u_0)\Big\},
\]
where the right-hand side is interpreted pointwise in $\omega$ through the random radius
$\varepsilon_{h,t}(u_0(\omega))$.

Consider the event $E:=\{\omega:\ \vu^t(\omega)\in C_h\}\in\gF$.
Fix $\omega\in E$. Then $\vz:=\vu^t(\omega)\in C_h$.
By the pathwise bound \cref{eq:pathwise_err},
\[
  \big\|u(\cdot,t)(\omega)-\mI_h \vz\big\|_{\gU}
  =\big\|u(\cdot,t)(\omega)-\mI_h \vu^t(\omega)\big\|_{\gU}
  \le \varepsilon_{h,t}(u_0(\omega)).
\]
Hence, by definition of $C$, we have $u(\cdot,t)(\omega)\in C$. Since $\omega\in E$ was arbitrary, this proves the event inclusion
\[
  \{\omega: \vu^t(\omega)\in C_h\}\subseteq \{\omega: u(\cdot,t)(\omega)\in C\}.
\]
Taking probabilities yields
\[
  \sP(u(\cdot,t)\in C)\ \ge\ \sP(\vu^t\in C_h)\ \ge\ 1-\alpha,
\]
which is the desired marginal coverage statement.
\end{proof}

\subsection{Proof of \texorpdfstring{\cref{thm:tv_general_drift}}{Theorem 4.5}}\label{app:proof_tv_general_drift}
\generalDriftTheorem*
\begin{proof}
We write $\vv:=c\,\vw$ for the drift vector. In addition to the TV distance, we derive the weights for likelihood-ratio weighting, which we use in the proof of \cref{thm:tv_gauss_same_cov}.
The drift PDE $\partial_t \vu^t=\vv$ has the solution
\[
  \vu^t=\vu^0+t\vv .
\]
Since $\vu^0\sim\gN(\vmu_0,\mSigma_0)$, we have for every $t\ge 0$,
\[
  \vu^t \sim \gN(\vm_t,\mSigma_0), \qquad \vm_t:=\vmu_0+t\vv,
\]
i.e., the drift shifts the mean and leaves the covariance unchanged.

\paragraph{Reduction to TV distance on inputs}
For $m \in \{1,2\}$, each pair $(\vu_i^{t_m}, \vu_i^{t_m+\delta})$ can be written as
\[
(\vu_i^{t_m}, \vu_i^{t_m+\delta})
=
\bigl(\vu_i^{t_m}, \gG_\delta(\vu_i^{t_m})\bigr)
\]
for some solution operator $\gG_\delta$. Consequently, the joint distribution of
$(\vu_i^{t_m}, \vu_i^{t_m+\delta})$ is supported on a $p$-dimensional submanifold of
$\sR^{2p}$. This is easiest to see in the case $p=1$, where the joint distribution is
concentrated on the graph of $\gG_\delta(x)$ in the $x$--$y$ plane.

The only difference between the joint distributions of
$(\vu_i^{t_1}, \vu_i^{t_1+\delta})$ and $(\vu_i^{t_2}, \vu_i^{t_2+\delta})$ lies in how
probability mass is distributed along this graph. In particular, any measurable set
either has probability zero under both joint distributions (if it does not intersect
the graph), or its probability is entirely determined by the corresponding subset of
$x$-values.

Define
\[
\bm{\xi}_m := \mathcal{L}(\vu_i^{t_m})
\quad\text{and}\quad
\bm{\nu}_m := \mathcal{L}\bigl(\vu_i^{t_m}, \vu_i^{t_m+\delta}\bigr),
\qquad m \in \{1,2\},
\]
where $\mathcal{L}$ denotes the law of a random variable, and define the mapping
\[
T(x) := \bigl(x, \gG_\delta(x)\bigr).
\]
Then $\bm{\nu}_m = T_\# \bm{\xi}_m$. Since $T$ is injective, with inverse given by
projection onto the first component restricted to $\mathrm{graph}(\gG_\delta)$, it
follows that
\[
\|\bm{\nu}_1 - \bm{\nu}_2\|_{\mathrm{TV}}
=
\|\bm{\xi}_1 - \bm{\xi}_2\|_{\mathrm{TV}}.
\]

From here on, we will use $\hat\vz := (\vu^{t_1}_1, \dots, \vu^{t_1}_n, \vu^{t_2}_{n+1})$ and $\hat\vz^i$ for $\hat\vz$ when the $i$th and last entry are swapped.
\paragraph{Exact TV distance for equal-covariance Gaussians}
As each $\vu_i^t$ is multivariate normal and independent of the others, the joint distributions of $\hat\vz$ and $\hat\vz^i$ are again multivariate normal with the same block-diagonal covariance $\mI_{n+1}\otimes\mSigma_0\in\sR^{p(n+1)\times p(n+1)}$.
The mean of $\hat\vz$ is $\bar\vmu:=(\vm_{t_1},\dots,\vm_{t_1},\vm_{t_2})\in\sR^{p(n+1)}$, consisting of $n$ blocks $\vm_{t_1}$ followed by one block $\vm_{t_2}$. The mean $\bar\vmu^i$ of $\hat\vz^i$ equals $\bar\vmu$ with the $i$th and last block swapped.

TV distance is invariant under bijective measurable maps, so we may apply the whitening map $\vx\mapsto(\mI_{n+1}\otimes\mSigma_0^{-1/2})\vx$ to both distributions, which turns the covariance into $\mI_{p(n+1)}$.
For identity covariances we use the following identity, which is used in various papers (e.g., \citet{he_information-theoretic_2025, karageorgiou_limit_2025}) and attributed to \citet{barsov_estimates_1987} (as we cannot find access to the original source we use the citation directly from other papers that used the identity). 
\[
\dTV(\vz, \vz^i) = \dTV(\hat\vz, \hat\vz^i) = \dTV(\mathcal{N}(\vmu_{1}, \mI_{p(n+1)}), \mathcal{N}(\vmu_{2}, \mI_{p(n+1)})) = 2\Phi(\frac{\big\|\vmu_{1} - \vmu_{2}\big\|_{2}}{2}) - 1
\]

Together with the whitening, it gives
\[
  \dTV(\vz, \vz^i) = \dTV(\hat\vz, \hat\vz^i)
  =2\Phi\!\Big(\frac{\big\|(\mI_{n+1}\otimes\mSigma_0^{-1/2})(\bar\vmu - \bar\vmu^i)\big\|_2}{2}\Big)-1.
\]
The vector $\bar\vmu - \bar\vmu^i$ is zero except for the $i$th block, $\vm_{t_1}-\vm_{t_2}=(t_1-t_2)\vv$, and the last block, $(t_2-t_1)\vv$. Hence
\[
  \big\|(\mI_{n+1}\otimes\mSigma_0^{-1/2})(\bar\vmu - \bar\vmu^i)\big\|_2^2 = 2(t_1-t_2)^2\,\vv^\top\mSigma_0^{-1}\vv
\]
and
\[
  \dTV\!\left(\vz, \vz^i\right)
  =2\Phi\!\Big(\frac{\sqrt{2\,\vv^\top\mSigma_0^{-1}\vv}\,|t_1-t_2|}{2}\Big)-1.
\]
Moreover, since $1-\Phi(z)\le e^{-z^2/2}$, we get
\[
  1-\dTV(\vz, \vz^i)
  =2\Bigl(1-\Phi\!\bigl(\tfrac{\sqrt{2\,\vv^\top\mSigma_0^{-1}\vv}\,|t_1-t_2|}{2}\bigr)\Bigr)
  \le 2\exp\!\Bigl(-\frac{(t_1-t_2)^2\,\vv^\top\mSigma_0^{-1}\vv}{4}\Bigr),
\]
so $\dTV(\vz, \vz^i)\to 1$ exponentially fast in $(t_1-t_2)^2\,\vv^\top\mSigma_0^{-1}\vv$. With $\vv=c\,\vw$, we have $\vv^\top\mSigma_0^{-1}\vv=c^2p_{\mathrm{eff}}$, which gives \cref{thm:tv_general_drift}.

\paragraph{Likelihood ratio at the centers}
Let $p_{t}$ denote the density of $\gN(\vm_t,\mSigma_0)$ and $\Delta:=\vm_{t_2}-\vm_{t_1}=(t_2-t_1)\vv$.
The Gaussian likelihood ratio is
\[
  r(\vx):=\frac{p_{t_2}(\vx)}{p_{t_1}(\vx)}
  =\exp\!\Big(\Delta^\top \mSigma_0^{-1}\vx-\frac12\big(\vm_{t_2}^\top\mSigma_0^{-1}\vm_{t_2}-\vm_{t_1}^\top\mSigma_0^{-1}\vm_{t_1}\big)\Big).
\]
Evaluating at $\vx_{\mathrm{test}}:=\vm_{t_2}$ and $\vx_{\mathrm{cal}}:=\vm_{t_1}$ gives
\[
  \log r(\vm_{t_2})=\tfrac12\Delta^\top\mSigma_0^{-1}\Delta,\qquad \log r(\vm_{t_1})=-\tfrac12\Delta^\top\mSigma_0^{-1}\Delta.
\]
Since $\Delta^\top\mSigma_0^{-1}\Delta=(t_1-t_2)^2\,\vv^\top\mSigma_0^{-1}\vv=:\kappa$, we obtain
\[
  r(\vx_{\mathrm{test}})=e^{\kappa/2},
  \qquad
  r(\vx_{\mathrm{cal}})=e^{-\kappa/2}.
\]

\paragraph{Normalized weights}
If the calibration set contains $n$ points with ratio $r(\vx_{\mathrm{cal}})$ and we normalize over these $n$ points plus the test point, then
\[
  w_{\mathrm{test}}
  =\frac{r(\vx_{\mathrm{test}})}{nr(\vx_{\mathrm{cal}})+r(\vx_{\mathrm{test}})}
  =\frac{e^{\kappa}}{n+e^{\kappa}}
  =1-O\!\big(n e^{-\kappa}\big),
\]
and for any such calibration point,
\[
  w_{\mathrm{cal}}
  =\frac{r(\vx_{\mathrm{cal}})}{nr(\vx_{\mathrm{cal}})+r(\vx_{\mathrm{test}})}
  =\frac{1}{n+e^{\kappa}}
  =O\!\big(e^{-\kappa}\big).
\]
\end{proof}

\subsection{Proof of \texorpdfstring{\cref{thm:tv_gauss_same_cov}}{Proposition 4.4}}\label{app:proof_tv_gauss_same_cov}
\blowupTheorem*
\begin{proof}
\cref{thm:tv_gauss_same_cov} is the special case $\vw=\mathbf{1}$ and $\mSigma_0=\mI_p$ of the proof in Appendix~\ref{app:proof_tv_general_drift}. Then $p_{\mathrm{eff}}=p$ and $\kappa=pc^2(t_1-t_2)^2$, which gives the TV distance, its exponential convergence in $p$, and the weights.
\end{proof}

\subsection{General TV distance bound}
\begin{proposition}[Extension of \cref{thm:tv_gauss_same_cov}]\label{app:prop:general_tv_distance}
Consider a linear partial differential equation discretized on a spatial grid with $p$ points. Let $\mathbf{u}^{t_1}, \mathbf{u}^{t_2} \in \mathbb{R}^p$ be the Gaussian numerical solutions $\mathcal{N}(\vmu_{t_1}, \mSigma_{t_1})$ and $\mathcal{N}(\vmu_{t_2}, \mSigma_{t_2})$ at two distinct time steps. 

Suppose there exist uniform spectral bounds $0 < \underline{\lambda} \le \overline{\lambda} < \infty$ such that for $t \in \{t_1, t_2\}$, the eigenvalues of the covariance matrices satisfy $\lambda(\mSigma_t) \in [\underline{\lambda}, \overline{\lambda}]$. Then the total variation distance between the laws of the numerical solutions satisfies
\[
\dTV(\mathbf{u}^{t_1}, \mathbf{u}^{t_2}) \le \frac{1}{2} \sqrt{ p \left( \frac{\overline{\lambda}}{\underline{\lambda}} - 1 + \log \frac{\overline{\lambda}}{\underline{\lambda}} \right) + \frac{\|\vmu_{t_1} - \vmu_{t_2}\|^2}{\underline{\lambda}} }.
\]
\end{proposition}
For a consistent discretization of a smooth Gaussian random field, $\underline{\lambda}\to 0$ as $p$ grows, so the uniform spectral bounds only hold for a fixed range of grid sizes.

\begin{proof}
The result follows by bounding the Kullback--Leibler (KL) divergence and applying Pinsker's inequality. Let $P = \mathcal{N}(\vmu_{t_1}, \mSigma_{t_1})$ and $Q = \mathcal{N}(\vmu_{t_2}, \mSigma_{t_2})$. By Pinsker's inequality, 
\[
\dTV(P, Q) \le \sqrt{\frac{1}{2} \operatorname{KL}(P \| Q)}.
\]
The KL divergence for $p$-dimensional Gaussian measures is well known as
\[
\operatorname{KL}(P\|Q) = \frac{1}{2} \left[ \operatorname{tr}(\mSigma_{t_2}^{-1}\mSigma_{t_1}) - p + \log \frac{\det \mSigma_{t_2}}{\det \mSigma_{t_1}} + (\vmu_{t_1} - \vmu_{t_2})^\top \mSigma_{t_2}^{-1} (\vmu_{t_1} - \vmu_{t_2}) \right].
\]
First, we bound the terms related to the covariance structure. Under the assumption that the eigenvalues of $\mSigma_{t_1}$ and $\mSigma_{t_2}$ are bounded by $\underline{\lambda}$ and $\overline{\lambda}$, the trace of the product satisfies
\[
\operatorname{tr}(\mSigma_{t_2}^{-1}\mSigma_{t_1}) \le p \cdot \lambda_{\max}(\mSigma_{t_2}^{-1}) \lambda_{\max}(\mSigma_{t_1}) \le p \frac{\overline{\lambda}}{\underline{\lambda}},
\]
where the first inequality follows from von Neumann's trace inequality as the matrices are symmetric and positive definite.\\
Similarly, the log-determinant ratio is bounded by the sum of the logs of the eigenvalue ratios:
\[
\log \frac{\det \mSigma_{t_2}}{\det \mSigma_{t_1}} = \sum_{i=1}^p \log \frac{\lambda_i(\mSigma_{t_2})}{\lambda_i(\mSigma_{t_1})} \le p \log \frac{\overline{\lambda}}{\underline{\lambda}}.
\]
Next, the quadratic form involving the mean shift is bounded using the spectral norm of the inverse covariance matrix:
\[
(\vmu_{t_1} - \vmu_{t_2})^\top \mSigma_{t_2}^{-1} (\vmu_{t_1} - \vmu_{t_2}) \le \lambda_{\max}(\mSigma_{t_2}^{-1}) \|\vmu_{t_1} - \vmu_{t_2}\|^2 \le \frac{\|\vmu_{t_1} - \vmu_{t_2}\|^2}{\underline{\lambda}}.
\]
Substituting these bounds into the KL divergence expression, we obtain
\[
\operatorname{KL}(P\|Q) \le \frac{1}{2} \left[ p \left( \frac{\overline{\lambda}}{\underline{\lambda}} - 1 + \log \frac{\overline{\lambda}}{\underline{\lambda}} \right) + \frac{\|\vmu_{t_1} - \vmu_{t_2}\|^2}{\underline{\lambda}} \right].
\]
Applying the square root from Pinsker's inequality yields the stated bound.
\end{proof}

\section{Experimental details}
We tracked our results with Weights \& Biases~\citep{biewald_experiment_2020}.

\begin{figure}[ht]
    \centering
    \includegraphics[width=0.99\textwidth]{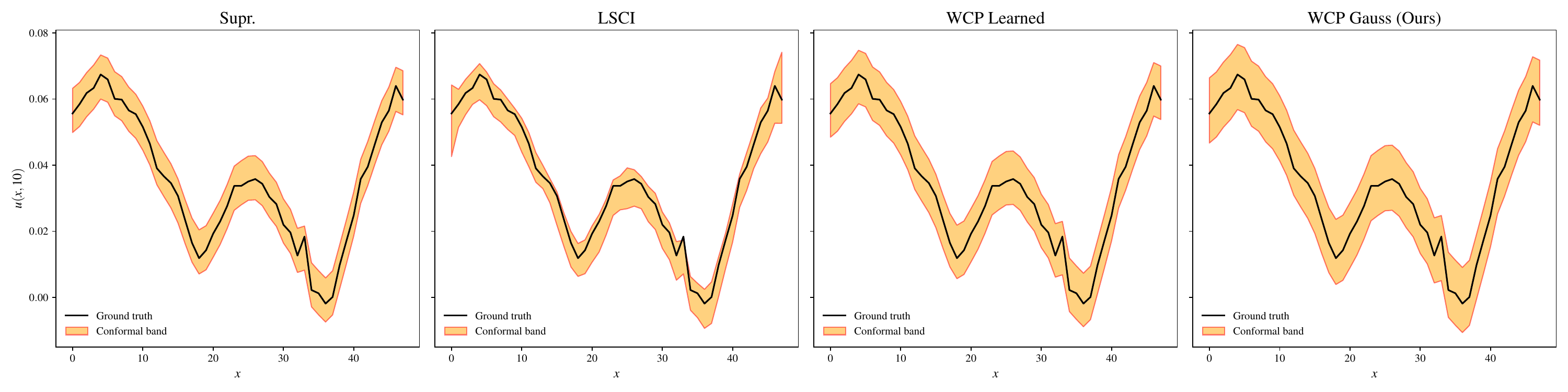}
    \caption{Exemplary coverage plots for the methods (from left to right) Supr., LSCI, WCP Learned, and WCP Gauss (Ours) for a sample at time step $10$ with parameters $a=-0.0075, b=0, c=0.5$.}
    \label{fig:conf_bands}
\end{figure}
\subsection{Methods}
\paragraph{Supr.}
The Supr. method follows the suggestion by \citet{diquigiovanni_conformal_2022}. We calculate the scores at calibration time as the maximum over the spatial dimension $x$. Then, at test time the respective quantile of these scores is added to each spatial location of the test prediction to generate the conformal interval. No further fine-tuning is necessary.
\paragraph{LSCI}
The LSCI method is difficult to fine-tune. In particular, we have four different parameters: $\lambda$, the embedding dimension, the metric, and the number of samples to use for the conformal bands. As the computational burden is high for this method, we opted for a mid-range setup, and chose (following the original work) $\lambda=3$, embedding dimension $20$, the supremum norm, and varied the number of samples between $500$ and $1000$, depending on what gave the closest to the target coverage for the first time step. This fine-tuning would not be feasible in practice, but as it favors the baseline we compare against, it does not harm the validity of our method.
\paragraph{DtACI}
To showcase adaptive methods, we use Dynamically-Tuned Adaptive Conformal Inference (DtACI) by \citet{gibbs_conformal_2024}. They dynamically adjust their $\alpha$ value according to previous miscoverage to satisfy the original $1-\alpha$ guarantee. In comparison to previous methods \citep[e.g.,][]{gibbs_adaptive_2021}, this method does not need manual tuning of the correction parameter, and the paper suggests calculation for some default parameters which we followed. For the rest of the free parameters, we followed the examples from their paper and chose 8 agents with $\gamma$ one of $[0.001,0.002,0.004,0.008,0.016,0.032,0.064,0.128]$ and a window size of 5 (due to the short time frame we observe).
\paragraph{WCP Learned}
The WCP Learned method is a WCP method according to the original proposal of \citet{tibshirani_conformal_2019}. There, they suggest estimating the (unknown) weights by the probabilistic output of a classifier. We experimented with multiple versions of LogisticRegression, Random Forests, and Support Vector Machines. We found gradient boosting to yield the best results in our experiments (n\_estimators=50, max\_depth=3, learning\_rate=0.1). There might be better models, e.g., certain deep learning models could model the underlying distributions better. However, the quality of a model depends on the data, i.e., a working model on one dataset might undercover on another. Also, as the models need to be trained at every prediction step, large deep learning models are practically infeasible.
\paragraph{WCP Gauss}
Our proposed method is also based on \citet{tibshirani_conformal_2019}. However, instead of estimating the weights, we use the knowledge of the underlying distribution to calculate them precisely. The mean and covariance are always estimated from the data using the average for the mean and Ledoit--Wolf \citep{ledoit_well-conditioned_2004} for the covariance.
\subsection{Data generation}\label{app:data_generation}
We generated multiple synthetic datasets from a general second-order formulation for periodic PDEs, each discretized with finite-difference schemes with exact exponential matrix evolution. Initial conditions were sampled from Gaussian processes with covariance $625( -\partial_x^2 + 25 I)^{-2}$ to provide smooth but nontrivial trajectories. Each dataset is stored as a compressed \texttt{.npz} file containing $(n_{\text{samples}}, n_t+1, n_x)$ trajectories together with grid and metadata.
The PDE looks as follows
\begin{alignat*}{2}
u_t(x,t) = au(x,t)_{xx} +bu(x,t)_{x} + cu(x,t)&, &\quad & x \in (0,2\pi),\ t \geq 0, \\
u(x,0)  = u_0(x)&, &\quad & x \in (0,2\pi),
\end{alignat*}
with periodic boundary conditions, and where $a, b, c\in\R$ are tunable parameters. We sampled data from all combinations of $a\in\{-0.005, -0.0075, -0.01\}$, $b\in\{-0.5, 0, 0.5\}$, and $c\in\{-0.5, 0, 0.5\}$. Note that we focused on negative values for $a$, as positive values tend to smooth the trajectory over time which leads to trivial coverage even for na\"ive CP. We sampled $21$ time steps (incl. initial condition) between $0$ and~$1$ (for the recalibration task, we extended the time to $2$, as otherwise all methods cover too easily) and sampled $5000$ trajectories for training, $500$ for validation, $5000$ for calibration, and $1000$ for testing, and use a spatial resolution of $n_x=48$. As surrogate, we train a 1D Fourier Neural Operator \citep{li_fourier_2021} (3 layers, 12 Fourier modes, 32 hidden channels) to predict one time step ahead, using AdamW (learning rate $10^{-3}$, cosine schedule) on the MSE loss for 5 epochs.

\subsection{Ablation study}\label{app:ablation_study}
\begin{figure}[ht]
    \centering
    \includegraphics[width=0.82\textwidth]{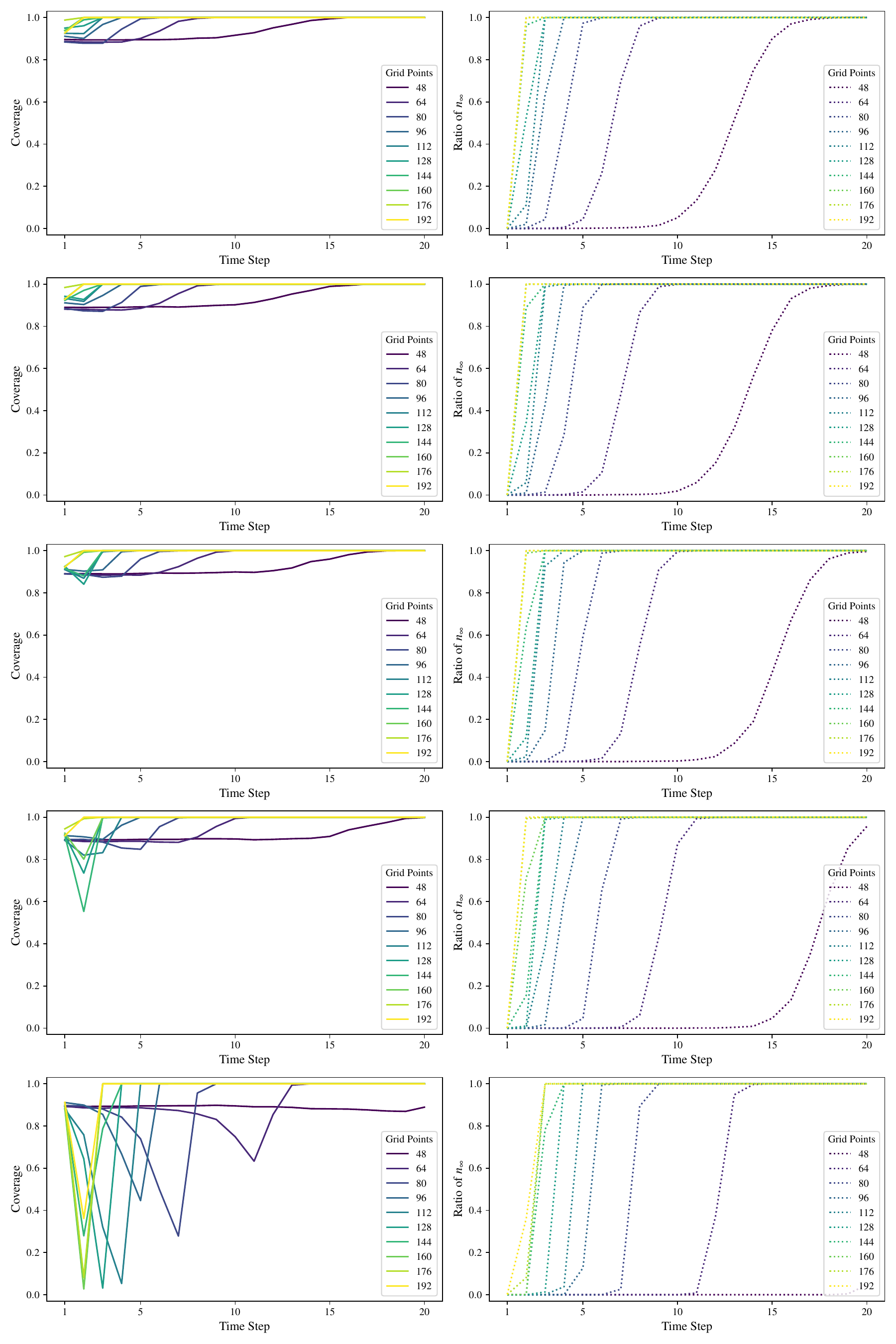}
    \caption{Ablation study for method stability. We use data from the synthetic task (no recalibration) with parameters $a=-0.0075,b=0,c=0$. From top to bottom, the rows correspond to 5000, 4000, 3000, 2000, and 1000 calibration samples. Each plot reports coverage (left) and the fraction of infinite conformal bands (right) over time, where color corresponds to grid granularity. We see that higher granularity leads to earlier onset of infinite bands, which aligns with \cref{thm:tv_gauss_same_cov}. When the number of samples is insufficient relative to the dimensionality, covariance estimation degrades and undercoverage can occur.}
    \label{fig:ablation_covariance}
\end{figure}

The critical part of the WCP Gauss method is the accurate estimation of the covariance. If the dimensionality increases or the number of samples for estimation drops, we expect instability in the covariance estimation and, therefore, instability in the coverage results. In \cref{fig:ablation_covariance} we see an example of the synthetic experiment when we increase grid granularity (i.e., the dimensionality of the problem) or reduce the number of calibration samples. The same would hold for reducing the number of test samples, but we omit this from further plots so as not to overload the appendix. We can see that a higher granularity results in an earlier onset of infinite bands (as suggested in \cref{thm:tv_gauss_same_cov}) and that the method becomes unstable when the number of estimation samples is too low for the dimensionality of the problem.

\subsection{Visualizations of theoretical results}\label{app:add_exp_expl}
\paragraph{\texorpdfstring{\cref{prop:cp_lifting}}{Proposition 4.7}} We use the backward heat equation to visualize this proposition. As we do not know the exact approximation error, we cannot construct conformal sets as described in the proposition. It is, however, reasonable to assume that the miscoverage should decrease as the approximation error drops (because what we would theoretically add to the numerical conformal intervals will get smaller with finer granularity). Therefore, we sampled numerical trajectories for the backward heat equation (with parameter $\nu=-0.001$ and the granularities seen in the figure) as described in Appendix~\ref{app:data_generation}. For each granularity, we used the sampled initial conditions to generate the analytic solution from its Fourier representation using $16$ modes for sine and cosine each. We then proceeded as in the synthetic case (general task), calibrated with the numerical data, predicted from numerical input, and checked coverage with the analytical output. The depicted coverages come from the first time step after the calibration (at later time steps, the negative effect from \cref{thm:tv_gauss_same_cov} reduced the observed effect, and the methods started undercovering or getting infinite bandwidth).

\paragraph{\texorpdfstring{\cref{thm:tv_gauss_same_cov}}{Proposition 4.4}}
To show that a higher granularity implies a higher TV distance and faster infinite bandwidth, we sampled from the PDE in the proposition (with drift parameter $1$ and the granularities seen in the figure) as described in Appendix~\ref{app:data_generation}. For each granularity, we calculated the coverage as in the synthetic case (general task). The depicted time step is $10$, as this was the time step where the effect for our chosen PDE was best visible (earlier time steps would need higher granularity to drop in coverage, and later ones dropped too quickly). The theoretical lower bound was calculated as described in \citet{barber_conformal_2023} as $1-\alpha-\sum_{i=1}^{n}\tilde w_i \dTV(\vz, \vz^i)$ with equal weights $w_i$. Here, $\vz$ and $\vz^i$ are described in \cref{thm:tv_gauss_same_cov}.

\begin{figure}[ht]
    \centering
    \includegraphics[width=0.6\textwidth]{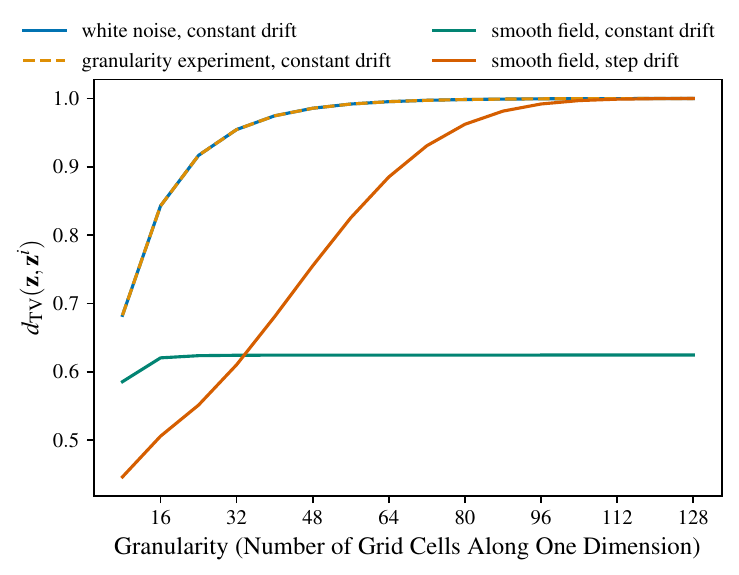}
    \caption{TV distance of \cref{thm:tv_general_drift} at $|t_1-t_2|=0.5$ over the grid granularity. For the constant drift $\vv=\mathbf{1}$ under white noise and the covariance of the granularity experiment in \cref{fig:granularity}, $\mathbf{1}^\top\mSigma_0^{-1}\mathbf{1}=p$ and both curves coincide. For the smooth Gaussian random field $\gN(0, 625(-\partial_x^2+25\mI)^{-2})$ on $(0,2\pi)$, evaluated on the grid, the TV distance levels off below $1$ for the constant drift, but tends to $1$ for a step drift ($v_j=1$ on the left half of the domain, $0$ otherwise).}
    \label{fig:tv_effective_dimension}
\end{figure}

\paragraph{Correlated initial conditions}
\cref{fig:tv_effective_dimension} evaluates the TV distance of \cref{thm:tv_general_drift} for the constant drift $\vv=\mathbf{1}$ under correlated initial conditions. If the initial field and the drift are smooth, neighboring grid points are strongly correlated and $\vv^\top\mSigma_0^{-1}\vv$ stays bounded as the grid is refined, so the TV distance levels off below $1$. A drift that is rougher than the initial field, such as a step, still drives the TV distance to $1$.

\section{Real-world examples}\label{app:real_world}
\subsection{Pulsed-thermography data}\label{app:heat_pulse}
\begin{figure}[ht]
    \centering
    \includegraphics[width=0.99\textwidth]{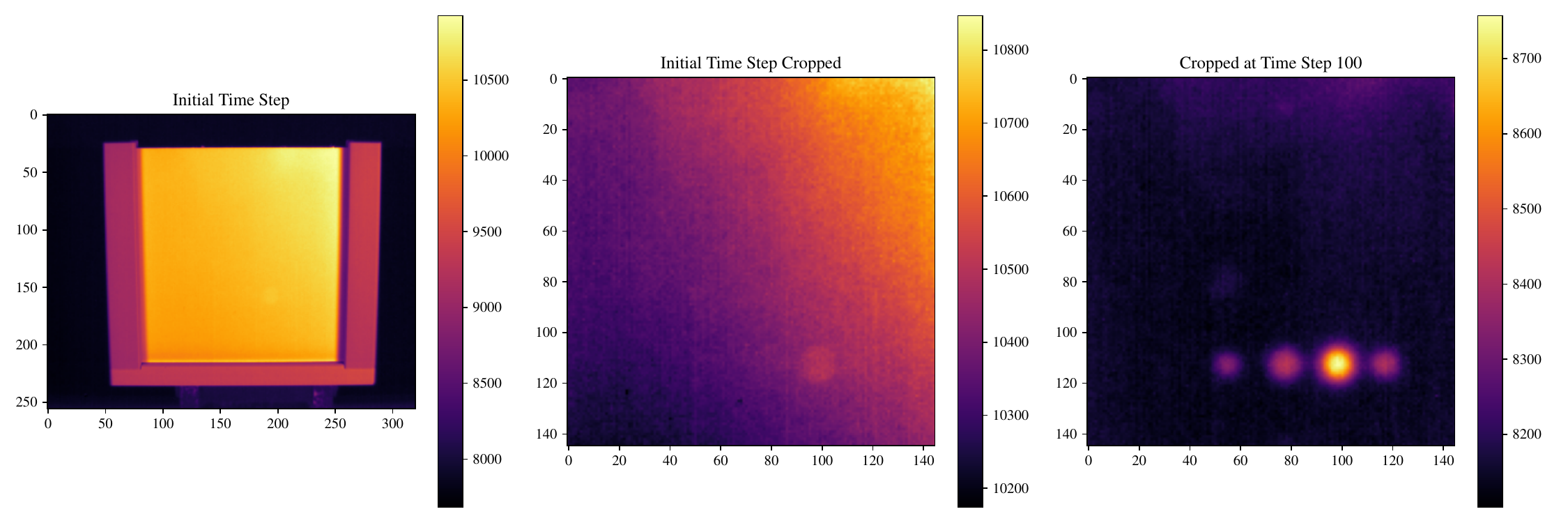}
    \caption{Exemplary frames from the real-world pulsed-thermography dataset. The left image shows a raw data frame from the beginning of the cooling period. The middle image shows how we cropped the frames to remove the background. The right image shows how the data looks after $100$ time steps.}
    \label{fig:heat_data_frame}
\end{figure}

\begin{figure}[h]
    \centering
    \begin{minipage}{0.99\columnwidth}
        \centering
        \begin{minipage}{0.48\linewidth}
            \centering
            \includegraphics[width=\linewidth]{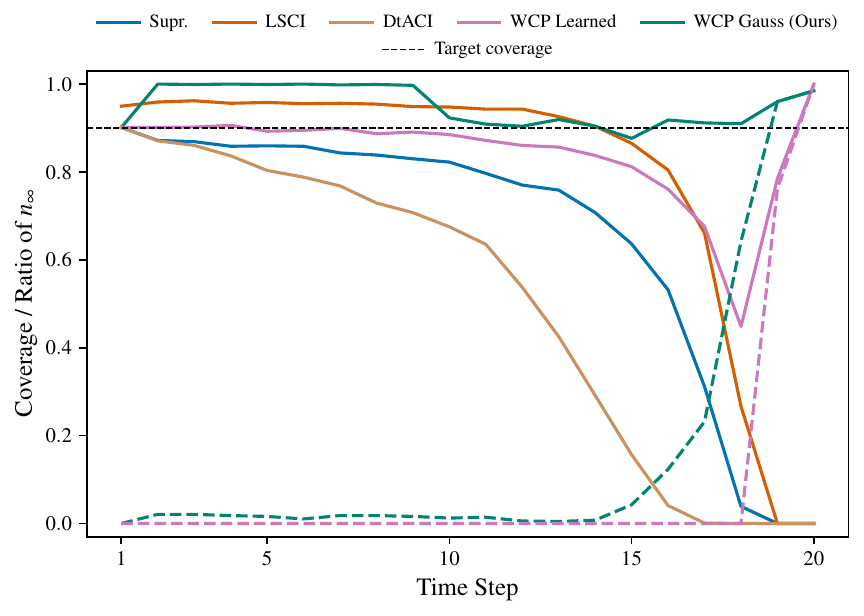}
        \end{minipage}%
        \hspace{0.03\linewidth}%
        \begin{minipage}{0.48\linewidth}
            \centering
            \includegraphics[width=\linewidth]{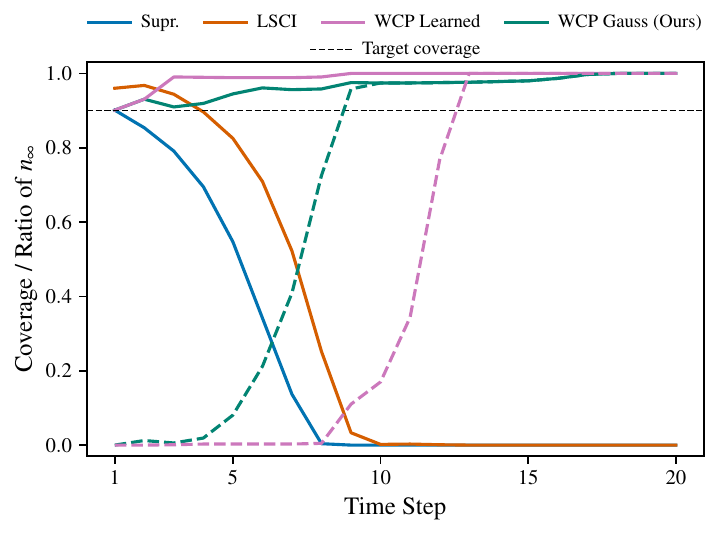}
        \end{minipage}
    \end{minipage}
    \caption{Coverages for the pulsed-thermography dataset for the recalibration task (left) and the general task (right). The 90\% coverage target and, for WCP Gauss (resp. Learned), the fraction of samples with infinite bandwidth $n_\infty$ are marked as a black dashed line and a green (resp. pink) dashed line, respectively. We see that all methods achieve target coverage at first, but only WCP Gauss can preserve coverage over time.}
    \label{fig:heat_coverage}
\end{figure}

The dataset of \citet{wei_pulsed_2023} contains 19 samples of 2D surfaces measured over 1810 time steps. We allocate 3 samples for training a Fourier Neural Operator \cite{li_fourier_2021}, 1 for validation, 13 for calibration and testing. 2 samples were excluded due to large outlier noise.
We crop the borders of the measurement data, since they correspond to non-heated background regions (see \cref{fig:heat_data_frame}). We then determine the time frame where cooling begins, by finding the frame with the largest temperature and use the 20 following time frames as the dataset. As we consider an inverse problem, we flip the time axis. For the recalibration task, we additionally use a time stride (i.e., we remove every second frame from the data), because otherwise all methods always covered (as they provide temporally very fine-grained data). Because the dataset is small, we divide each surface into square patches of size $10\times 10$ and treat them as identically distributed samples (to mitigate dependency between the patches, we leave an empty space of size 5 to all neighboring patches). While this induces some dependency between patches, and the variations in the metal plates (e.g., differently drilled backside holes) introduce slight distribution shifts, both effects are minor for a real-world measurement dataset and have a negligible impact on CP performance. We further do not split the trajectories into different calibration and test sets. We calibrate at an earlier time than we test, but for the same trajectories. This also introduces dependency structures, but again mitigates the low dataset size and mirrors the real-world setting, where the same processes are observed over time. We can see in \cref{fig:heat_coverage} that WCP Gauss is the only method with consistent coverage over time.

\subsection{Lake Maggiore thermistor data}\label{app:lake_maggiore}
\begin{figure}[ht]
    \centering
    \includegraphics[width=0.99\textwidth]{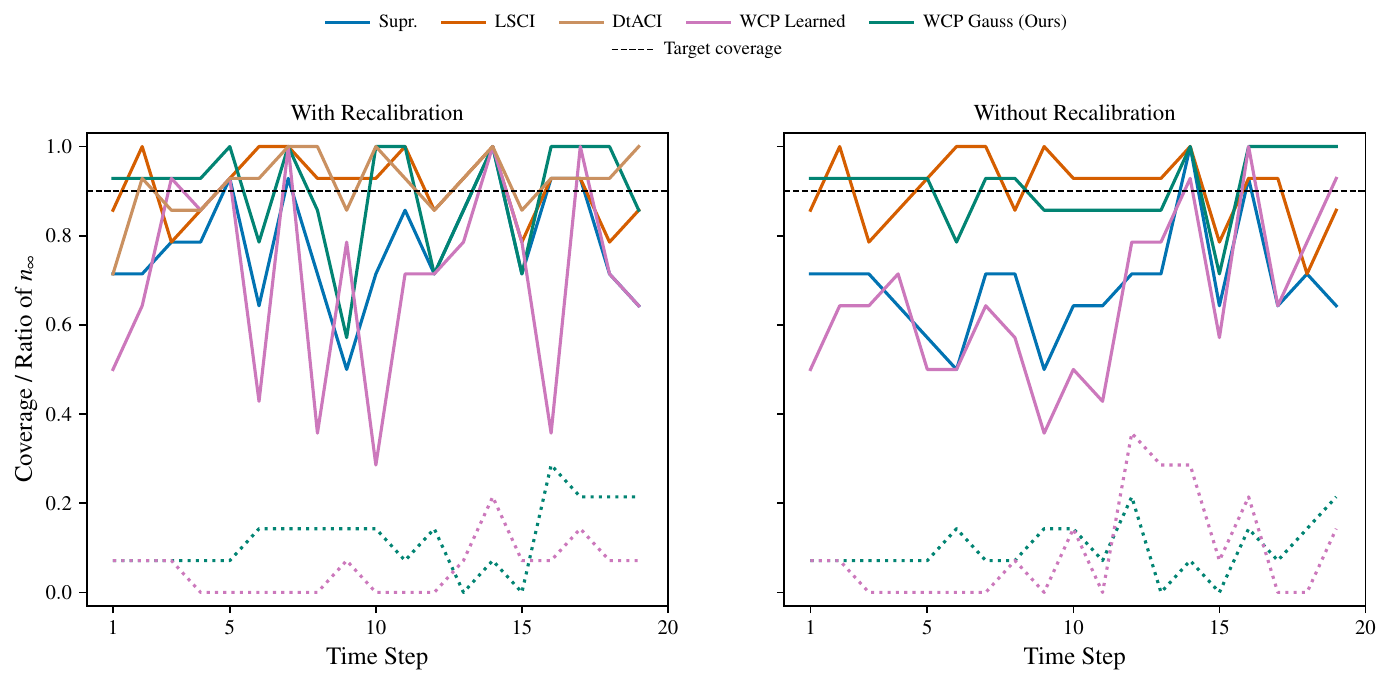}
    \caption{Coverages for the Lake Maggiore nighttime cooling dataset for the recalibration task (left) and the general task (right). The 90\% target is marked by a black dashed line. For WCP Gauss (resp. Learned), the fraction of samples with infinite bandwidth $n_\infty$ is shown as a green (resp. pink) dotted line. WCP Gauss and LSCI maintain the target coverage on both tasks. Deviations at single time steps are statistically expected given the test size of $14$.}
    \label{fig:lake_maggiore_combined}
\end{figure}

The Lake Maggiore thermistor dataset of \citet{scheu_observation_2018}\footnote{\url{https://doi.org/10.5281/zenodo.1206188}} contains $1$-minute-resolution thermistor chain measurements at multiple depths in Pallanza Bay. We use 50 nights in early spring for calibration and 14 nights in late spring for testing. A 1D Fourier Neural Operator \citep{li_fourier_2021} (3 layers, 3 Fourier modes, 32 hidden channels) is trained on 15 further early-spring nights (and validated on 5) to predict one time step ahead, for up to 200 epochs with early stopping. We subsample to $30$-minute resolution across $7$ depth sensors and extract nighttime windows as independent trajectories of vertical heat diffusion without solar forcing, which fits the linear-PDE setting of \cref{prop:wcp_linear_pde}. We z-scored with the global mean and standard deviation of the training nights. As for the pulsed-thermography data, we frame the prediction as an inverse problem and reverse the time axis.
Following the discussion after \cref{thm:tv_general_drift}, we restrict CP to the middle depth sensors ($27$, $30$, and $35$\,m above bed) rather than the full water column, exploiting the more favorable marginal distribution.
\cref{fig:lake_maggiore_combined} shows the resulting coverages for both tasks: WCP Gauss again maintains the target coverage almost everywhere (note that with a test size of $14$, single-step deviations from $90\%$ are statistically expected).

\subsection{FRAP data}\label{app:frap}
\begin{figure}[ht]
    \centering
    \includegraphics[width=0.99\textwidth]{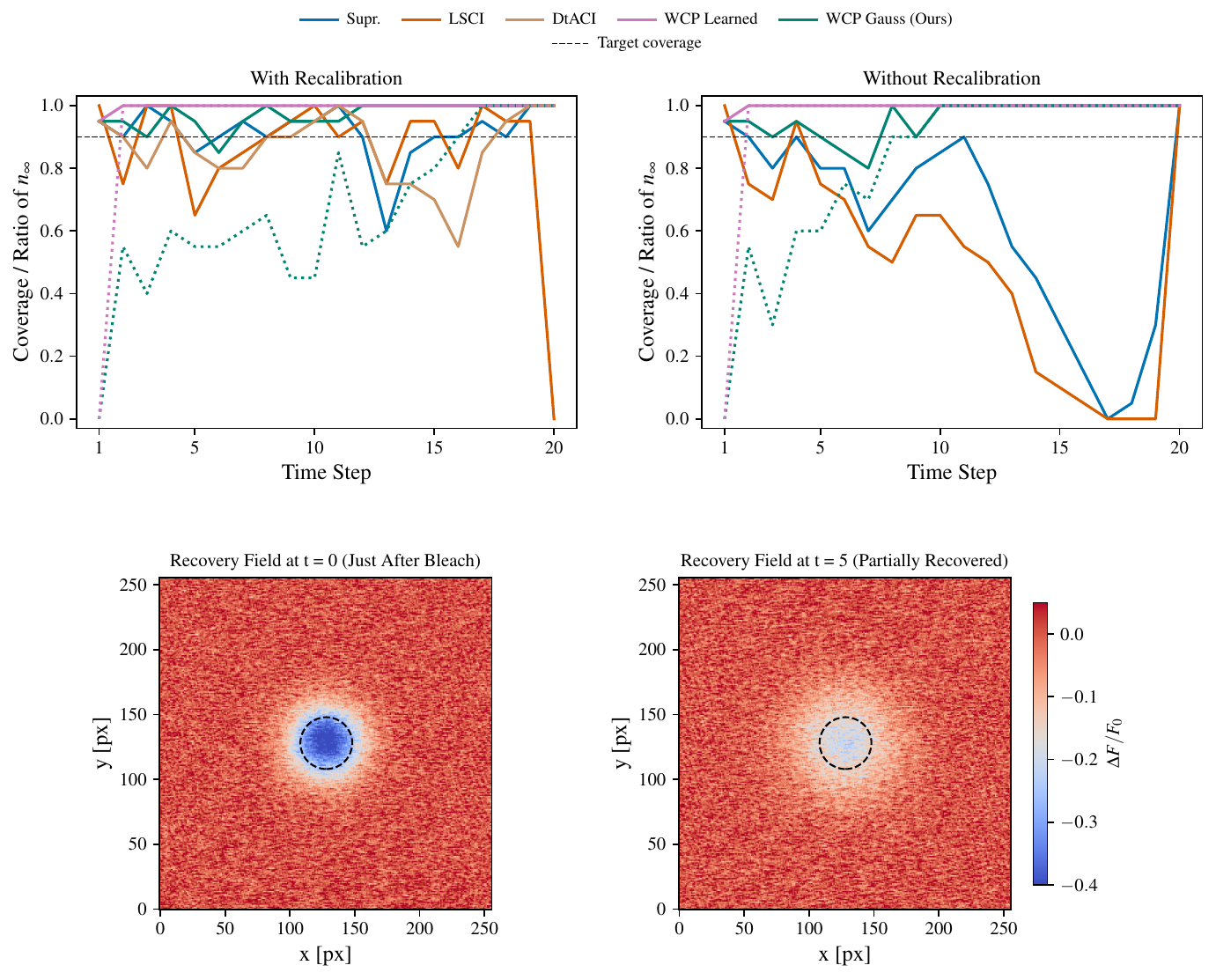}
    \caption{(\emph{Top}) Coverages for the FRAP dataset for the recalibration task (left) and the general task (right). The 90\% target is marked by a black dashed line. For WCP Gauss (resp. Learned), the fraction of samples with infinite bandwidth $n_\infty$ is shown as a green (resp. pink) dotted line. (\emph{Bottom}) Recovery field snapshots from a real FRAP experiment on $32$\,wt\% sucrose solution, showing the normalized fluorescence change $\Delta F / F_0$ just after photobleaching ($t=0$) and after partial recovery ($t=5$). The dashed circle marks the bleach center. WCP Gauss is the only method that provides non-trivial target coverage for every time step on both tasks, where WCP Learned immediately provides trivial bands.}
    \label{fig:frap_combined}
\end{figure}

The FRAP dataset of \citet{wahlstrand_skarstrom_deepfrap_2021}\footnote{\url{https://doi.org/10.5281/zenodo.3874218}} contains fluorescence-recovery-after-photobleaching experiments on $32$\,wt\% sucrose solutions, whose post-bleach intensity field obeys the standard diffusion equation and again fits \cref{prop:wcp_linear_pde}. We train a $2$D Fourier Neural Operator \cite{li_fourier_2021} on $400$ frame pairs from $20$ simulated recovery fields of size $256\times 256$ pixels, and use $20$ real experiments for calibration and test (we use different time steps for calibration and test, not different trajectories). The training fields are produced by a spectral diffusion simulator with randomly sampled diffusion coefficient, bleach depth, additive-noise level, and bleach radius according to the original paper. Real recoveries are computed as $\Delta F/F_0$ relative to the per-pixel pre-bleach mean, percentile-and-MAD clipped to suppress imaging outliers, Gaussian-smoothed with $\sigma{=}0.6$\,px, and per-trajectory z-scored before training. The 5$\times$5 box filter is applied only at CP score extraction and is not seen by the FNO. From every $256\times 256$ frame we extract a $3\times 3$ grid of pixels around the bleach center, spaced $5$ pixels apart and each smoothed with a $5\times 5$ box filter, yielding a $9$-dimensional time series. The time axis is then truncated, time-strided, and reversed to frame the problem as an inverse one.
Following the discussion after \cref{thm:tv_general_drift}, restricting CP to the center of the bleach region (where the laser is focused and the recovery dynamics are concentrated) rather than to the full $256\times 256$ field exploits the more favorable marginal distribution.
\cref{fig:frap_combined} shows the resulting coverages together with sample recovery snapshots. Only WCP Gauss attains non-trivial target coverage at every time step on both tasks (with the same test-size caveat as above).

\section{Robustness experiments}\label{app:robustness}
\subsection{Multi-seed stability}\label{app:multi_seed}
\begin{figure}[ht]
    \centering
    \includegraphics[width=0.99\textwidth]{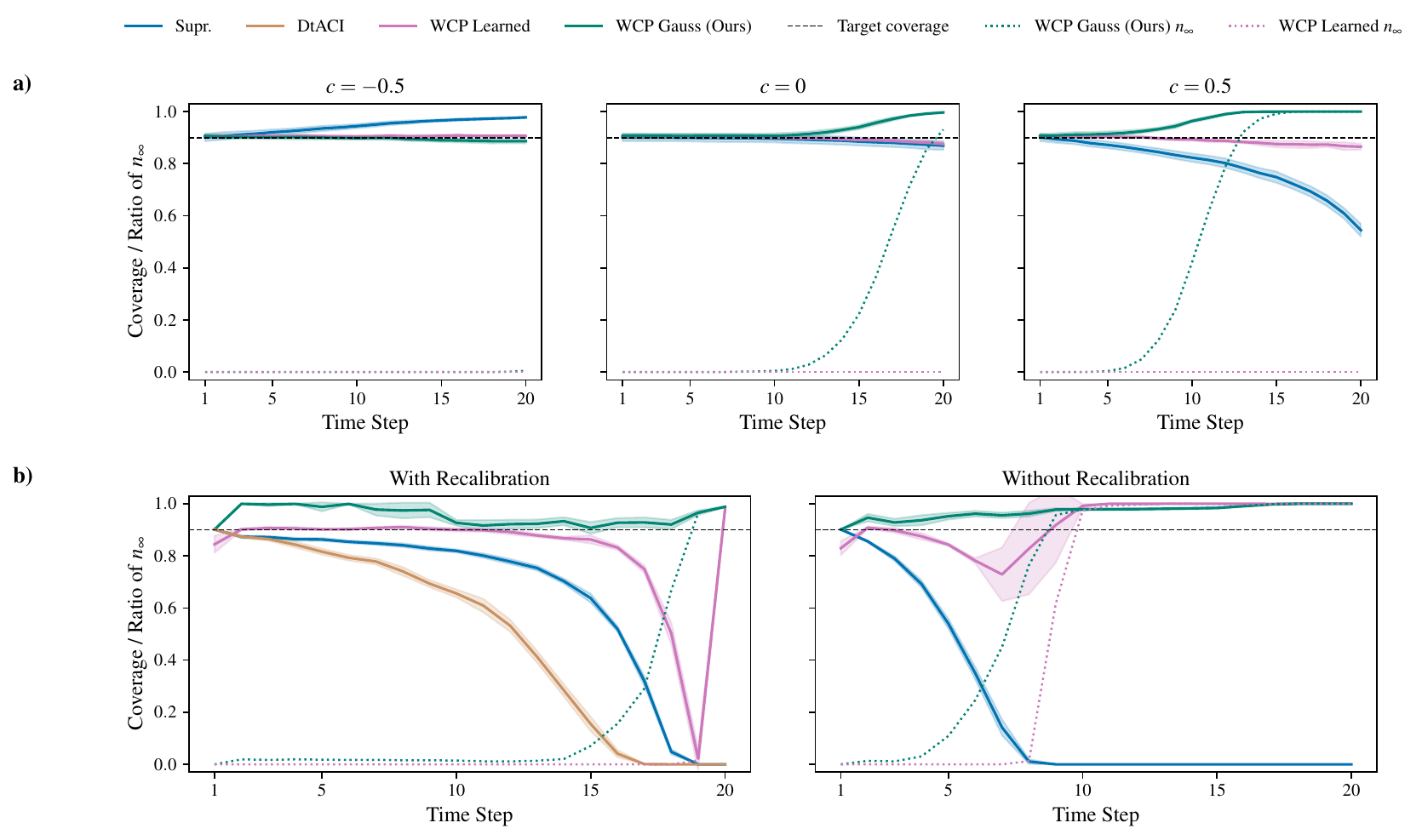}
    \caption{Multi-seed stability. \emph{(a)} Coverages for the general task of the synthetic PDE ($a=-0.0075$, $b=0$) for $c\in\{-0.5,0,0.5\}$, mean $\pm$ one standard deviation over 10 reshuffled calibration/test splits. \emph{(b)} Coverages for the pulsed-thermography dataset for the recalibration task (left) and the general task (right), mean $\pm$ one standard deviation over 10 plate-level reshuffles with a retrained surrogate per seed. The 90\% target is marked by a black dashed line. For WCP Gauss (resp. Learned), the fraction of samples with infinite bandwidth $n_\infty$ is shown as a green (resp. pink) dotted line.}
    \label{fig:multi_seed_stability}
\end{figure}

\cref{fig:multi_seed_stability} repeats the general task of the synthetic experiment (\cref{fig:coverages_b0}, $a=-0.0075$) over 10 reshuffled calibration/test splits and the pulsed-thermography experiment over 10 plate-level reshuffles of the train/validation/calibration/test assignment, each with a retrained surrogate. Since each seed uses a different plate assignment and surrogate, the pulsed-thermography curves differ from those in \cref{fig:heat_coverage}. The standard deviation across seeds is small and the ranking of the methods is the same as in the single-seed figures. LSCI is omitted due to its computational cost.

\subsection{Nonlinear PDEs}\label{app:nonlinear}
\begin{figure}[ht]
    \centering
    \includegraphics[width=0.99\textwidth]{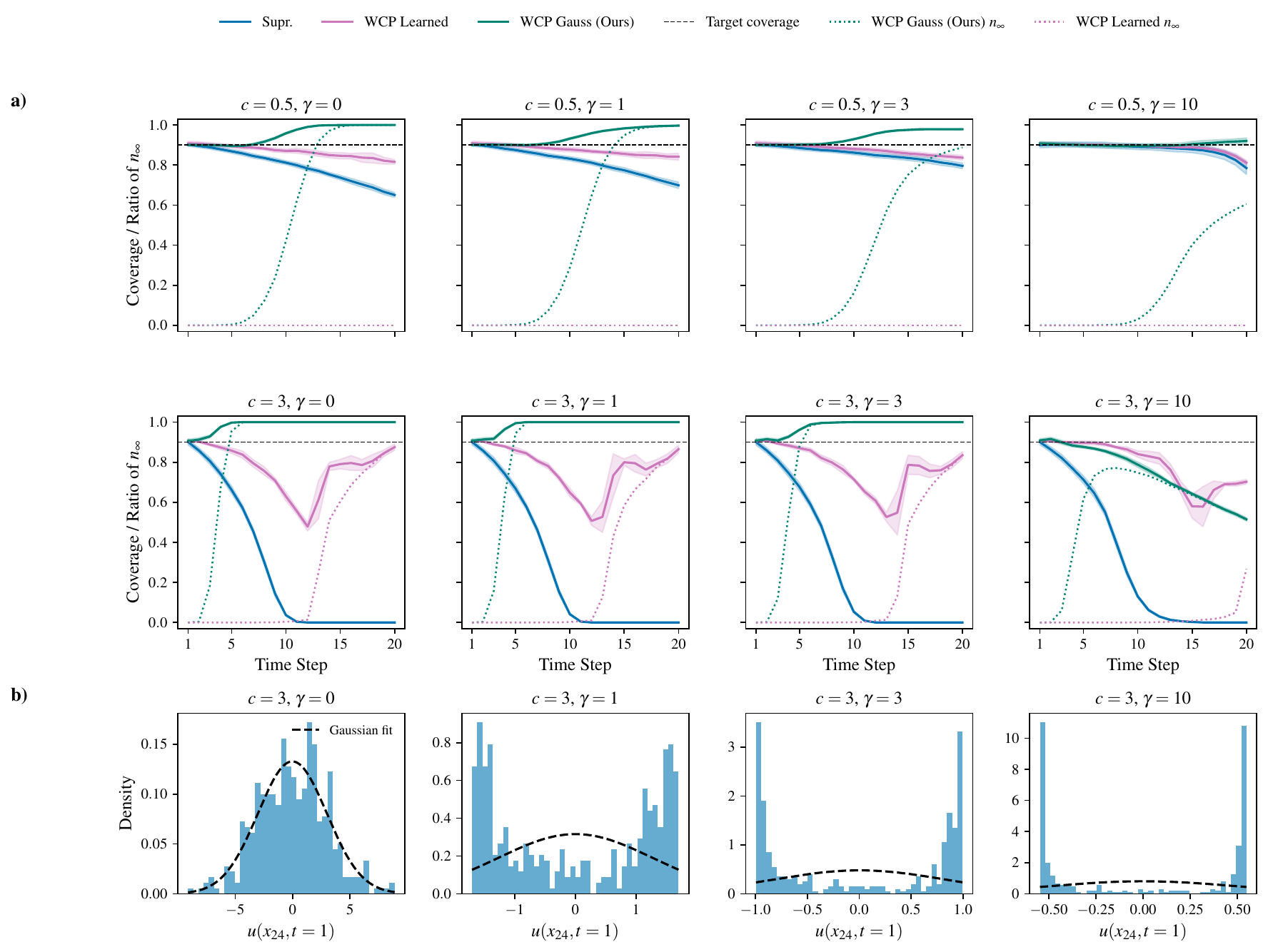}
    \caption{Nonlinear ablation for $u_t = au_{xx} + cu - \gamma u^3$ ($a=-0.0075$, $b=0$). \emph{(a)} Coverages, mean $\pm$ one standard deviation over 10 reshuffled calibration/test splits, for $c=0.5$ (top row) and $c=3$ (bottom row). The 90\% target is marked by a black dashed line. For WCP Gauss (resp. Learned), the fraction of samples with infinite bandwidth $n_\infty$ is shown as a green (resp. pink) dotted line. \emph{(b)} Marginal distribution of the middle grid point at $t=1$ for $c=3$ with a Gaussian fit.}
    \label{fig:nonlinear_ablation}
\end{figure}

We add a cubic reaction term to the synthetic PDE, $u_t = au_{xx} + cu - \gamma u^3$ with $a=-0.0075$ and $b=0$, where $\gamma=0$ recovers the linear PDE. For $\gamma>0$, the solutions are no longer Gaussian and \cref{prop:wcp_linear_pde} does not apply. \cref{fig:nonlinear_ablation} shows that coverage then depends on the PDE parameters: for $c=0.5$, WCP Gauss attains the target for all $\gamma$, while for $c=3$ the solutions saturate into a bimodal distribution and at $\gamma=10$ the coverage of WCP Gauss drops to about $0.5$. Coverage on nonlinear PDEs is therefore not guaranteed and can be lost by changing the parameters.

\subsection{Bandwidths}\label{app:widths}
\begin{figure}[ht]
    \centering
    \includegraphics[width=0.99\textwidth]{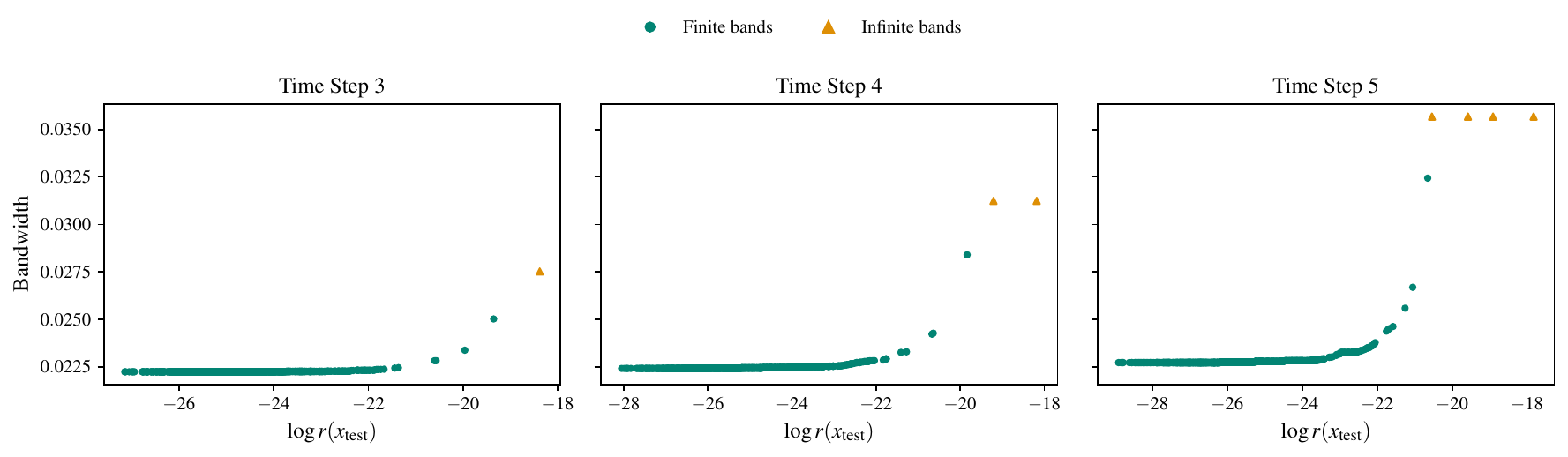}
    \caption{Bandwidth of WCP Gauss per test sample against the test log likelihood-ratio $\log r(x_{\mathrm{test}})$ for the synthetic PDE ($a=-0.0075$, $b=0$, $c=0.5$, general task) at the first time steps with infinite bands. Infinite bands are marked by triangles at the top of each panel.}
    \label{fig:width_vs_loglik}
\end{figure}

The bandwidth of WCP Gauss is determined by the likelihood ratio of the test sample. A test sample that is likely under the test distribution but unlikely under the calibration distribution has a large weight $\tilde w_{n+1}$, which moves the weighted quantile in \cref{eq:wcp_set} upwards. Once $\tilde w_{n+1}>\alpha$, the band is infinite. \cref{fig:width_vs_loglik} shows this for the synthetic PDE: the width increases monotonically with the likelihood ratio, and infinite bands occur exactly for the samples with $\tilde w_{n+1}>\alpha$.

\subsection{Other initial distributions}
\begin{figure}[ht]
    \centering
    \includegraphics[width=0.99\textwidth]{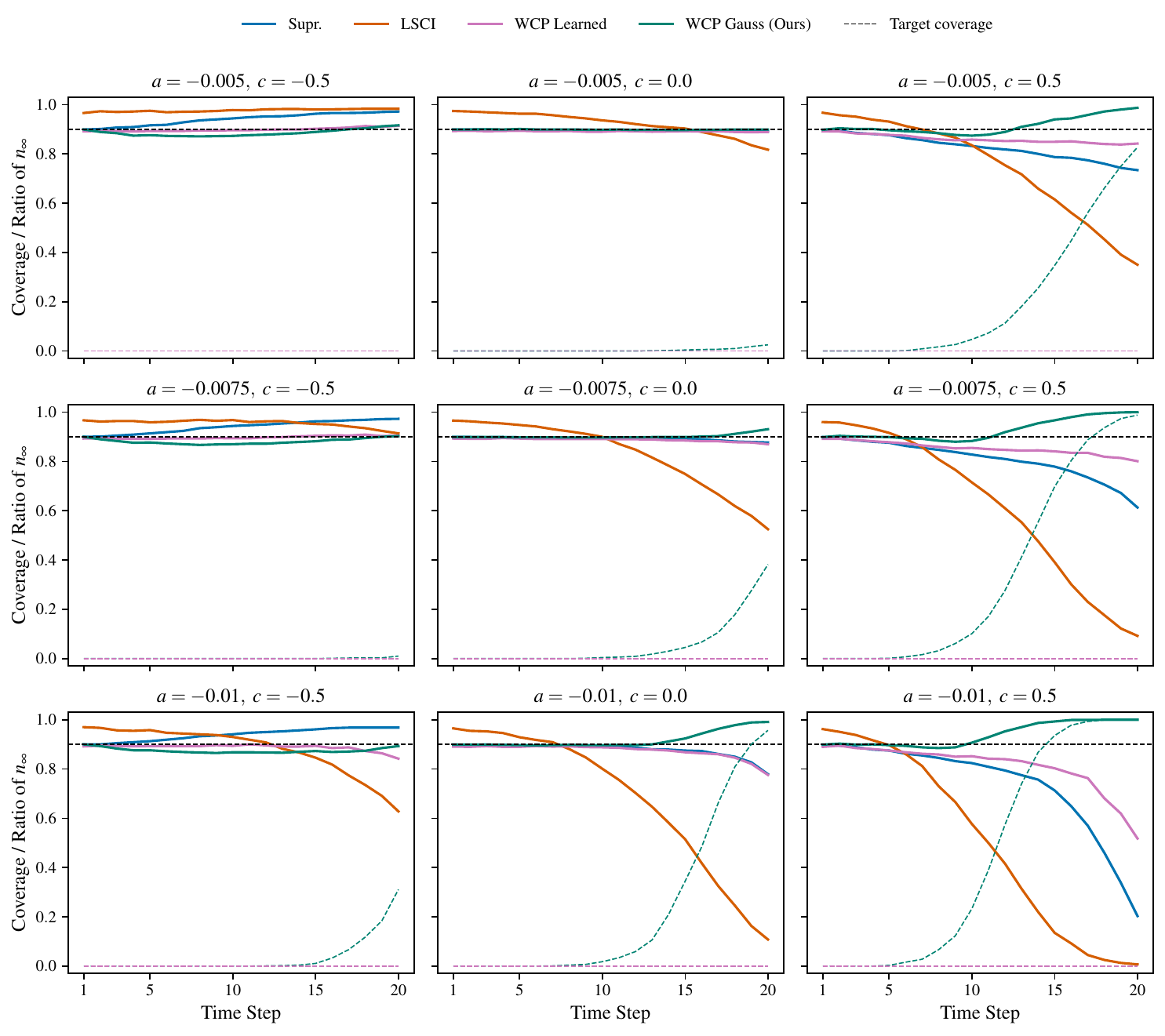}
    \caption{Coverages for the general task for varying $a,c$ and fixed $b=0$ with Laplace initial distribution across increasing prediction horizon. The 90\% coverage target is marked as black dashed line and the fraction of samples with infinite bandwidth $n_\infty$ for WCP Learned and WCP Gauss as dotted line in their respective color.}
    \label{fig:coverages_b0_laplace}
\end{figure}

\begin{figure}[ht]
    \centering
    \includegraphics[width=0.99\textwidth]{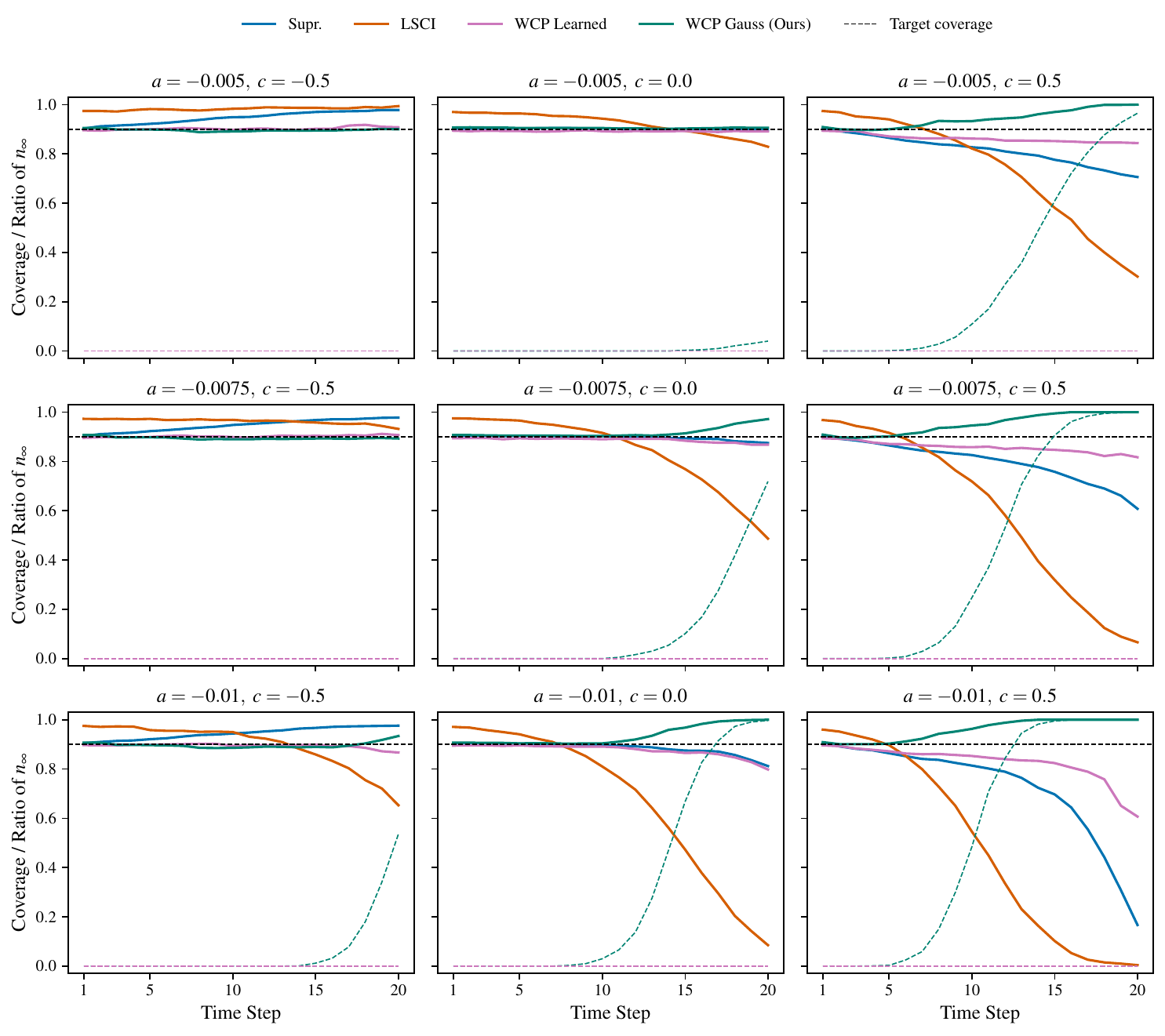}
    \caption{Coverages for the general task for varying $a,c$ and fixed $b=0$ with Logistic initial distribution across increasing prediction horizon. The 90\% coverage target is marked as black dashed line and the fraction of samples with infinite bandwidth $n_\infty$ for WCP Learned and WCP Gauss as dotted line in their respective color.}
    \label{fig:coverages_b0_logistic}
\end{figure}

\begin{figure}[ht]
    \centering
    \includegraphics[width=0.99\textwidth]{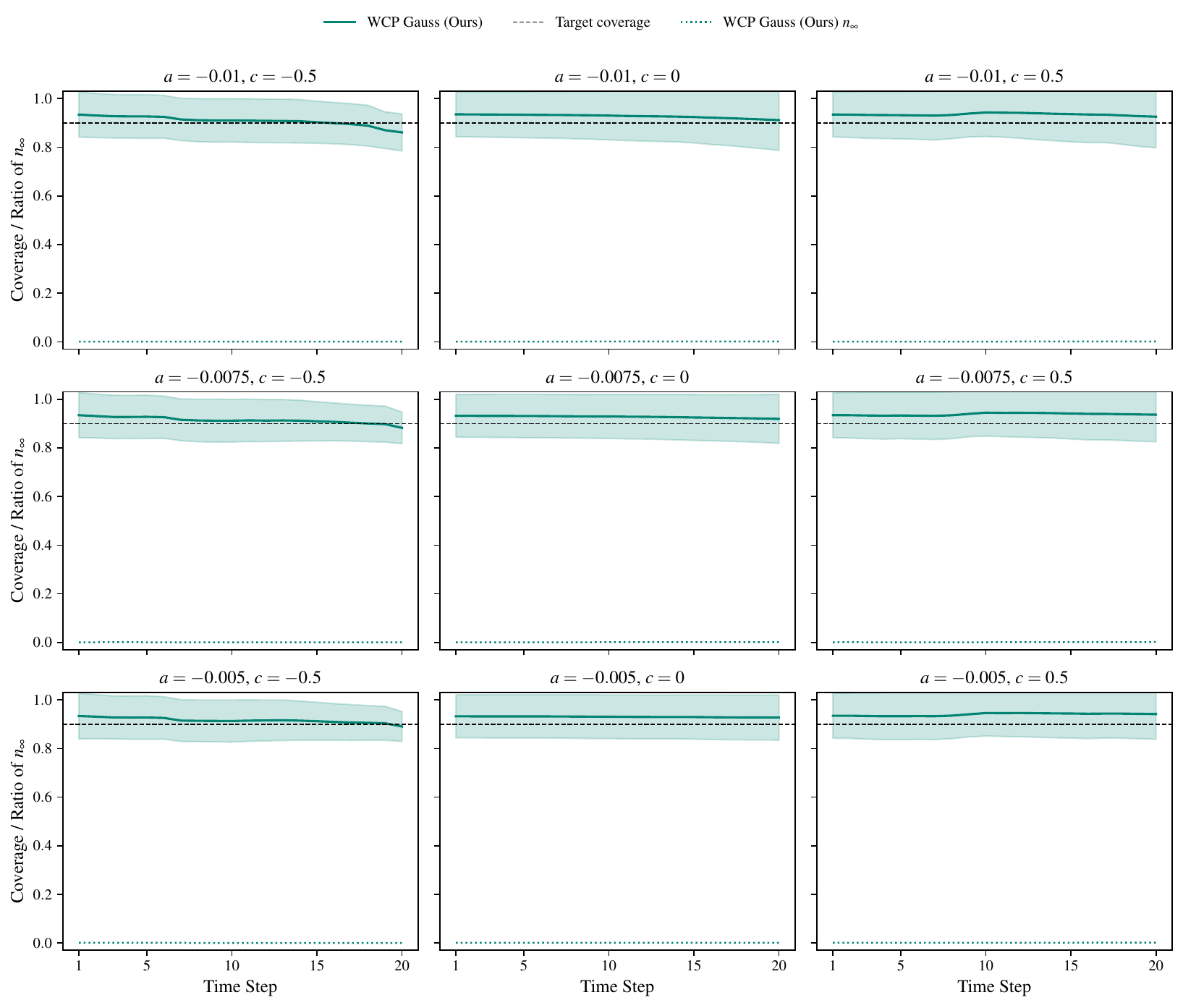}
    \caption{Coverages of WCP with Cauchy likelihood-ratio weights for the general task for varying $a,c$ and fixed $b=0$ with heavy-tailed elliptical Cauchy initial distribution, mean $\pm$ one standard deviation over 10 reshuffled calibration/test splits. The 90\% coverage target is marked as black dashed line and the fraction of samples with infinite bandwidth $n_\infty$ as dotted line. The Cauchy likelihood ratios stay bounded, so no infinite bands occur. The heavy tails increase the variance of the coverage across splits, and for $c=-0.5$ the mean coverage falls slightly below the target at late time steps.}
    \label{fig:coverages_b0_cauchy}
\end{figure}

We further show results when starting sampling from other initial distributions. In \cref{fig:coverages_b0_laplace}, we can see coverages when initially sampling from a Laplace distribution and in \cref{fig:coverages_b0_logistic} when sampling from a Logistic distribution. As the results are almost equivalent to those with Gaussian initial conditions, we only show the graphs for $b=0$ and refrain from further plots or tables. In \cref{fig:coverages_b0_cauchy}, we show coverages of WCP Gauss (with Cauchy instead of Gaussian likelihood-ratio weights) when sampling from heavy-tailed Cauchy initial conditions over multiple seeds. The method stays stable over the seed distribution.

\section{Additional results on synthetic data}\label{app:additional_results}
\begin{figure}[ht]
    \centering
    \includegraphics[width=0.99\textwidth]{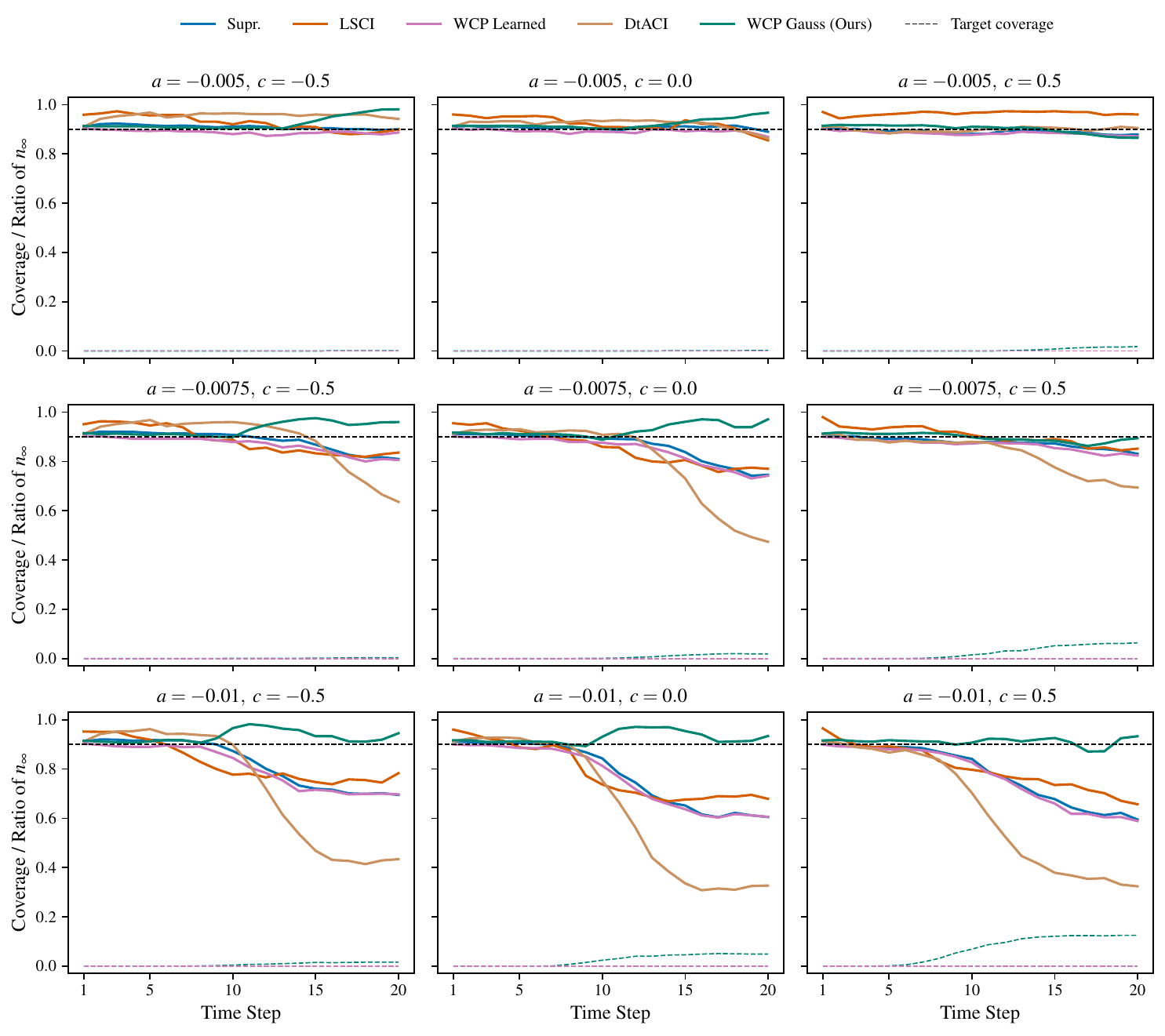}
    \caption{Coverages for the recalibration task for varying $a,c$ and fixed $b=-0.5$ across increasing prediction horizon. The 90\% coverage target is marked as black dashed line and the fraction of samples with infinite bandwidth $n_\infty$ for WCP Learned and WCP Gauss as dotted line in their respective color.}
    \label{fig:coverages_re_b-0p5}
\end{figure}

\begin{figure}[ht]
    \centering
    \includegraphics[width=0.99\textwidth]{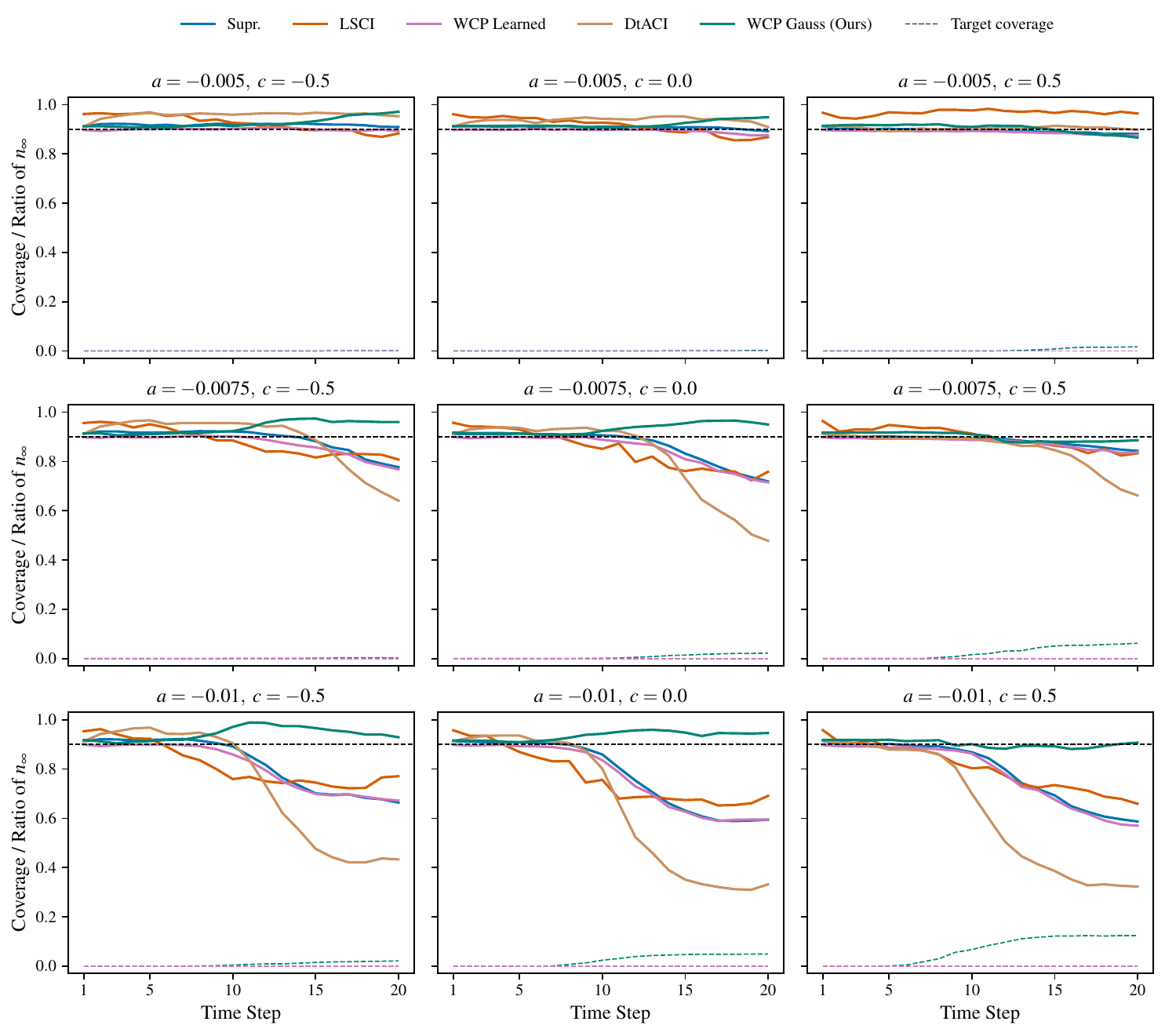}
    \caption{Coverages for the recalibration task for varying $a,c$ and fixed $b=0$ across increasing prediction horizon. The 90\% coverage target is marked as black dashed line and the fraction of samples with infinite bandwidth $n_\infty$ for WCP Learned and WCP Gauss as dotted line in their respective color.}
    \label{fig:coverages_re_b0}
\end{figure}

\begin{figure}[ht]
    \centering
    \includegraphics[width=0.99\textwidth]{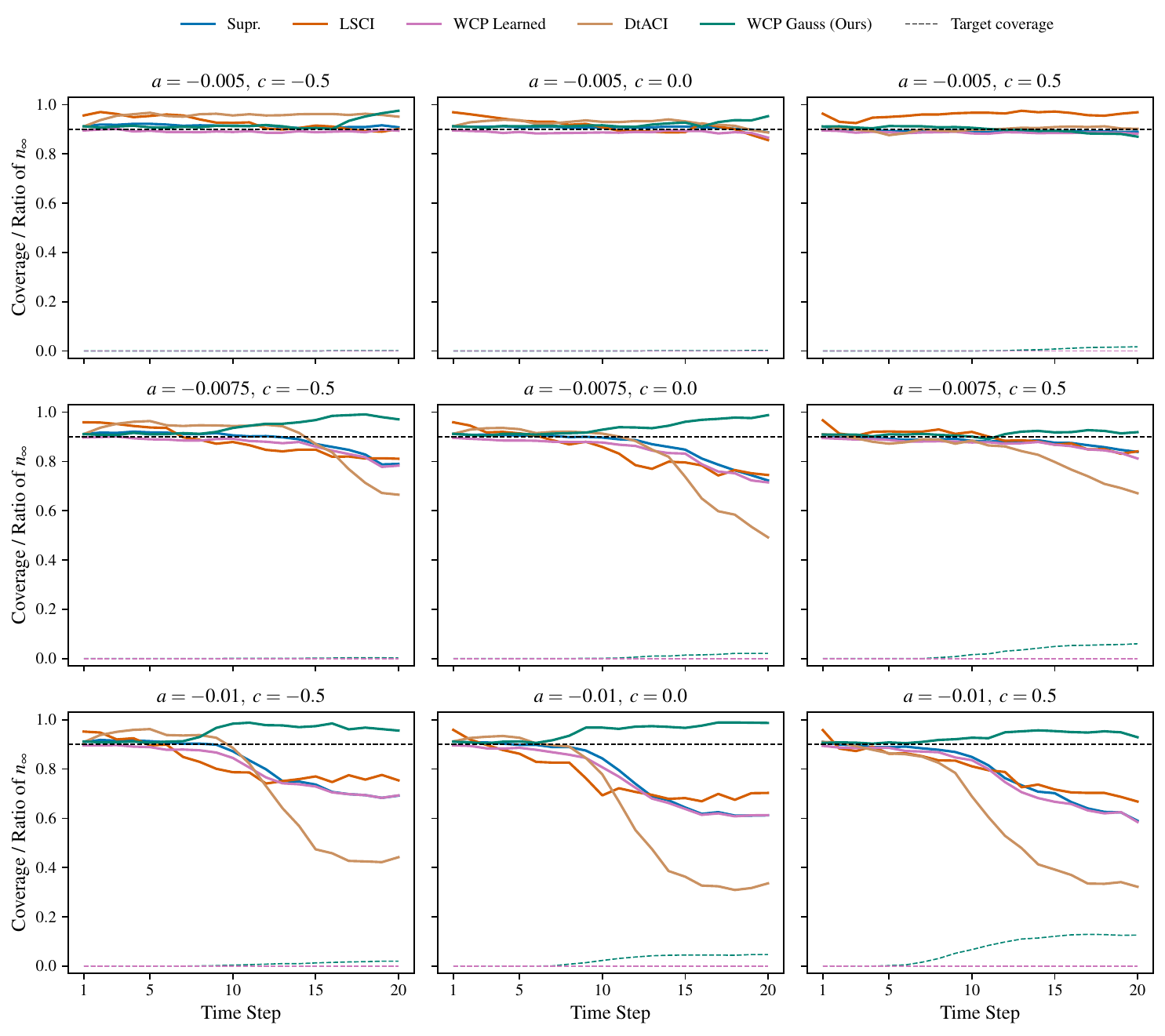}
    \caption{Coverages for the recalibration task for varying $a,c$ and fixed $b=0.5$ across increasing prediction horizon. The 90\% coverage target is marked as black dashed line and the fraction of samples with infinite bandwidth $n_\infty$ for WCP Learned and WCP Gauss as dotted line in their respective color.}
    \label{fig:coverages_re_b0p5}
\end{figure}

\begin{figure}[ht]
    \centering
    \includegraphics[width=0.99\textwidth]{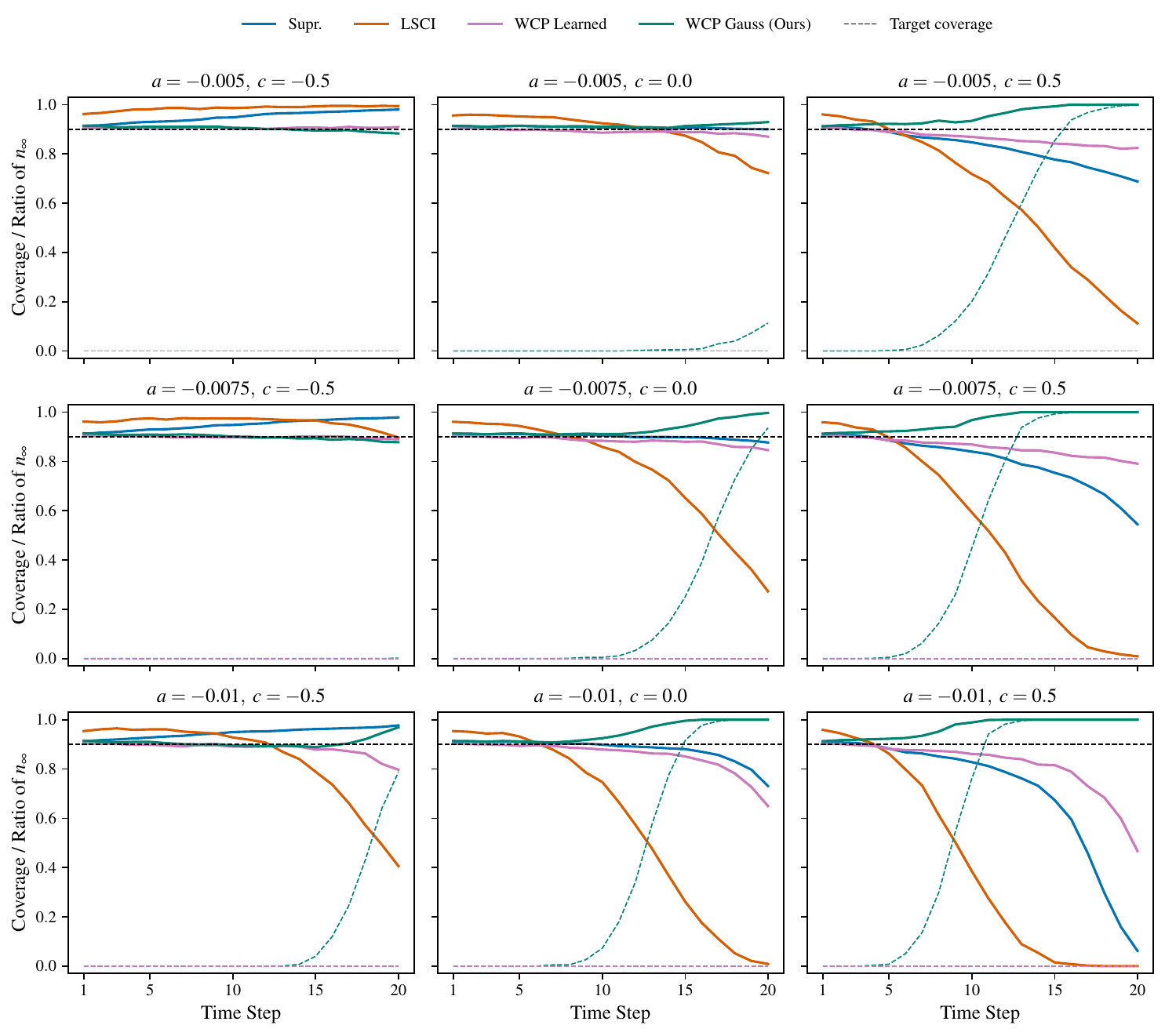}
    \caption{Mean coverages for the general task (no recalibration) for varying $a,c$ and fixed $b=-0.5$ across an increasing prediction horizon. The 90\% coverage target is marked as black dashed line and the fraction of samples with infinite bandwidth $n_\infty$ for WCP Learned and WCP Gauss as dotted line in their respective color.}
    \label{fig:coverages_b-0p5}
\end{figure}

\begin{figure}[ht]
    \centering
    \includegraphics[width=0.99\textwidth]{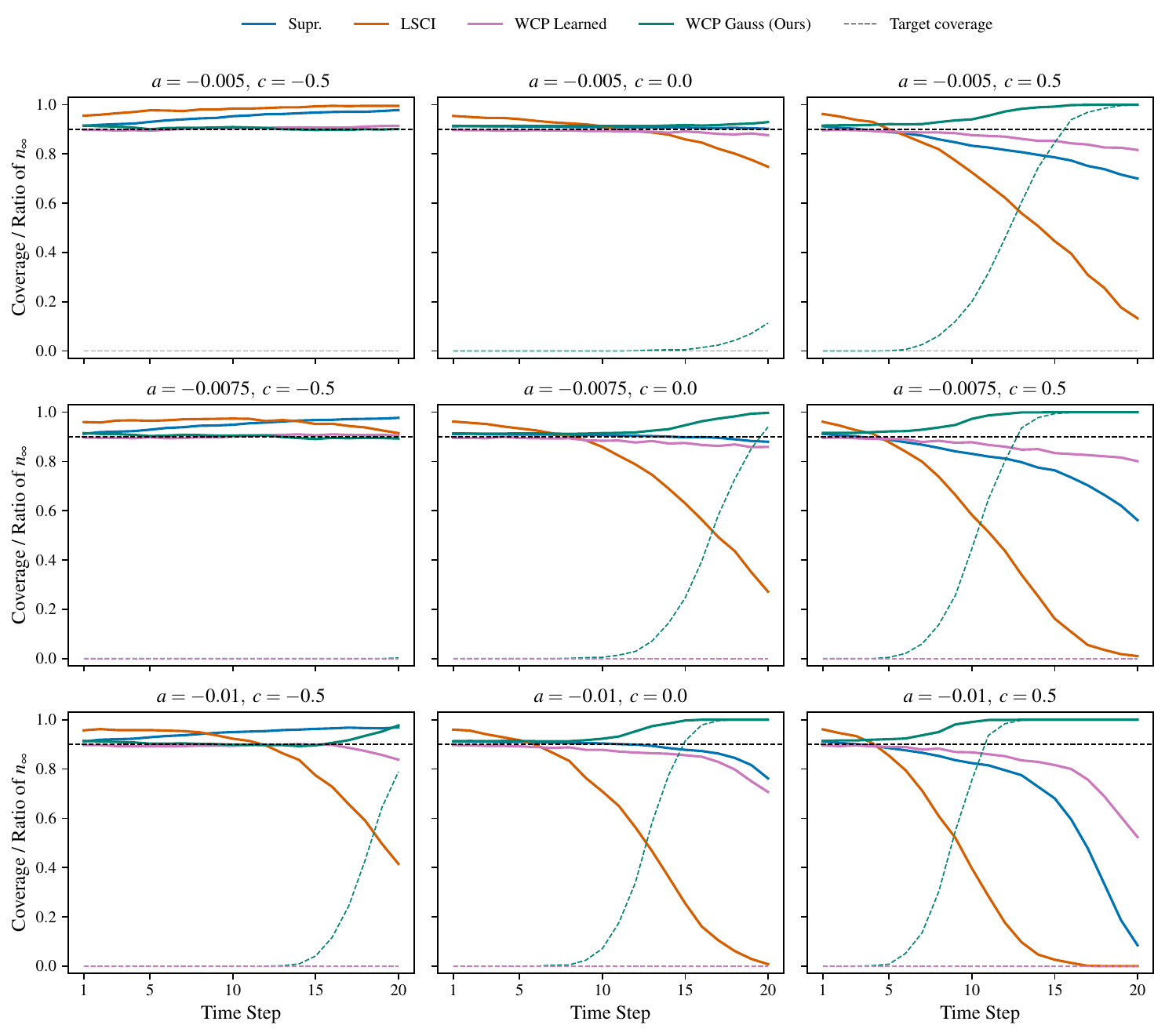}
    \caption{Mean coverages for the general task (no recalibration) for varying $a,c$ and fixed $b=0$ across increasing prediction horizon. The 90\% coverage target is marked as black dashed line and the fraction of samples with infinite bandwidth $n_\infty$ for WCP Learned and WCP Gauss as dotted line in their respective color.}
    \label{fig:coverages_b0}
\end{figure}

\begin{figure}[ht]
    \centering
    \includegraphics[width=0.99\textwidth]{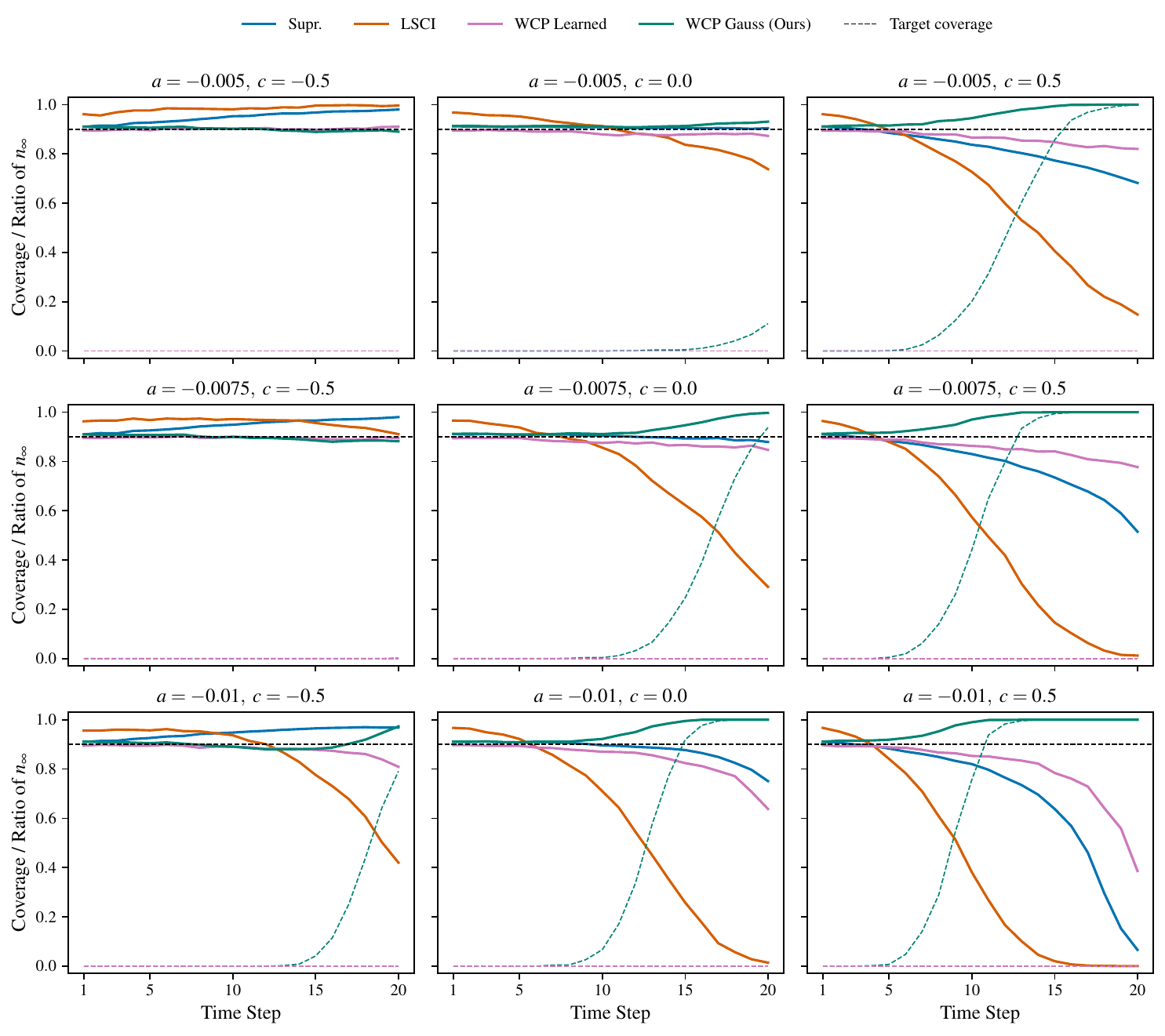}
    \caption{Mean coverages for the general task (no recalibration) for varying $a,c$ and fixed $b=0.5$ across increasing prediction horizon. The 90\% coverage target is marked as black dashed line and the fraction of samples with infinite bandwidth $n_\infty$ for WCP Learned and WCP Gauss as dotted line in their respective color.}
    \label{fig:coverages_b0p5}
\end{figure}

We present the additional results (plots and tables) for the synthetic experiments. For the recalibration task, we find them in \cref{fig:coverages_re_b-0p5} and \cref{tab:synthetic_re_beta-0p5} for $b=-0.5$, in \cref{fig:coverages_re_b0} and \cref{tab:synthetic_re_beta0p0} for $b=0$, and in \cref{fig:coverages_re_b0p5} and \cref{tab:synthetic_re_beta0p5} for $b=0.5$. For the general task, we find them in \cref{fig:coverages_b-0p5} and \cref{tab:synthetic_beta-0p5} for $b=-0.5$, in \cref{fig:coverages_b0} and \cref{tab:synthetic_beta0p0} for $b=0$, and in \cref{fig:coverages_b0p5} and \cref{tab:synthetic_beta0p5} for $b=0.5$. 

All results show similar structures. If the parameter set yields a simple dynamic, all methods cover even far beyond the calibration points. But when the dynamics get rougher, all but the WCP Gauss method undercover. WCP Gauss refrains from false predictions and instead yields infinite bands. DtACI yields worse results than is usually seen in the literature. The reason is that we observe only a short prediction horizon. When using adaptive CP on stock market data for example, thousands of time steps are available and changes happen slowly.

\begingroup
\setlength{\tabcolsep}{2pt}
\renewcommand{\arraystretch}{1.2}
\begin{table}[ht]
\caption{Coverage and mean bandwidth for the recalibration task for varying $a, c$ and fixed $b=-0.5$. We also report the fraction of samples where infinite bands were reported $n_\infty$. The mean bandwidth is taken over finite bands only. The gray font is chosen for better readability.}
\label{tab:synthetic_re_beta-0p5}
\centering\small
\begin{tabular}{lll | ccccc|ccccc|ccccc |}
\hline
& & & \multicolumn{5}{c|}{$c=-0.5$} & \multicolumn{5}{c|}{$c=0.0$} & \multicolumn{5}{c|}{$c=0.5$} \\ \hline
& & T: & 1 & 5 & 10 & 15 & 20 & 1 & 5 & 10 & 15 & 20 & 1 & 5 & 10 & 15 & 20 \\ \hline
\hline
\multirow{12}*{\rotatebox[origin=c]{90}{$a=-0.005$}}
  & \multirow{2}{*}{Supr.} & Cov. & .91 & .92 & .91 & .90 & .90 & .91 & .91 & .90 & .91 & .89 & .91 & .89 & .88 & .89 & .88 \\
  &                            & \textcolor{gray}{Bw.} & \tinygray{.0125} & \tinygray{.0109} & \tinygray{.0089} & \tinygray{.0073} & \tinygray{.0065} & \tinygray{.0133} & \tinygray{.0134} & \tinygray{.0135} & \tinygray{.0145} & \tinygray{.0161} & \tinygray{.0142} & \tinygray{.0171} & \tinygray{.0243} & \tinygray{.0419} & \tinygray{.0793} \\ \cline{2-18}
  & \multirow{2}{*}{LSCI}     & Cov. & .96 & .96 & .92 & .91 & .90 & .96 & .95 & .91 & .94 & .86 & .97 & .96 & .97 & .97 & .96 \\
  &                           & \textcolor{gray}{Bw.} & \tinygray{.0121} & \tinygray{.0069} & \tinygray{.0052} & \tinygray{.0052} & \tinygray{.0056} & \tinygray{.0126} & \tinygray{.0108} & \tinygray{.0101} & \tinygray{.0148} & \tinygray{.0152} & \tinygray{.0133} & \tinygray{.0167} & \tinygray{.0238} & \tinygray{.0508} & \tinygray{.0988} \\ \cline{2-18}
  & \multirow{2}{*}{DtACI} & Cov. & .91 & .97 & .96 & .96 & .94 & .91 & .93 & .93 & .93 & .86 & .91 & .88 & .89 & .91 & .91 \\
  &                            & \textcolor{gray}{Bw.} & \tinygray{.0125} & \tinygray{.0117} & \tinygray{.0094} & \tinygray{.0080} & \tinygray{.0068} & \tinygray{.0133} & \tinygray{.0129} & \tinygray{.0131} & \tinygray{.0138} & \tinygray{.0150} & \tinygray{.0142} & \tinygray{.0145} & \tinygray{.0212} & \tinygray{.0344} & \tinygray{.0650} \\ \cline{2-18}
  & \multirow{3}{*}{\parbox{2cm}{WCP\\Learned}} & Cov. & .90 & .89 & .88 & .89 & .89 & .90 & .89 & .89 & .89 & .87 & .90 & .88 & .88 & .89 & .87 \\
  &                            & \textcolor{gray}{Bw.} & \tinygray{.0117} & \tinygray{.0100} & \tinygray{.0083} & \tinygray{.0070} & \tinygray{.0063} & \tinygray{.0124} & \tinygray{.0125} & \tinygray{.0128} & \tinygray{.0138} & \tinygray{.0158} & \tinygray{.0133} & \tinygray{.0165} & \tinygray{.0238} & \tinygray{.0410} & \tinygray{.0768} \\
  &                            & \textcolor{gray}{$n_\infty$} & \tinygray{0\%} & \tinygray{0\%} & \tinygray{0\%} & \tinygray{0\%} & \tinygray{0\%} & \tinygray{0\%} & \tinygray{0\%} & \tinygray{0\%} & \tinygray{0\%} & \tinygray{0\%} & \tinygray{0\%} & \tinygray{0\%} & \tinygray{0\%} & \tinygray{0\%} & \tinygray{0\%} \\ \cline{2-18}
  & \multirow{3}{*}{\parbox{2cm}{WCP\\Gauss (Ours)}} & Cov. & .91 & .91 & .91 & .93 & .98 & .91 & .91 & .90 & .93 & .97 & .91 & .92 & .91 & .89 & .86 \\
  &                            & \textcolor{gray}{Bw.} & \tinygray{.0127} & \tinygray{.0106} & \tinygray{.0087} & \tinygray{.0080} & \tinygray{.0086} & \tinygray{.0134} & \tinygray{.0138} & \tinygray{.0135} & \tinygray{.0158} & \tinygray{.0218} & \tinygray{.0143} & \tinygray{.0192} & \tinygray{.0283} & \tinygray{.0430} & \tinygray{.0708} \\
  &                            & \textcolor{gray}{$n_\infty$} & \tinygray{0\%} & \tinygray{0\%} & \tinygray{0\%} & \tinygray{0\%} & \tinygray{0\%} & \tinygray{0\%} & \tinygray{0\%} & \tinygray{0\%} & \tinygray{0\%} & \tinygray{0\%} & \tinygray{0\%} & \tinygray{0\%} & \tinygray{0\%} & \tinygray{1\%} & \tinygray{2\%} \\ \hline
\hline
\multirow{12}*{\rotatebox[origin=c]{90}{$a=-0.0075$}}
  & \multirow{2}{*}{Supr.} & Cov. & .91 & .92 & .91 & .87 & .81 & .91 & .91 & .89 & .84 & .75 & .91 & .89 & .88 & .87 & .83 \\
  &                            & \textcolor{gray}{Bw.} & \tinygray{.0126} & \tinygray{.0110} & \tinygray{.0093} & \tinygray{.0089} & \tinygray{.0122} & \tinygray{.0133} & \tinygray{.0136} & \tinygray{.0142} & \tinygray{.0179} & \tinygray{.0316} & \tinygray{.0142} & \tinygray{.0172} & \tinygray{.0251} & \tinygray{.0461} & \tinygray{.1042} \\ \cline{2-18}
  & \multirow{2}{*}{LSCI}     & Cov. & .95 & .95 & .89 & .83 & .84 & .96 & .93 & .86 & .81 & .77 & .98 & .94 & .91 & .89 & .85 \\
  &                           & \textcolor{gray}{Bw.} & \tinygray{.0122} & \tinygray{.0070} & \tinygray{.0068} & \tinygray{.0086} & \tinygray{.0129} & \tinygray{.0127} & \tinygray{.0110} & \tinygray{.0132} & \tinygray{.0182} & \tinygray{.0333} & \tinygray{.0134} & \tinygray{.0142} & \tinygray{.0256} & \tinygray{.0584} & \tinygray{.1263} \\ \cline{2-18}
  & \multirow{2}{*}{DtACI} & Cov. & .91 & .97 & .96 & .88 & .64 & .91 & .93 & .91 & .73 & .47 & .91 & .88 & .88 & .78 & .69 \\
  &                            & \textcolor{gray}{Bw.} & \tinygray{.0126} & \tinygray{.0118} & \tinygray{.0097} & \tinygray{.0090} & \tinygray{.0111} & \tinygray{.0133} & \tinygray{.0130} & \tinygray{.0137} & \tinygray{.0165} & \tinygray{.0279} & \tinygray{.0142} & \tinygray{.0146} & \tinygray{.0220} & \tinygray{.0394} & \tinygray{.0914} \\ \cline{2-18}
  & \multirow{3}{*}{\parbox{2cm}{WCP\\Learned}} & Cov. & .90 & .89 & .88 & .85 & .81 & .90 & .89 & .88 & .81 & .74 & .90 & .88 & .87 & .85 & .82 \\
  &                            & \textcolor{gray}{Bw.} & \tinygray{.0118} & \tinygray{.0102} & \tinygray{.0088} & \tinygray{.0087} & \tinygray{.0122} & \tinygray{.0124} & \tinygray{.0128} & \tinygray{.0138} & \tinygray{.0175} & \tinygray{.0314} & \tinygray{.0133} & \tinygray{.0168} & \tinygray{.0246} & \tinygray{.0442} & \tinygray{.1017} \\
  &                            & \textcolor{gray}{$n_\infty$} & \tinygray{0\%} & \tinygray{0\%} & \tinygray{0\%} & \tinygray{0\%} & \tinygray{0\%} & \tinygray{0\%} & \tinygray{0\%} & \tinygray{0\%} & \tinygray{0\%} & \tinygray{0\%} & \tinygray{0\%} & \tinygray{0\%} & \tinygray{0\%} & \tinygray{0\%} & \tinygray{0\%} \\ \cline{2-18}
  & \multirow{3}{*}{\parbox{2cm}{WCP\\Gauss (Ours)}} & Cov. & .92 & .91 & .90 & .98 & .96 & .92 & .91 & .89 & .96 & .97 & .91 & .91 & .90 & .88 & .89 \\
  &                            & \textcolor{gray}{Bw.} & \tinygray{.0127} & \tinygray{.0107} & \tinygray{.0091} & \tinygray{.0112} & \tinygray{.0146} & \tinygray{.0136} & \tinygray{.0140} & \tinygray{.0141} & \tinygray{.0223} & \tinygray{.0395} & \tinygray{.0144} & \tinygray{.0192} & \tinygray{.0276} & \tinygray{.0453} & \tinygray{.1134} \\
  &                            & \textcolor{gray}{$n_\infty$} & \tinygray{0\%} & \tinygray{0\%} & \tinygray{0\%} & \tinygray{0\%} & \tinygray{0\%} & \tinygray{0\%} & \tinygray{0\%} & \tinygray{0\%} & \tinygray{1\%} & \tinygray{2\%} & \tinygray{0\%} & \tinygray{0\%} & \tinygray{2\%} & \tinygray{5\%} & \tinygray{6\%} \\ \hline
\hline
\multirow{12}*{\rotatebox[origin=c]{90}{$a=-0.01$}}
  & \multirow{2}{*}{Supr.} & Cov. & .91 & .92 & .87 & .72 & .69 & .91 & .91 & .84 & .65 & .60 & .91 & .89 & .84 & .68 & .59 \\
  &                            & \textcolor{gray}{Bw.} & \tinygray{.0126} & \tinygray{.0112} & \tinygray{.0102} & \tinygray{.0154} & \tinygray{.0320} & \tinygray{.0133} & \tinygray{.0139} & \tinygray{.0160} & \tinygray{.0312} & \tinygray{.0829} & \tinygray{.0143} & \tinygray{.0174} & \tinygray{.0274} & \tinygray{.0696} & \tinygray{.2281} \\ \cline{2-18}
  & \multirow{2}{*}{LSCI}     & Cov. & .95 & .92 & .78 & .75 & .78 & .96 & .89 & .74 & .68 & .68 & .97 & .89 & .80 & .74 & .66 \\
  &                           & \textcolor{gray}{Bw.} & \tinygray{.0123} & \tinygray{.0075} & \tinygray{.0090} & \tinygray{.0161} & \tinygray{.0349} & \tinygray{.0127} & \tinygray{.0094} & \tinygray{.0151} & \tinygray{.0328} & \tinygray{.0897} & \tinygray{.0134} & \tinygray{.0140} & \tinygray{.0304} & \tinygray{.0810} & \tinygray{.2472} \\ \cline{2-18}
  & \multirow{2}{*}{DtACI} & Cov. & .91 & .96 & .90 & .47 & .43 & .91 & .93 & .75 & .34 & .33 & .91 & .87 & .70 & .38 & .32 \\
  &                            & \textcolor{gray}{Bw.} & \tinygray{.0126} & \tinygray{.0118} & \tinygray{.0102} & \tinygray{.0137} & \tinygray{.0282} & \tinygray{.0133} & \tinygray{.0130} & \tinygray{.0149} & \tinygray{.0275} & \tinygray{.0731} & \tinygray{.0143} & \tinygray{.0147} & \tinygray{.0242} & \tinygray{.0617} & \tinygray{.2040} \\ \cline{2-18}
  & \multirow{3}{*}{\parbox{2cm}{WCP\\Learned}} & Cov. & .90 & .89 & .84 & .72 & .70 & .90 & .89 & .81 & .64 & .61 & .90 & .88 & .83 & .66 & .59 \\
  &                            & \textcolor{gray}{Bw.} & \tinygray{.0118} & \tinygray{.0103} & \tinygray{.0099} & \tinygray{.0154} & \tinygray{.0320} & \tinygray{.0124} & \tinygray{.0129} & \tinygray{.0155} & \tinygray{.0310} & \tinygray{.0829} & \tinygray{.0134} & \tinygray{.0171} & \tinygray{.0269} & \tinygray{.0685} & \tinygray{.2275} \\
  &                            & \textcolor{gray}{$n_\infty$} & \tinygray{0\%} & \tinygray{0\%} & \tinygray{0\%} & \tinygray{0\%} & \tinygray{0\%} & \tinygray{0\%} & \tinygray{0\%} & \tinygray{0\%} & \tinygray{0\%} & \tinygray{0\%} & \tinygray{0\%} & \tinygray{0\%} & \tinygray{0\%} & \tinygray{0\%} & \tinygray{0\%} \\ \cline{2-18}
  & \multirow{3}{*}{\parbox{2cm}{WCP\\Gauss (Ours)}} & Cov. & .92 & .91 & .97 & .93 & .94 & .92 & .91 & .93 & .95 & .93 & .92 & .92 & .91 & .93 & .93 \\
  &                            & \textcolor{gray}{Bw.} & \tinygray{.0128} & \tinygray{.0111} & \tinygray{.0125} & \tinygray{.0179} & \tinygray{.0394} & \tinygray{.0137} & \tinygray{.0144} & \tinygray{.0190} & \tinygray{.0390} & \tinygray{.1034} & \tinygray{.0146} & \tinygray{.0196} & \tinygray{.0318} & \tinygray{.0876} & \tinygray{.2969} \\
  &                            & \textcolor{gray}{$n_\infty$} & \tinygray{0\%} & \tinygray{0\%} & \tinygray{0\%} & \tinygray{2\%} & \tinygray{2\%} & \tinygray{0\%} & \tinygray{0\%} & \tinygray{2\%} & \tinygray{5\%} & \tinygray{5\%} & \tinygray{0\%} & \tinygray{0\%} & \tinygray{7\%} & \tinygray{12\%} & \tinygray{12\%} \\ \hline
\hline
\end{tabular}
\end{table}
\endgroup

\begingroup
\setlength{\tabcolsep}{2pt}
\renewcommand{\arraystretch}{1.2}
\begin{table}[ht]
\caption{Coverage and mean bandwidth for the recalibration task for varying $a, c$ and fixed $b=0.0$. We also report the fraction of samples where infinite bands were reported $n_\infty$. The mean bandwidth is taken over finite bands only. The gray font is chosen for better readability.}
\label{tab:synthetic_re_beta0p0}
\centering\small
\begin{tabular}{lll | ccccc|ccccc|ccccc |}
\hline
& & & \multicolumn{5}{c|}{$c=-0.5$} & \multicolumn{5}{c|}{$c=0.0$} & \multicolumn{5}{c|}{$c=0.5$} \\ \hline
& & T: & 1 & 5 & 10 & 15 & 20 & 1 & 5 & 10 & 15 & 20 & 1 & 5 & 10 & 15 & 20 \\ \hline
\hline
\multirow{12}*{\rotatebox[origin=c]{90}{$a=-0.005$}}
  & \multirow{2}{*}{Supr.} & Cov. & .91 & .92 & .92 & .92 & .91 & .91 & .91 & .91 & .91 & .89 & .91 & .90 & .90 & .89 & .88 \\
  &                            & \textcolor{gray}{Bw.} & \tinygray{.0125} & \tinygray{.0108} & \tinygray{.0089} & \tinygray{.0073} & \tinygray{.0063} & \tinygray{.0132} & \tinygray{.0133} & \tinygray{.0136} & \tinygray{.0144} & \tinygray{.0157} & \tinygray{.0141} & \tinygray{.0172} & \tinygray{.0249} & \tinygray{.0417} & \tinygray{.0798} \\ \cline{2-18}
  & \multirow{2}{*}{LSCI}     & Cov. & .96 & .97 & .93 & .90 & .88 & .96 & .95 & .93 & .89 & .87 & .97 & .97 & .98 & .97 & .96 \\
  &                           & \textcolor{gray}{Bw.} & \tinygray{.0120} & \tinygray{.0073} & \tinygray{.0049} & \tinygray{.0048} & \tinygray{.0051} & \tinygray{.0124} & \tinygray{.0126} & \tinygray{.0098} & \tinygray{.0110} & \tinygray{.0145} & \tinygray{.0131} & \tinygray{.0187} & \tinygray{.0235} & \tinygray{.0495} & \tinygray{.0972} \\ \cline{2-18}
  & \multirow{2}{*}{DtACI} & Cov. & .91 & .97 & .96 & .97 & .95 & .91 & .94 & .94 & .95 & .91 & .91 & .89 & .91 & .91 & .90 \\
  &                            & \textcolor{gray}{Bw.} & \tinygray{.0125} & \tinygray{.0116} & \tinygray{.0091} & \tinygray{.0076} & \tinygray{.0065} & \tinygray{.0132} & \tinygray{.0127} & \tinygray{.0128} & \tinygray{.0135} & \tinygray{.0149} & \tinygray{.0141} & \tinygray{.0144} & \tinygray{.0204} & \tinygray{.0336} & \tinygray{.0634} \\ \cline{2-18}
  & \multirow{3}{*}{\parbox{2cm}{WCP\\Learned}} & Cov. & .90 & .90 & .90 & .90 & .89 & .90 & .90 & .90 & .89 & .88 & .90 & .89 & .89 & .88 & .88 \\
  &                            & \textcolor{gray}{Bw.} & \tinygray{.0116} & \tinygray{.0099} & \tinygray{.0083} & \tinygray{.0069} & \tinygray{.0061} & \tinygray{.0122} & \tinygray{.0124} & \tinygray{.0129} & \tinygray{.0137} & \tinygray{.0152} & \tinygray{.0132} & \tinygray{.0164} & \tinygray{.0241} & \tinygray{.0394} & \tinygray{.0759} \\
  &                            & \textcolor{gray}{$n_\infty$} & \tinygray{0\%} & \tinygray{0\%} & \tinygray{0\%} & \tinygray{0\%} & \tinygray{0\%} & \tinygray{0\%} & \tinygray{0\%} & \tinygray{0\%} & \tinygray{0\%} & \tinygray{0\%} & \tinygray{0\%} & \tinygray{0\%} & \tinygray{0\%} & \tinygray{0\%} & \tinygray{0\%} \\ \cline{2-18}
  & \multirow{3}{*}{\parbox{2cm}{WCP\\Gauss (Ours)}} & Cov. & .91 & .91 & .91 & .93 & .97 & .91 & .92 & .91 & .93 & .95 & .91 & .92 & .91 & .90 & .87 \\
  &                            & \textcolor{gray}{Bw.} & \tinygray{.0125} & \tinygray{.0103} & \tinygray{.0086} & \tinygray{.0078} & \tinygray{.0078} & \tinygray{.0133} & \tinygray{.0135} & \tinygray{.0135} & \tinygray{.0153} & \tinygray{.0192} & \tinygray{.0143} & \tinygray{.0188} & \tinygray{.0276} & \tinygray{.0437} & \tinygray{.0681} \\
  &                            & \textcolor{gray}{$n_\infty$} & \tinygray{0\%} & \tinygray{0\%} & \tinygray{0\%} & \tinygray{0\%} & \tinygray{0\%} & \tinygray{0\%} & \tinygray{0\%} & \tinygray{0\%} & \tinygray{0\%} & \tinygray{0\%} & \tinygray{0\%} & \tinygray{0\%} & \tinygray{0\%} & \tinygray{1\%} & \tinygray{2\%} \\ \hline
\hline
\multirow{12}*{\rotatebox[origin=c]{90}{$a=-0.0075$}}
  & \multirow{2}{*}{Supr.} & Cov. & .91 & .92 & .92 & .88 & .78 & .91 & .91 & .91 & .83 & .72 & .91 & .90 & .89 & .88 & .84 \\
  &                            & \textcolor{gray}{Bw.} & \tinygray{.0125} & \tinygray{.0109} & \tinygray{.0092} & \tinygray{.0087} & \tinygray{.0116} & \tinygray{.0132} & \tinygray{.0136} & \tinygray{.0144} & \tinygray{.0172} & \tinygray{.0300} & \tinygray{.0142} & \tinygray{.0173} & \tinygray{.0257} & \tinygray{.0460} & \tinygray{.1006} \\ \cline{2-18}
  & \multirow{2}{*}{LSCI}     & Cov. & .96 & .95 & .89 & .82 & .81 & .96 & .93 & .85 & .76 & .76 & .96 & .95 & .91 & .86 & .83 \\
  &                           & \textcolor{gray}{Bw.} & \tinygray{.0121} & \tinygray{.0068} & \tinygray{.0062} & \tinygray{.0079} & \tinygray{.0121} & \tinygray{.0125} & \tinygray{.0097} & \tinygray{.0111} & \tinygray{.0164} & \tinygray{.0316} & \tinygray{.0132} & \tinygray{.0156} & \tinygray{.0250} & \tinygray{.0567} & \tinygray{.1235} \\ \cline{2-18}
  & \multirow{2}{*}{DtACI} & Cov. & .91 & .97 & .96 & .89 & .64 & .91 & .94 & .92 & .73 & .48 & .91 & .89 & .89 & .85 & .66 \\
  &                            & \textcolor{gray}{Bw.} & \tinygray{.0125} & \tinygray{.0117} & \tinygray{.0093} & \tinygray{.0086} & \tinygray{.0107} & \tinygray{.0132} & \tinygray{.0128} & \tinygray{.0133} & \tinygray{.0159} & \tinygray{.0269} & \tinygray{.0142} & \tinygray{.0144} & \tinygray{.0210} & \tinygray{.0382} & \tinygray{.0876} \\ \cline{2-18}
  & \multirow{3}{*}{\parbox{2cm}{WCP\\Learned}} & Cov. & .90 & .90 & .90 & .86 & .77 & .90 & .90 & .89 & .81 & .71 & .90 & .89 & .89 & .87 & .84 \\
  &                            & \textcolor{gray}{Bw.} & \tinygray{.0116} & \tinygray{.0101} & \tinygray{.0086} & \tinygray{.0085} & \tinygray{.0116} & \tinygray{.0123} & \tinygray{.0127} & \tinygray{.0137} & \tinygray{.0168} & \tinygray{.0298} & \tinygray{.0132} & \tinygray{.0164} & \tinygray{.0248} & \tinygray{.0449} & \tinygray{.0991} \\
  &                            & \textcolor{gray}{$n_\infty$} & \tinygray{0\%} & \tinygray{0\%} & \tinygray{0\%} & \tinygray{0\%} & \tinygray{0\%} & \tinygray{0\%} & \tinygray{0\%} & \tinygray{0\%} & \tinygray{0\%} & \tinygray{0\%} & \tinygray{0\%} & \tinygray{0\%} & \tinygray{0\%} & \tinygray{0\%} & \tinygray{0\%} \\ \cline{2-18}
  & \multirow{3}{*}{\parbox{2cm}{WCP\\Gauss (Ours)}} & Cov. & .91 & .91 & .92 & .97 & .96 & .92 & .91 & .92 & .95 & .95 & .92 & .92 & .91 & .88 & .89 \\
  &                            & \textcolor{gray}{Bw.} & \tinygray{.0126} & \tinygray{.0105} & \tinygray{.0094} & \tinygray{.0111} & \tinygray{.0140} & \tinygray{.0135} & \tinygray{.0136} & \tinygray{.0153} & \tinygray{.0220} & \tinygray{.0368} & \tinygray{.0144} & \tinygray{.0189} & \tinygray{.0279} & \tinygray{.0447} & \tinygray{.1084} \\
  &                            & \textcolor{gray}{$n_\infty$} & \tinygray{0\%} & \tinygray{0\%} & \tinygray{0\%} & \tinygray{0\%} & \tinygray{0\%} & \tinygray{0\%} & \tinygray{0\%} & \tinygray{0\%} & \tinygray{1\%} & \tinygray{2\%} & \tinygray{0\%} & \tinygray{0\%} & \tinygray{2\%} & \tinygray{5\%} & \tinygray{6\%} \\ \hline
\hline
\multirow{12}*{\rotatebox[origin=c]{90}{$a=-0.01$}}
  & \multirow{2}{*}{Supr.} & Cov. & .91 & .92 & .89 & .70 & .66 & .91 & .91 & .86 & .63 & .59 & .91 & .90 & .87 & .69 & .59 \\
  &                            & \textcolor{gray}{Bw.} & \tinygray{.0125} & \tinygray{.0111} & \tinygray{.0100} & \tinygray{.0147} & \tinygray{.0307} & \tinygray{.0133} & \tinygray{.0137} & \tinygray{.0156} & \tinygray{.0300} & \tinygray{.0796} & \tinygray{.0143} & \tinygray{.0175} & \tinygray{.0275} & \tinygray{.0669} & \tinygray{.2207} \\ \cline{2-18}
  & \multirow{2}{*}{LSCI}     & Cov. & .95 & .92 & .76 & .75 & .77 & .96 & .87 & .76 & .67 & .69 & .96 & .89 & .80 & .74 & .66 \\
  &                           & \textcolor{gray}{Bw.} & \tinygray{.0121} & \tinygray{.0074} & \tinygray{.0085} & \tinygray{.0154} & \tinygray{.0342} & \tinygray{.0126} & \tinygray{.0097} & \tinygray{.0145} & \tinygray{.0314} & \tinygray{.0882} & \tinygray{.0133} & \tinygray{.0148} & \tinygray{.0294} & \tinygray{.0792} & \tinygray{.2418} \\ \cline{2-18}
  & \multirow{2}{*}{DtACI} & Cov. & .91 & .97 & .91 & .48 & .43 & .91 & .94 & .80 & .35 & .33 & .91 & .88 & .70 & .39 & .32 \\
  &                            & \textcolor{gray}{Bw.} & \tinygray{.0125} & \tinygray{.0118} & \tinygray{.0099} & \tinygray{.0133} & \tinygray{.0272} & \tinygray{.0133} & \tinygray{.0129} & \tinygray{.0144} & \tinygray{.0266} & \tinygray{.0707} & \tinygray{.0143} & \tinygray{.0145} & \tinygray{.0229} & \tinygray{.0591} & \tinygray{.1976} \\ \cline{2-18}
  & \multirow{3}{*}{\parbox{2cm}{WCP\\Learned}} & Cov. & .90 & .90 & .86 & .70 & .67 & .90 & .89 & .84 & .63 & .59 & .90 & .89 & .86 & .68 & .57 \\
  &                            & \textcolor{gray}{Bw.} & \tinygray{.0117} & \tinygray{.0103} & \tinygray{.0095} & \tinygray{.0147} & \tinygray{.0308} & \tinygray{.0124} & \tinygray{.0129} & \tinygray{.0152} & \tinygray{.0299} & \tinygray{.0797} & \tinygray{.0132} & \tinygray{.0169} & \tinygray{.0270} & \tinygray{.0663} & \tinygray{.2184} \\
  &                            & \textcolor{gray}{$n_\infty$} & \tinygray{0\%} & \tinygray{0\%} & \tinygray{0\%} & \tinygray{0\%} & \tinygray{0\%} & \tinygray{0\%} & \tinygray{0\%} & \tinygray{0\%} & \tinygray{0\%} & \tinygray{0\%} & \tinygray{0\%} & \tinygray{0\%} & \tinygray{0\%} & \tinygray{0\%} & \tinygray{0\%} \\ \cline{2-18}
  & \multirow{3}{*}{\parbox{2cm}{WCP\\Gauss (Ours)}} & Cov. & .92 & .91 & .97 & .97 & .93 & .92 & .91 & .94 & .95 & .95 & .92 & .92 & .90 & .89 & .91 \\
  &                            & \textcolor{gray}{Bw.} & \tinygray{.0128} & \tinygray{.0108} & \tinygray{.0132} & \tinygray{.0183} & \tinygray{.0378} & \tinygray{.0135} & \tinygray{.0137} & \tinygray{.0192} & \tinygray{.0376} & \tinygray{.1030} & \tinygray{.0145} & \tinygray{.0190} & \tinygray{.0301} & \tinygray{.0770} & \tinygray{.2767} \\
  &                            & \textcolor{gray}{$n_\infty$} & \tinygray{0\%} & \tinygray{0\%} & \tinygray{0\%} & \tinygray{2\%} & \tinygray{2\%} & \tinygray{0\%} & \tinygray{0\%} & \tinygray{2\%} & \tinygray{5\%} & \tinygray{5\%} & \tinygray{0\%} & \tinygray{0\%} & \tinygray{7\%} & \tinygray{12\%} & \tinygray{12\%} \\ \hline
\hline
\end{tabular}
\end{table}
\endgroup

\begingroup
\setlength{\tabcolsep}{2pt}
\renewcommand{\arraystretch}{1.2}
\begin{table}[ht]
\caption{Coverage and mean bandwidth for the recalibration task for varying $a, c$ and fixed $b=0.5$. We also report the fraction of samples where infinite bands were reported $n_\infty$. The mean bandwidth is taken over finite bands only. The gray font is chosen for better readability.}
\label{tab:synthetic_re_beta0p5}
\centering\small
\begin{tabular}{lll | ccccc|ccccc|ccccc |}
\hline
& & & \multicolumn{5}{c|}{$c=-0.5$} & \multicolumn{5}{c|}{$c=0.0$} & \multicolumn{5}{c|}{$c=0.5$} \\ \hline
& & T: & 1 & 5 & 10 & 15 & 20 & 1 & 5 & 10 & 15 & 20 & 1 & 5 & 10 & 15 & 20 \\ \hline
\hline
\multirow{12}*{\rotatebox[origin=c]{90}{$a=-0.005$}}
  & \multirow{2}{*}{Supr.} & Cov. & .91 & .92 & .92 & .91 & .91 & .91 & .91 & .91 & .91 & .89 & .91 & .89 & .89 & .89 & .89 \\
  &                            & \textcolor{gray}{Bw.} & \tinygray{.0127} & \tinygray{.0109} & \tinygray{.0090} & \tinygray{.0075} & \tinygray{.0067} & \tinygray{.0134} & \tinygray{.0135} & \tinygray{.0139} & \tinygray{.0147} & \tinygray{.0164} & \tinygray{.0144} & \tinygray{.0172} & \tinygray{.0248} & \tinygray{.0426} & \tinygray{.0809} \\ \cline{2-18}
  & \multirow{2}{*}{LSCI}     & Cov. & .96 & .95 & .93 & .92 & .90 & .97 & .94 & .91 & .89 & .86 & .96 & .95 & .97 & .97 & .97 \\
  &                           & \textcolor{gray}{Bw.} & \tinygray{.0100} & \tinygray{.0067} & \tinygray{.0053} & \tinygray{.0053} & \tinygray{.0056} & \tinygray{.0102} & \tinygray{.0096} & \tinygray{.0095} & \tinygray{.0127} & \tinygray{.0150} & \tinygray{.0108} & \tinygray{.0137} & \tinygray{.0238} & \tinygray{.0505} & \tinygray{.0986} \\ \cline{2-18}
  & \multirow{2}{*}{DtACI} & Cov. & .91 & .97 & .96 & .96 & .95 & .91 & .94 & .93 & .93 & .89 & .91 & .88 & .90 & .91 & .90 \\
  &                            & \textcolor{gray}{Bw.} & \tinygray{.0127} & \tinygray{.0117} & \tinygray{.0096} & \tinygray{.0080} & \tinygray{.0068} & \tinygray{.0134} & \tinygray{.0130} & \tinygray{.0133} & \tinygray{.0139} & \tinygray{.0154} & \tinygray{.0144} & \tinygray{.0146} & \tinygray{.0213} & \tinygray{.0346} & \tinygray{.0638} \\ \cline{2-18}
  & \multirow{3}{*}{\parbox{2cm}{WCP\\Learned}} & Cov. & .90 & .89 & .89 & .89 & .89 & .90 & .89 & .89 & .89 & .87 & .90 & .89 & .88 & .89 & .88 \\
  &                            & \textcolor{gray}{Bw.} & \tinygray{.0117} & \tinygray{.0101} & \tinygray{.0084} & \tinygray{.0072} & \tinygray{.0064} & \tinygray{.0124} & \tinygray{.0127} & \tinygray{.0129} & \tinygray{.0140} & \tinygray{.0157} & \tinygray{.0133} & \tinygray{.0168} & \tinygray{.0238} & \tinygray{.0403} & \tinygray{.0766} \\
  &                            & \textcolor{gray}{$n_\infty$} & \tinygray{0\%} & \tinygray{0\%} & \tinygray{0\%} & \tinygray{0\%} & \tinygray{0\%} & \tinygray{0\%} & \tinygray{0\%} & \tinygray{0\%} & \tinygray{0\%} & \tinygray{0\%} & \tinygray{0\%} & \tinygray{0\%} & \tinygray{0\%} & \tinygray{0\%} & \tinygray{0\%} \\ \cline{2-18}
  & \multirow{3}{*}{\parbox{2cm}{WCP\\Gauss (Ours)}} & Cov. & .91 & .91 & .91 & .91 & .97 & .91 & .91 & .92 & .93 & .95 & .91 & .91 & .91 & .89 & .87 \\
  &                            & \textcolor{gray}{Bw.} & \tinygray{.0129} & \tinygray{.0106} & \tinygray{.0089} & \tinygray{.0074} & \tinygray{.0080} & \tinygray{.0134} & \tinygray{.0137} & \tinygray{.0143} & \tinygray{.0155} & \tinygray{.0199} & \tinygray{.0144} & \tinygray{.0187} & \tinygray{.0276} & \tinygray{.0431} & \tinygray{.0683} \\
  &                            & \textcolor{gray}{$n_\infty$} & \tinygray{0\%} & \tinygray{0\%} & \tinygray{0\%} & \tinygray{0\%} & \tinygray{0\%} & \tinygray{0\%} & \tinygray{0\%} & \tinygray{0\%} & \tinygray{0\%} & \tinygray{0\%} & \tinygray{0\%} & \tinygray{0\%} & \tinygray{0\%} & \tinygray{1\%} & \tinygray{2\%} \\ \hline
\hline
\multirow{12}*{\rotatebox[origin=c]{90}{$a=-0.0075$}}
  & \multirow{2}{*}{Supr.} & Cov. & .91 & .92 & .91 & .87 & .79 & .91 & .91 & .90 & .85 & .72 & .91 & .89 & .88 & .88 & .84 \\
  &                            & \textcolor{gray}{Bw.} & \tinygray{.0127} & \tinygray{.0110} & \tinygray{.0093} & \tinygray{.0092} & \tinygray{.0123} & \tinygray{.0134} & \tinygray{.0137} & \tinygray{.0145} & \tinygray{.0182} & \tinygray{.0316} & \tinygray{.0144} & \tinygray{.0174} & \tinygray{.0256} & \tinygray{.0472} & \tinygray{.1050} \\ \cline{2-18}
  & \multirow{2}{*}{LSCI}     & Cov. & .96 & .94 & .88 & .85 & .81 & .96 & .91 & .86 & .80 & .75 & .97 & .92 & .92 & .87 & .84 \\
  &                           & \textcolor{gray}{Bw.} & \tinygray{.0092} & \tinygray{.0070} & \tinygray{.0068} & \tinygray{.0086} & \tinygray{.0129} & \tinygray{.0098} & \tinygray{.0091} & \tinygray{.0135} & \tinygray{.0177} & \tinygray{.0333} & \tinygray{.0102} & \tinygray{.0129} & \tinygray{.0255} & \tinygray{.0580} & \tinygray{.1262} \\ \cline{2-18}
  & \multirow{2}{*}{DtACI} & Cov. & .91 & .96 & .94 & .87 & .67 & .91 & .93 & .91 & .74 & .49 & .91 & .87 & .89 & .80 & .67 \\
  &                            & \textcolor{gray}{Bw.} & \tinygray{.0127} & \tinygray{.0118} & \tinygray{.0098} & \tinygray{.0090} & \tinygray{.0112} & \tinygray{.0134} & \tinygray{.0131} & \tinygray{.0138} & \tinygray{.0168} & \tinygray{.0282} & \tinygray{.0144} & \tinygray{.0147} & \tinygray{.0219} & \tinygray{.0396} & \tinygray{.0908} \\ \cline{2-18}
  & \multirow{3}{*}{\parbox{2cm}{WCP\\Learned}} & Cov. & .90 & .89 & .89 & .86 & .78 & .90 & .89 & .88 & .83 & .71 & .90 & .89 & .88 & .87 & .81 \\
  &                            & \textcolor{gray}{Bw.} & \tinygray{.0118} & \tinygray{.0102} & \tinygray{.0088} & \tinygray{.0090} & \tinygray{.0122} & \tinygray{.0125} & \tinygray{.0128} & \tinygray{.0136} & \tinygray{.0179} & \tinygray{.0314} & \tinygray{.0133} & \tinygray{.0170} & \tinygray{.0244} & \tinygray{.0456} & \tinygray{.1014} \\
  &                            & \textcolor{gray}{$n_\infty$} & \tinygray{0\%} & \tinygray{0\%} & \tinygray{0\%} & \tinygray{0\%} & \tinygray{0\%} & \tinygray{0\%} & \tinygray{0\%} & \tinygray{0\%} & \tinygray{0\%} & \tinygray{0\%} & \tinygray{0\%} & \tinygray{0\%} & \tinygray{0\%} & \tinygray{0\%} & \tinygray{0\%} \\ \cline{2-18}
  & \multirow{3}{*}{\parbox{2cm}{WCP\\Gauss (Ours)}} & Cov. & .91 & .91 & .94 & .97 & .97 & .91 & .91 & .93 & .96 & .99 & .91 & .91 & .90 & .92 & .92 \\
  &                            & \textcolor{gray}{Bw.} & \tinygray{.0128} & \tinygray{.0108} & \tinygray{.0102} & \tinygray{.0113} & \tinygray{.0153} & \tinygray{.0134} & \tinygray{.0139} & \tinygray{.0163} & \tinygray{.0231} & \tinygray{.0436} & \tinygray{.0143} & \tinygray{.0188} & \tinygray{.0274} & \tinygray{.0559} & \tinygray{.1276} \\
  &                            & \textcolor{gray}{$n_\infty$} & \tinygray{0\%} & \tinygray{0\%} & \tinygray{0\%} & \tinygray{0\%} & \tinygray{0\%} & \tinygray{0\%} & \tinygray{0\%} & \tinygray{0\%} & \tinygray{1\%} & \tinygray{2\%} & \tinygray{0\%} & \tinygray{0\%} & \tinygray{2\%} & \tinygray{5\%} & \tinygray{6\%} \\ \hline
\hline
\multirow{12}*{\rotatebox[origin=c]{90}{$a=-0.01$}}
  & \multirow{2}{*}{Supr.} & Cov. & .91 & .91 & .87 & .74 & .69 & .91 & .90 & .84 & .64 & .61 & .91 & .89 & .85 & .70 & .59 \\
  &                            & \textcolor{gray}{Bw.} & \tinygray{.0127} & \tinygray{.0112} & \tinygray{.0103} & \tinygray{.0155} & \tinygray{.0319} & \tinygray{.0135} & \tinygray{.0138} & \tinygray{.0161} & \tinygray{.0313} & \tinygray{.0825} & \tinygray{.0144} & \tinygray{.0175} & \tinygray{.0277} & \tinygray{.0699} & \tinygray{.2270} \\ \cline{2-18}
  & \multirow{2}{*}{LSCI}     & Cov. & .95 & .90 & .79 & .77 & .75 & .96 & .86 & .69 & .68 & .70 & .96 & .86 & .81 & .72 & .67 \\
  &                           & \textcolor{gray}{Bw.} & \tinygray{.0088} & \tinygray{.0071} & \tinygray{.0091} & \tinygray{.0163} & \tinygray{.0349} & \tinygray{.0090} & \tinygray{.0094} & \tinygray{.0144} & \tinygray{.0331} & \tinygray{.0907} & \tinygray{.0093} & \tinygray{.0124} & \tinygray{.0305} & \tinygray{.0802} & \tinygray{.2482} \\ \cline{2-18}
  & \multirow{2}{*}{DtACI} & Cov. & .91 & .96 & .89 & .47 & .44 & .91 & .93 & .78 & .36 & .34 & .91 & .86 & .69 & .39 & .32 \\
  &                            & \textcolor{gray}{Bw.} & \tinygray{.0127} & \tinygray{.0118} & \tinygray{.0104} & \tinygray{.0138} & \tinygray{.0282} & \tinygray{.0135} & \tinygray{.0131} & \tinygray{.0150} & \tinygray{.0277} & \tinygray{.0731} & \tinygray{.0144} & \tinygray{.0149} & \tinygray{.0241} & \tinygray{.0616} & \tinygray{.2033} \\ \cline{2-18}
  & \multirow{3}{*}{\parbox{2cm}{WCP\\Learned}} & Cov. & .90 & .89 & .84 & .73 & .69 & .90 & .89 & .81 & .64 & .61 & .89 & .89 & .84 & .67 & .58 \\
  &                            & \textcolor{gray}{Bw.} & \tinygray{.0118} & \tinygray{.0104} & \tinygray{.0100} & \tinygray{.0155} & \tinygray{.0319} & \tinygray{.0125} & \tinygray{.0132} & \tinygray{.0155} & \tinygray{.0312} & \tinygray{.0825} & \tinygray{.0134} & \tinygray{.0172} & \tinygray{.0271} & \tinygray{.0685} & \tinygray{.2266} \\
  &                            & \textcolor{gray}{$n_\infty$} & \tinygray{0\%} & \tinygray{0\%} & \tinygray{0\%} & \tinygray{0\%} & \tinygray{0\%} & \tinygray{0\%} & \tinygray{0\%} & \tinygray{0\%} & \tinygray{0\%} & \tinygray{0\%} & \tinygray{0\%} & \tinygray{0\%} & \tinygray{0\%} & \tinygray{0\%} & \tinygray{0\%} \\ \cline{2-18}
  & \multirow{3}{*}{\parbox{2cm}{WCP\\Gauss (Ours)}} & Cov. & .91 & .91 & .98 & .97 & .96 & .91 & .91 & .97 & .97 & .99 & .91 & .91 & .93 & .95 & .93 \\
  &                            & \textcolor{gray}{Bw.} & \tinygray{.0127} & \tinygray{.0110} & \tinygray{.0149} & \tinygray{.0202} & \tinygray{.0403} & \tinygray{.0134} & \tinygray{.0142} & \tinygray{.0235} & \tinygray{.0405} & \tinygray{.1204} & \tinygray{.0141} & \tinygray{.0189} & \tinygray{.0358} & \tinygray{.0997} & \tinygray{.2914} \\
  &                            & \textcolor{gray}{$n_\infty$} & \tinygray{0\%} & \tinygray{0\%} & \tinygray{0\%} & \tinygray{1\%} & \tinygray{2\%} & \tinygray{0\%} & \tinygray{0\%} & \tinygray{2\%} & \tinygray{4\%} & \tinygray{5\%} & \tinygray{0\%} & \tinygray{0\%} & \tinygray{7\%} & \tinygray{12\%} & \tinygray{13\%} \\ \hline
\hline
\end{tabular}
\end{table}
\endgroup

\begingroup
\setlength{\tabcolsep}{2pt}
\renewcommand{\arraystretch}{1.2}
\begin{table}[ht]
\caption{Coverage and mean bandwidth for the general task (no recalibration) for varying $a, c$ and fixed $b=-0.5$. We also report the fraction of samples where infinite bands were reported $n_\infty$. The mean bandwidth is taken over finite bands only. A dash (--) marks cells where every band was infinite. The gray font is chosen for better readability.}
\label{tab:synthetic_beta-0p5}
\centering\small
\begin{tabular}{lll | ccccc|ccccc|ccccc |}
\hline
& & & \multicolumn{5}{c|}{$c=-0.5$} & \multicolumn{5}{c|}{$c=0.0$} & \multicolumn{5}{c|}{$c=0.5$} \\ \hline
& & T: & 1 & 5 & 10 & 15 & 20 & 1 & 5 & 10 & 15 & 20 & 1 & 5 & 10 & 15 & 20 \\ \hline
\hline
\multirow{10}*{\rotatebox[origin=c]{90}{$a=-0.005$}}
  & \multirow{2}{*}{Supr.} & Cov. & .91 & .93 & .95 & .97 & .98 & .91 & .91 & .91 & .90 & .90 & .91 & .89 & .85 & .78 & .69 \\
  &                            & \textcolor{gray}{Bw.} & \tinygray{.0129} & \tinygray{.0129} & \tinygray{.0129} & \tinygray{.0129} & \tinygray{.0129} & \tinygray{.0132} & \tinygray{.0132} & \tinygray{.0132} & \tinygray{.0132} & \tinygray{.0132} & \tinygray{.0136} & \tinygray{.0136} & \tinygray{.0136} & \tinygray{.0136} & \tinygray{.0136} \\ \cline{2-18}
  & \multirow{2}{*}{LSCI}     & Cov. & .96 & .98 & .99 & .99 & .99 & .96 & .95 & .92 & .87 & .72 & .96 & .90 & .72 & .42 & .11 \\
  &                           & \textcolor{gray}{Bw.} & \tinygray{.0122} & \tinygray{.0122} & \tinygray{.0123} & \tinygray{.0123} & \tinygray{.0123} & \tinygray{.0124} & \tinygray{.0124} & \tinygray{.0124} & \tinygray{.0124} & \tinygray{.0124} & \tinygray{.0127} & \tinygray{.0127} & \tinygray{.0126} & \tinygray{.0125} & \tinygray{.0124} \\ \cline{2-18}
  & \multirow{3}{*}{\parbox{2cm}{WCP\\Learned}} & Cov. & .90 & .90 & .90 & .91 & .91 & .90 & .90 & .89 & .89 & .87 & .90 & .89 & .87 & .84 & .82 \\
  &                            & \textcolor{gray}{Bw.} & \tinygray{.0119} & \tinygray{.0111} & \tinygray{.0103} & \tinygray{.0095} & \tinygray{.0087} & \tinygray{.0122} & \tinygray{.0123} & \tinygray{.0122} & \tinygray{.0124} & \tinygray{.0122} & \tinygray{.0127} & \tinygray{.0136} & \tinygray{.0153} & \tinygray{.0165} & \tinygray{.0184} \\
  &                            & \textcolor{gray}{$n_\infty$} & \tinygray{0\%} & \tinygray{0\%} & \tinygray{0\%} & \tinygray{0\%} & \tinygray{0\%} & \tinygray{0\%} & \tinygray{0\%} & \tinygray{0\%} & \tinygray{0\%} & \tinygray{0\%} & \tinygray{0\%} & \tinygray{0\%} & \tinygray{0\%} & \tinygray{0\%} & \tinygray{0\%} \\ \cline{2-18}
  & \multirow{3}{*}{\parbox{2cm}{WCP\\Gauss (Ours)}} & Cov. & .91 & .91 & .91 & .90 & .88 & .91 & .91 & .91 & .91 & .93 & .91 & .92 & .93 & .99 & 1.00 \\
  &                            & \textcolor{gray}{Bw.} & \tinygray{.0129} & \tinygray{.0115} & \tinygray{.0104} & \tinygray{.0093} & \tinygray{.0083} & \tinygray{.0133} & \tinygray{.0133} & \tinygray{.0133} & \tinygray{.0136} & \tinygray{.0148} & \tinygray{.0137} & \tinygray{.0164} & \tinygray{.0191} & \tinygray{.0231} & \tinygray{.0264} \\
  &                            & \textcolor{gray}{$n_\infty$} & \tinygray{0\%} & \tinygray{0\%} & \tinygray{0\%} & \tinygray{0\%} & \tinygray{0\%} & \tinygray{0\%} & \tinygray{0\%} & \tinygray{0\%} & \tinygray{1\%} & \tinygray{11\%} & \tinygray{0\%} & \tinygray{0\%} & \tinygray{20\%} & \tinygray{85\%} & \tinygray{$>$99\%} \\ \hline
\hline
\multirow{10}*{\rotatebox[origin=c]{90}{$a=-0.0075$}}
  & \multirow{2}{*}{Supr.} & Cov. & .91 & .93 & .95 & .97 & .98 & .91 & .91 & .91 & .90 & .88 & .91 & .89 & .84 & .75 & .55 \\
  &                            & \textcolor{gray}{Bw.} & \tinygray{.0128} & \tinygray{.0128} & \tinygray{.0128} & \tinygray{.0128} & \tinygray{.0128} & \tinygray{.0132} & \tinygray{.0132} & \tinygray{.0132} & \tinygray{.0132} & \tinygray{.0132} & \tinygray{.0137} & \tinygray{.0137} & \tinygray{.0137} & \tinygray{.0137} & \tinygray{.0137} \\ \cline{2-18}
  & \multirow{2}{*}{LSCI}     & Cov. & .96 & .98 & .97 & .97 & .90 & .96 & .94 & .86 & .65 & .27 & .96 & .90 & .59 & .17 & .01 \\
  &                           & \textcolor{gray}{Bw.} & \tinygray{.0122} & \tinygray{.0123} & \tinygray{.0123} & \tinygray{.0123} & \tinygray{.0123} & \tinygray{.0125} & \tinygray{.0125} & \tinygray{.0125} & \tinygray{.0125} & \tinygray{.0125} & \tinygray{.0128} & \tinygray{.0127} & \tinygray{.0126} & \tinygray{.0125} & \tinygray{.0124} \\ \cline{2-18}
  & \multirow{3}{*}{\parbox{2cm}{WCP\\Learned}} & Cov. & .90 & .90 & .90 & .90 & .89 & .90 & .90 & .88 & .88 & .85 & .90 & .89 & .87 & .84 & .79 \\
  &                            & \textcolor{gray}{Bw.} & \tinygray{.0119} & \tinygray{.0111} & \tinygray{.0104} & \tinygray{.0095} & \tinygray{.0087} & \tinygray{.0123} & \tinygray{.0124} & \tinygray{.0123} & \tinygray{.0125} & \tinygray{.0122} & \tinygray{.0127} & \tinygray{.0137} & \tinygray{.0154} & \tinygray{.0165} & \tinygray{.0180} \\
  &                            & \textcolor{gray}{$n_\infty$} & \tinygray{0\%} & \tinygray{0\%} & \tinygray{0\%} & \tinygray{0\%} & \tinygray{0\%} & \tinygray{0\%} & \tinygray{0\%} & \tinygray{0\%} & \tinygray{0\%} & \tinygray{0\%} & \tinygray{0\%} & \tinygray{0\%} & \tinygray{0\%} & \tinygray{0\%} & \tinygray{0\%} \\ \cline{2-18}
  & \multirow{3}{*}{\parbox{2cm}{WCP\\Gauss (Ours)}} & Cov. & .91 & .91 & .90 & .89 & .88 & .91 & .91 & .91 & .94 & 1.00 & .91 & .92 & .97 & 1.00 & 1.00 \\
  &                            & \textcolor{gray}{Bw.} & \tinygray{.0129} & \tinygray{.0115} & \tinygray{.0104} & \tinygray{.0093} & \tinygray{.0086} & \tinygray{.0133} & \tinygray{.0134} & \tinygray{.0136} & \tinygray{.0157} & \tinygray{.0201} & \tinygray{.0138} & \tinygray{.0165} & \tinygray{.0204} & \tinygray{.0252} & \tinygray{--} \\
  &                            & \textcolor{gray}{$n_\infty$} & \tinygray{0\%} & \tinygray{0\%} & \tinygray{0\%} & \tinygray{0\%} & \tinygray{0\%} & \tinygray{0\%} & \tinygray{0\%} & \tinygray{0\%} & \tinygray{25\%} & \tinygray{94\%} & \tinygray{0\%} & \tinygray{0\%} & \tinygray{45\%} & \tinygray{99\%} & \tinygray{100\%} \\ \hline
\hline
\multirow{10}*{\rotatebox[origin=c]{90}{$a=-0.01$}}
  & \multirow{2}{*}{Supr.} & Cov. & .91 & .93 & .95 & .96 & .98 & .91 & .91 & .90 & .88 & .73 & .91 & .88 & .83 & .67 & .06 \\
  &                            & \textcolor{gray}{Bw.} & \tinygray{.0129} & \tinygray{.0129} & \tinygray{.0129} & \tinygray{.0129} & \tinygray{.0129} & \tinygray{.0132} & \tinygray{.0132} & \tinygray{.0132} & \tinygray{.0132} & \tinygray{.0132} & \tinygray{.0137} & \tinygray{.0137} & \tinygray{.0137} & \tinygray{.0137} & \tinygray{.0137} \\ \cline{2-18}
  & \multirow{2}{*}{LSCI}     & Cov. & .95 & .96 & .93 & .79 & .41 & .95 & .93 & .75 & .26 & .01 & .96 & .86 & .39 & .02 & .00 \\
  &                           & \textcolor{gray}{Bw.} & \tinygray{.0122} & \tinygray{.0123} & \tinygray{.0123} & \tinygray{.0123} & \tinygray{.0123} & \tinygray{.0125} & \tinygray{.0125} & \tinygray{.0125} & \tinygray{.0125} & \tinygray{.0125} & \tinygray{.0128} & \tinygray{.0127} & \tinygray{.0126} & \tinygray{.0126} & \tinygray{.0124} \\ \cline{2-18}
  & \multirow{3}{*}{\parbox{2cm}{WCP\\Learned}} & Cov. & .90 & .90 & .89 & .88 & .80 & .90 & .89 & .88 & .85 & .65 & .90 & .88 & .86 & .82 & .47 \\
  &                            & \textcolor{gray}{Bw.} & \tinygray{.0119} & \tinygray{.0111} & \tinygray{.0103} & \tinygray{.0095} & \tinygray{.0087} & \tinygray{.0123} & \tinygray{.0123} & \tinygray{.0124} & \tinygray{.0124} & \tinygray{.0125} & \tinygray{.0127} & \tinygray{.0136} & \tinygray{.0153} & \tinygray{.0167} & \tinygray{.0183} \\
  &                            & \textcolor{gray}{$n_\infty$} & \tinygray{0\%} & \tinygray{0\%} & \tinygray{0\%} & \tinygray{0\%} & \tinygray{0\%} & \tinygray{0\%} & \tinygray{0\%} & \tinygray{0\%} & \tinygray{0\%} & \tinygray{0\%} & \tinygray{0\%} & \tinygray{0\%} & \tinygray{0\%} & \tinygray{0\%} & \tinygray{0\%} \\ \cline{2-18}
  & \multirow{3}{*}{\parbox{2cm}{WCP\\Gauss (Ours)}} & Cov. & .91 & .91 & .90 & .89 & .97 & .91 & .91 & .93 & 1.00 & 1.00 & .91 & .92 & .99 & 1.00 & 1.00 \\
  &                            & \textcolor{gray}{Bw.} & \tinygray{.0129} & \tinygray{.0116} & \tinygray{.0105} & \tinygray{.0099} & \tinygray{.0100} & \tinygray{.0133} & \tinygray{.0134} & \tinygray{.0146} & \tinygray{.0182} & \tinygray{--} & \tinygray{.0138} & \tinygray{.0167} & \tinygray{.0217} & \tinygray{--} & \tinygray{--} \\
  &                            & \textcolor{gray}{$n_\infty$} & \tinygray{0\%} & \tinygray{0\%} & \tinygray{0\%} & \tinygray{4\%} & \tinygray{79\%} & \tinygray{0\%} & \tinygray{0\%} & \tinygray{7\%} & \tinygray{92\%} & \tinygray{100\%} & \tinygray{0\%} & \tinygray{1\%} & \tinygray{76\%} & \tinygray{100\%} & \tinygray{100\%} \\ \hline
\hline
\end{tabular}
\end{table}
\endgroup

\begingroup
\setlength{\tabcolsep}{2pt}
\renewcommand{\arraystretch}{1.2}
\begin{table}[ht]
\caption{Coverage and mean bandwidth for the general task (no recalibration) for varying $a, c$ and fixed $b=0.0$. We also report the fraction of samples where infinite bands were reported $n_\infty$. The mean bandwidth is taken over finite bands only. A dash (--) marks cells where every band was infinite. The gray font is chosen for better readability.}
\label{tab:synthetic_beta0p0}
\centering\small
\begin{tabular}{lll | ccccc|ccccc|ccccc |}
\hline
& & & \multicolumn{5}{c|}{$c=-0.5$} & \multicolumn{5}{c|}{$c=0.0$} & \multicolumn{5}{c|}{$c=0.5$} \\ \hline
& & T: & 1 & 5 & 10 & 15 & 20 & 1 & 5 & 10 & 15 & 20 & 1 & 5 & 10 & 15 & 20 \\ \hline
\hline
\multirow{10}*{\rotatebox[origin=c]{90}{$a=-0.005$}}
  & \multirow{2}{*}{Supr.} & Cov. & .91 & .93 & .95 & .97 & .98 & .91 & .91 & .91 & .91 & .90 & .91 & .89 & .83 & .79 & .70 \\
  &                            & \textcolor{gray}{Bw.} & \tinygray{.0128} & \tinygray{.0128} & \tinygray{.0128} & \tinygray{.0128} & \tinygray{.0128} & \tinygray{.0132} & \tinygray{.0132} & \tinygray{.0132} & \tinygray{.0132} & \tinygray{.0132} & \tinygray{.0136} & \tinygray{.0136} & \tinygray{.0136} & \tinygray{.0136} & \tinygray{.0136} \\ \cline{2-18}
  & \multirow{2}{*}{LSCI}     & Cov. & .96 & .98 & .98 & .99 & 1.00 & .95 & .94 & .91 & .86 & .75 & .96 & .90 & .73 & .45 & .13 \\
  &                           & \textcolor{gray}{Bw.} & \tinygray{.0121} & \tinygray{.0121} & \tinygray{.0122} & \tinygray{.0122} & \tinygray{.0122} & \tinygray{.0123} & \tinygray{.0123} & \tinygray{.0123} & \tinygray{.0123} & \tinygray{.0123} & \tinygray{.0127} & \tinygray{.0126} & \tinygray{.0125} & \tinygray{.0124} & \tinygray{.0123} \\ \cline{2-18}
  & \multirow{3}{*}{\parbox{2cm}{WCP\\Learned}} & Cov. & .90 & .89 & .90 & .91 & .91 & .90 & .90 & .89 & .89 & .88 & .90 & .89 & .88 & .85 & .82 \\
  &                            & \textcolor{gray}{Bw.} & \tinygray{.0118} & \tinygray{.0110} & \tinygray{.0102} & \tinygray{.0095} & \tinygray{.0087} & \tinygray{.0122} & \tinygray{.0122} & \tinygray{.0123} & \tinygray{.0123} & \tinygray{.0122} & \tinygray{.0127} & \tinygray{.0138} & \tinygray{.0151} & \tinygray{.0167} & \tinygray{.0178} \\
  &                            & \textcolor{gray}{$n_\infty$} & \tinygray{0\%} & \tinygray{0\%} & \tinygray{0\%} & \tinygray{0\%} & \tinygray{0\%} & \tinygray{0\%} & \tinygray{0\%} & \tinygray{0\%} & \tinygray{0\%} & \tinygray{0\%} & \tinygray{0\%} & \tinygray{0\%} & \tinygray{0\%} & \tinygray{0\%} & \tinygray{0\%} \\ \cline{2-18}
  & \multirow{3}{*}{\parbox{2cm}{WCP\\Gauss (Ours)}} & Cov. & .92 & .90 & .91 & .90 & .90 & .91 & .91 & .91 & .92 & .93 & .92 & .92 & .94 & .99 & 1.00 \\
  &                            & \textcolor{gray}{Bw.} & \tinygray{.0129} & \tinygray{.0112} & \tinygray{.0103} & \tinygray{.0091} & \tinygray{.0083} & \tinygray{.0133} & \tinygray{.0133} & \tinygray{.0134} & \tinygray{.0138} & \tinygray{.0151} & \tinygray{.0138} & \tinygray{.0164} & \tinygray{.0196} & \tinygray{.0250} & \tinygray{.0275} \\
  &                            & \textcolor{gray}{$n_\infty$} & \tinygray{0\%} & \tinygray{0\%} & \tinygray{0\%} & \tinygray{0\%} & \tinygray{0\%} & \tinygray{0\%} & \tinygray{0\%} & \tinygray{0\%} & \tinygray{0\%} & \tinygray{11\%} & \tinygray{0\%} & \tinygray{0\%} & \tinygray{20\%} & \tinygray{85\%} & \tinygray{$>$99\%} \\ \hline
\hline
\multirow{10}*{\rotatebox[origin=c]{90}{$a=-0.0075$}}
  & \multirow{2}{*}{Supr.} & Cov. & .91 & .93 & .95 & .97 & .98 & .91 & .91 & .91 & .90 & .88 & .91 & .89 & .83 & .76 & .56 \\
  &                            & \textcolor{gray}{Bw.} & \tinygray{.0128} & \tinygray{.0128} & \tinygray{.0128} & \tinygray{.0128} & \tinygray{.0128} & \tinygray{.0132} & \tinygray{.0132} & \tinygray{.0132} & \tinygray{.0132} & \tinygray{.0132} & \tinygray{.0136} & \tinygray{.0136} & \tinygray{.0136} & \tinygray{.0136} & \tinygray{.0136} \\ \cline{2-18}
  & \multirow{2}{*}{LSCI}     & Cov. & .96 & .97 & .97 & .95 & .92 & .96 & .93 & .86 & .63 & .27 & .96 & .88 & .58 & .16 & .01 \\
  &                           & \textcolor{gray}{Bw.} & \tinygray{.0122} & \tinygray{.0122} & \tinygray{.0122} & \tinygray{.0123} & \tinygray{.0123} & \tinygray{.0124} & \tinygray{.0124} & \tinygray{.0124} & \tinygray{.0124} & \tinygray{.0124} & \tinygray{.0127} & \tinygray{.0126} & \tinygray{.0125} & \tinygray{.0125} & \tinygray{.0123} \\ \cline{2-18}
  & \multirow{3}{*}{\parbox{2cm}{WCP\\Learned}} & Cov. & .90 & .90 & .90 & .91 & .91 & .90 & .90 & .88 & .88 & .86 & .90 & .89 & .88 & .83 & .80 \\
  &                            & \textcolor{gray}{Bw.} & \tinygray{.0118} & \tinygray{.0111} & \tinygray{.0102} & \tinygray{.0095} & \tinygray{.0087} & \tinygray{.0122} & \tinygray{.0122} & \tinygray{.0122} & \tinygray{.0123} & \tinygray{.0123} & \tinygray{.0127} & \tinygray{.0139} & \tinygray{.0153} & \tinygray{.0164} & \tinygray{.0184} \\
  &                            & \textcolor{gray}{$n_\infty$} & \tinygray{0\%} & \tinygray{0\%} & \tinygray{0\%} & \tinygray{0\%} & \tinygray{0\%} & \tinygray{0\%} & \tinygray{0\%} & \tinygray{0\%} & \tinygray{0\%} & \tinygray{0\%} & \tinygray{0\%} & \tinygray{0\%} & \tinygray{0\%} & \tinygray{0\%} & \tinygray{0\%} \\ \cline{2-18}
  & \multirow{3}{*}{\parbox{2cm}{WCP\\Gauss (Ours)}} & Cov. & .92 & .90 & .91 & .89 & .89 & .91 & .91 & .92 & .95 & 1.00 & .92 & .92 & .97 & 1.00 & 1.00 \\
  &                            & \textcolor{gray}{Bw.} & \tinygray{.0129} & \tinygray{.0113} & \tinygray{.0103} & \tinygray{.0092} & \tinygray{.0085} & \tinygray{.0133} & \tinygray{.0133} & \tinygray{.0137} & \tinygray{.0160} & \tinygray{.0188} & \tinygray{.0138} & \tinygray{.0165} & \tinygray{.0207} & \tinygray{.0240} & \tinygray{--} \\
  &                            & \textcolor{gray}{$n_\infty$} & \tinygray{0\%} & \tinygray{0\%} & \tinygray{0\%} & \tinygray{0\%} & \tinygray{0\%} & \tinygray{0\%} & \tinygray{0\%} & \tinygray{0\%} & \tinygray{25\%} & \tinygray{94\%} & \tinygray{0\%} & \tinygray{0\%} & \tinygray{45\%} & \tinygray{99\%} & \tinygray{100\%} \\ \hline
\hline
\multirow{10}*{\rotatebox[origin=c]{90}{$a=-0.01$}}
  & \multirow{2}{*}{Supr.} & Cov. & .91 & .93 & .95 & .96 & .97 & .91 & .91 & .90 & .88 & .76 & .91 & .89 & .82 & .68 & .09 \\
  &                            & \textcolor{gray}{Bw.} & \tinygray{.0128} & \tinygray{.0128} & \tinygray{.0128} & \tinygray{.0128} & \tinygray{.0128} & \tinygray{.0132} & \tinygray{.0132} & \tinygray{.0132} & \tinygray{.0132} & \tinygray{.0132} & \tinygray{.0136} & \tinygray{.0136} & \tinygray{.0136} & \tinygray{.0136} & \tinygray{.0136} \\ \cline{2-18}
  & \multirow{2}{*}{LSCI}     & Cov. & .96 & .96 & .92 & .77 & .42 & .96 & .92 & .71 & .26 & .01 & .96 & .85 & .40 & .03 & .00 \\
  &                           & \textcolor{gray}{Bw.} & \tinygray{.0122} & \tinygray{.0122} & \tinygray{.0123} & \tinygray{.0123} & \tinygray{.0123} & \tinygray{.0124} & \tinygray{.0124} & \tinygray{.0124} & \tinygray{.0124} & \tinygray{.0124} & \tinygray{.0127} & \tinygray{.0127} & \tinygray{.0126} & \tinygray{.0125} & \tinygray{.0124} \\ \cline{2-18}
  & \multirow{3}{*}{\parbox{2cm}{WCP\\Learned}} & Cov. & .90 & .89 & .89 & .90 & .84 & .90 & .89 & .88 & .86 & .71 & .90 & .89 & .87 & .82 & .52 \\
  &                            & \textcolor{gray}{Bw.} & \tinygray{.0119} & \tinygray{.0111} & \tinygray{.0102} & \tinygray{.0095} & \tinygray{.0088} & \tinygray{.0122} & \tinygray{.0122} & \tinygray{.0123} & \tinygray{.0123} & \tinygray{.0125} & \tinygray{.0127} & \tinygray{.0139} & \tinygray{.0154} & \tinygray{.0164} & \tinygray{.0185} \\
  &                            & \textcolor{gray}{$n_\infty$} & \tinygray{0\%} & \tinygray{0\%} & \tinygray{0\%} & \tinygray{0\%} & \tinygray{0\%} & \tinygray{0\%} & \tinygray{0\%} & \tinygray{0\%} & \tinygray{0\%} & \tinygray{0\%} & \tinygray{0\%} & \tinygray{0\%} & \tinygray{0\%} & \tinygray{0\%} & \tinygray{0\%} \\ \cline{2-18}
  & \multirow{3}{*}{\parbox{2cm}{WCP\\Gauss (Ours)}} & Cov. & .91 & .90 & .90 & .90 & .98 & .91 & .91 & .92 & 1.00 & 1.00 & .91 & .92 & .99 & 1.00 & 1.00 \\
  &                            & \textcolor{gray}{Bw.} & \tinygray{.0129} & \tinygray{.0114} & \tinygray{.0104} & \tinygray{.0097} & \tinygray{.0099} & \tinygray{.0133} & \tinygray{.0134} & \tinygray{.0147} & \tinygray{.0194} & \tinygray{--} & \tinygray{.0138} & \tinygray{.0166} & \tinygray{.0228} & \tinygray{--} & \tinygray{--} \\
  &                            & \textcolor{gray}{$n_\infty$} & \tinygray{0\%} & \tinygray{0\%} & \tinygray{0\%} & \tinygray{4\%} & \tinygray{79\%} & \tinygray{0\%} & \tinygray{0\%} & \tinygray{7\%} & \tinygray{91\%} & \tinygray{100\%} & \tinygray{0\%} & \tinygray{1\%} & \tinygray{76\%} & \tinygray{100\%} & \tinygray{100\%} \\ \hline
\hline
\end{tabular}
\end{table}
\endgroup

\begingroup
\setlength{\tabcolsep}{2pt}
\renewcommand{\arraystretch}{1.2}
\begin{table}[ht]
\caption{Coverage and mean bandwidth for the general task (no recalibration) for varying $a, c$ and fixed $b=0.5$. We also report the fraction of samples where infinite bands were reported $n_\infty$. The mean bandwidth is taken over finite bands only. A dash (--) marks cells where every band was infinite. The gray font is chosen for better readability.}
\label{tab:synthetic_beta0p5}
\centering\small
\begin{tabular}{lll | ccccc|ccccc|ccccc |}
\hline
& & & \multicolumn{5}{c|}{$c=-0.5$} & \multicolumn{5}{c|}{$c=0.0$} & \multicolumn{5}{c|}{$c=0.5$} \\ \hline
& & T: & 1 & 5 & 10 & 15 & 20 & 1 & 5 & 10 & 15 & 20 & 1 & 5 & 10 & 15 & 20 \\ \hline
\hline
\multirow{10}*{\rotatebox[origin=c]{90}{$a=-0.005$}}
  & \multirow{2}{*}{Supr.} & Cov. & .91 & .93 & .95 & .97 & .98 & .91 & .91 & .91 & .91 & .90 & .91 & .89 & .84 & .77 & .68 \\
  &                            & \textcolor{gray}{Bw.} & \tinygray{.0129} & \tinygray{.0129} & \tinygray{.0129} & \tinygray{.0129} & \tinygray{.0129} & \tinygray{.0133} & \tinygray{.0133} & \tinygray{.0133} & \tinygray{.0133} & \tinygray{.0133} & \tinygray{.0137} & \tinygray{.0137} & \tinygray{.0137} & \tinygray{.0137} & \tinygray{.0137} \\ \cline{2-18}
  & \multirow{2}{*}{LSCI}     & Cov. & .96 & .98 & .98 & 1.00 & 1.00 & .97 & .95 & .91 & .84 & .74 & .96 & .90 & .73 & .41 & .15 \\
  &                           & \textcolor{gray}{Bw.} & \tinygray{.0122} & \tinygray{.0122} & \tinygray{.0122} & \tinygray{.0123} & \tinygray{.0123} & \tinygray{.0124} & \tinygray{.0124} & \tinygray{.0124} & \tinygray{.0124} & \tinygray{.0124} & \tinygray{.0127} & \tinygray{.0126} & \tinygray{.0125} & \tinygray{.0124} & \tinygray{.0123} \\ \cline{2-18}
  & \multirow{3}{*}{\parbox{2cm}{WCP\\Learned}} & Cov. & .90 & .90 & .90 & .90 & .91 & .90 & .89 & .88 & .88 & .87 & .90 & .89 & .87 & .85 & .82 \\
  &                            & \textcolor{gray}{Bw.} & \tinygray{.0119} & \tinygray{.0109} & \tinygray{.0103} & \tinygray{.0095} & \tinygray{.0087} & \tinygray{.0123} & \tinygray{.0122} & \tinygray{.0122} & \tinygray{.0123} & \tinygray{.0122} & \tinygray{.0127} & \tinygray{.0140} & \tinygray{.0149} & \tinygray{.0167} & \tinygray{.0182} \\
  &                            & \textcolor{gray}{$n_\infty$} & \tinygray{0\%} & \tinygray{0\%} & \tinygray{0\%} & \tinygray{0\%} & \tinygray{0\%} & \tinygray{0\%} & \tinygray{0\%} & \tinygray{0\%} & \tinygray{0\%} & \tinygray{0\%} & \tinygray{0\%} & \tinygray{0\%} & \tinygray{0\%} & \tinygray{0\%} & \tinygray{0\%} \\ \cline{2-18}
  & \multirow{3}{*}{\parbox{2cm}{WCP\\Gauss (Ours)}} & Cov. & .91 & .91 & .90 & .89 & .89 & .91 & .91 & .91 & .91 & .93 & .91 & .92 & .94 & .99 & 1.00 \\
  &                            & \textcolor{gray}{Bw.} & \tinygray{.0130} & \tinygray{.0113} & \tinygray{.0102} & \tinygray{.0092} & \tinygray{.0084} & \tinygray{.0134} & \tinygray{.0134} & \tinygray{.0135} & \tinygray{.0138} & \tinygray{.0149} & \tinygray{.0138} & \tinygray{.0158} & \tinygray{.0196} & \tinygray{.0232} & \tinygray{.0241} \\
  &                            & \textcolor{gray}{$n_\infty$} & \tinygray{0\%} & \tinygray{0\%} & \tinygray{0\%} & \tinygray{0\%} & \tinygray{0\%} & \tinygray{0\%} & \tinygray{0\%} & \tinygray{0\%} & \tinygray{0\%} & \tinygray{11\%} & \tinygray{0\%} & \tinygray{0\%} & \tinygray{20\%} & \tinygray{86\%} & \tinygray{$>$99\%} \\ \hline
\hline
\multirow{10}*{\rotatebox[origin=c]{90}{$a=-0.0075$}}
  & \multirow{2}{*}{Supr.} & Cov. & .91 & .93 & .95 & .97 & .98 & .91 & .91 & .90 & .89 & .88 & .91 & .88 & .83 & .73 & .52 \\
  &                            & \textcolor{gray}{Bw.} & \tinygray{.0129} & \tinygray{.0129} & \tinygray{.0129} & \tinygray{.0129} & \tinygray{.0129} & \tinygray{.0133} & \tinygray{.0133} & \tinygray{.0133} & \tinygray{.0133} & \tinygray{.0133} & \tinygray{.0137} & \tinygray{.0137} & \tinygray{.0137} & \tinygray{.0137} & \tinygray{.0137} \\ \cline{2-18}
  & \multirow{2}{*}{LSCI}     & Cov. & .96 & .97 & .97 & .96 & .91 & .97 & .94 & .86 & .62 & .29 & .96 & .88 & .57 & .15 & .01 \\
  &                           & \textcolor{gray}{Bw.} & \tinygray{.0122} & \tinygray{.0123} & \tinygray{.0123} & \tinygray{.0123} & \tinygray{.0123} & \tinygray{.0124} & \tinygray{.0125} & \tinygray{.0125} & \tinygray{.0125} & \tinygray{.0125} & \tinygray{.0127} & \tinygray{.0127} & \tinygray{.0126} & \tinygray{.0125} & \tinygray{.0124} \\ \cline{2-18}
  & \multirow{3}{*}{\parbox{2cm}{WCP\\Learned}} & Cov. & .90 & .90 & .90 & .89 & .89 & .89 & .90 & .88 & .87 & .85 & .90 & .89 & .86 & .84 & .78 \\
  &                            & \textcolor{gray}{Bw.} & \tinygray{.0119} & \tinygray{.0110} & \tinygray{.0102} & \tinygray{.0095} & \tinygray{.0087} & \tinygray{.0123} & \tinygray{.0123} & \tinygray{.0122} & \tinygray{.0121} & \tinygray{.0123} & \tinygray{.0127} & \tinygray{.0140} & \tinygray{.0151} & \tinygray{.0170} & \tinygray{.0179} \\
  &                            & \textcolor{gray}{$n_\infty$} & \tinygray{0\%} & \tinygray{0\%} & \tinygray{0\%} & \tinygray{0\%} & \tinygray{0\%} & \tinygray{0\%} & \tinygray{0\%} & \tinygray{0\%} & \tinygray{0\%} & \tinygray{0\%} & \tinygray{0\%} & \tinygray{0\%} & \tinygray{0\%} & \tinygray{0\%} & \tinygray{0\%} \\ \cline{2-18}
  & \multirow{3}{*}{\parbox{2cm}{WCP\\Gauss (Ours)}} & Cov. & .91 & .91 & .90 & .89 & .88 & .91 & .91 & .91 & .95 & 1.00 & .91 & .92 & .97 & 1.00 & 1.00 \\
  &                            & \textcolor{gray}{Bw.} & \tinygray{.0129} & \tinygray{.0113} & \tinygray{.0103} & \tinygray{.0092} & \tinygray{.0086} & \tinygray{.0134} & \tinygray{.0134} & \tinygray{.0137} & \tinygray{.0156} & \tinygray{.0203} & \tinygray{.0139} & \tinygray{.0160} & \tinygray{.0208} & \tinygray{.0244} & \tinygray{--} \\
  &                            & \textcolor{gray}{$n_\infty$} & \tinygray{0\%} & \tinygray{0\%} & \tinygray{0\%} & \tinygray{0\%} & \tinygray{0\%} & \tinygray{0\%} & \tinygray{0\%} & \tinygray{0\%} & \tinygray{25\%} & \tinygray{94\%} & \tinygray{0\%} & \tinygray{0\%} & \tinygray{44\%} & \tinygray{99\%} & \tinygray{100\%} \\ \hline
\hline
\multirow{10}*{\rotatebox[origin=c]{90}{$a=-0.01$}}
  & \multirow{2}{*}{Supr.} & Cov. & .91 & .93 & .95 & .96 & .97 & .91 & .91 & .90 & .88 & .75 & .91 & .88 & .82 & .64 & .07 \\
  &                            & \textcolor{gray}{Bw.} & \tinygray{.0129} & \tinygray{.0129} & \tinygray{.0129} & \tinygray{.0129} & \tinygray{.0129} & \tinygray{.0133} & \tinygray{.0133} & \tinygray{.0133} & \tinygray{.0133} & \tinygray{.0133} & \tinygray{.0138} & \tinygray{.0138} & \tinygray{.0138} & \tinygray{.0138} & \tinygray{.0138} \\ \cline{2-18}
  & \multirow{2}{*}{LSCI}     & Cov. & .96 & .96 & .94 & .78 & .42 & .97 & .92 & .71 & .26 & .01 & .97 & .84 & .38 & .02 & .00 \\
  &                           & \textcolor{gray}{Bw.} & \tinygray{.0123} & \tinygray{.0123} & \tinygray{.0123} & \tinygray{.0123} & \tinygray{.0124} & \tinygray{.0125} & \tinygray{.0125} & \tinygray{.0125} & \tinygray{.0125} & \tinygray{.0125} & \tinygray{.0128} & \tinygray{.0127} & \tinygray{.0126} & \tinygray{.0125} & \tinygray{.0124} \\ \cline{2-18}
  & \multirow{3}{*}{\parbox{2cm}{WCP\\Learned}} & Cov. & .90 & .90 & .89 & .88 & .81 & .90 & .89 & .87 & .82 & .64 & .90 & .89 & .85 & .78 & .39 \\
  &                            & \textcolor{gray}{Bw.} & \tinygray{.0119} & \tinygray{.0111} & \tinygray{.0103} & \tinygray{.0096} & \tinygray{.0088} & \tinygray{.0123} & \tinygray{.0124} & \tinygray{.0124} & \tinygray{.0122} & \tinygray{.0123} & \tinygray{.0128} & \tinygray{.0140} & \tinygray{.0150} & \tinygray{.0165} & \tinygray{.0175} \\
  &                            & \textcolor{gray}{$n_\infty$} & \tinygray{0\%} & \tinygray{0\%} & \tinygray{0\%} & \tinygray{0\%} & \tinygray{0\%} & \tinygray{0\%} & \tinygray{0\%} & \tinygray{0\%} & \tinygray{0\%} & \tinygray{0\%} & \tinygray{0\%} & \tinygray{0\%} & \tinygray{0\%} & \tinygray{0\%} & \tinygray{0\%} \\ \cline{2-18}
  & \multirow{3}{*}{\parbox{2cm}{WCP\\Gauss (Ours)}} & Cov. & .91 & .91 & .89 & .88 & .97 & .91 & .91 & .92 & .99 & 1.00 & .91 & .92 & .99 & 1.00 & 1.00 \\
  &                            & \textcolor{gray}{Bw.} & \tinygray{.0130} & \tinygray{.0115} & \tinygray{.0103} & \tinygray{.0098} & \tinygray{.0101} & \tinygray{.0133} & \tinygray{.0134} & \tinygray{.0146} & \tinygray{.0186} & \tinygray{--} & \tinygray{.0139} & \tinygray{.0162} & \tinygray{.0222} & \tinygray{--} & \tinygray{--} \\
  &                            & \textcolor{gray}{$n_\infty$} & \tinygray{0\%} & \tinygray{0\%} & \tinygray{0\%} & \tinygray{4\%} & \tinygray{79\%} & \tinygray{0\%} & \tinygray{0\%} & \tinygray{7\%} & \tinygray{92\%} & \tinygray{100\%} & \tinygray{0\%} & \tinygray{1\%} & \tinygray{76\%} & \tinygray{100\%} & \tinygray{100\%} \\ \hline
\hline
\end{tabular}
\end{table}
\endgroup

\section{Computing resources}
All the experiments were conducted on a mix of CPUs and GPUs (A40, A100, H100), based on the availability of our academic cluster. Most runs are comfortably executed on CPUs, but for the image data, GPUs speed up the process. No run takes longer than a few minutes. For the LSCI method, it is highly advised to use the computation cache that JAX provides. Otherwise, these calculations can take far longer, even on GPUs. Memory and storage do not exceed 10 GB per dataset. The full research project required more compute than the experiments reported in the paper, due to failed runs, parameter tuning, or experiments that were not relevant enough for the final cut.

\clearpage

\end{document}